%% file: main.tex
\documentclass[10pt,journal,compsoc]{IEEEtran} %
\usepackage[english]{babel}

\ifCLASSOPTIONcompsoc
  \usepackage[nocompress]{cite}
\else
  \usepackage{cite}
\fi

\ifCLASSINFOpdf
  \usepackage[pdftex]{graphicx}
  \DeclareGraphicsExtensions{.pdf,.jpeg,.png}
\else

\fi

\usepackage[cmex10]{amsmath}
\usepackage{amsfonts}
\input{chapters/math_macros.tex}

  \usepackage[caption=false,font=footnotesize,labelfont=sf,textfont=sf]{subfig}

\usepackage{booktabs}
\usepackage{multirow}

\usepackage{color}
\usepackage{pifont}%
\newcommand{\cmark}{\ding{51}}%
\newcommand{\xmark}{\ding{55}}%

\usepackage{url}

\usepackage{siunitx}
\DeclareSIUnit{\nothing}{\relax}

\usepackage{cite} %

\newcommand\MYhyperrefoptions{bookmarks=true,bookmarksnumbered=true,
pdfpagemode={UseOutlines},plainpages=false,pdfpagelabels=true,
colorlinks=true,citecolor={black},
urlcolor={black},
pdftitle={Event-based Vision: A Survey},%
pdfsubject={Computer Vision, Robotics, Neuromorphic Engineering},%
pdfauthor={G. Gallego, T. Delbruck, G. Orchard, C. Bartolozzi, B. Taba, A. Censi, S. Leutenegger, A. Davison, J. Conradt, K. Daniilidis, D. Scaramuzza},%
pdfkeywords={Event Cameras, Bio-Inspired Vision, Asynchronous sensor, Low Latency, High Dynamic Range, Low Power}}%
\usepackage[\MYhyperrefoptions,pdftex,hidelinks]{hyperref}

\usepackage{adjustbox} %

\usepackage{etoolbox} %

\newif\ifclearsectionlook
\clearsectionlookfalse

\newif\ifclearsubseclook
\clearsubseclookfalse

\newif\iflongversion
\longversiontrue

\usepackage{xcolor}

\newif\ifblacktext
\blacktexttrue

\ifblacktext
\newcommand{\td}[1]{#1} %
\newcommand{\cb}[1]{#1} %
\newcommand{\sle}[1]{#1} %
\else
\newcommand{\td}[1]{\textcolor{magenta}{TD: #1}} %
\newcommand{\cb}[1]{\textcolor{violet}{CB: #1}} %
\newcommand{\sle}[1]{\textcolor{teal}{SL: #1}} %
\fi

\usepackage[draft,authormarkup=none]{changes}
\definechangesauthor[color=blue]{GG}
\definechangesauthor[name=TD, color=magenta]{TD}
\definechangesauthor[name=GO, color=orange]{GO}
\definechangesauthor[name=CB, color=violet]{CB}

\definecolor{somegray}{rgb}{0.5, 0.5, 0.5}
\newcommand{\darkgrayed}[1]{\textcolor{somegray}{#1}}
\makeatletter
\newcommand*\titleheader[1]{\gdef\@titleheader{#1}}
\AtBeginDocument{%
  \let\st@red@title\@title
  \def\@title{%
    \vskip-1.3em
    \bgroup\normalfont\large\centering\@titleheader\par\egroup
    \vskip1.5em\st@red@title}
}
\makeatother

\titleheader{\darkgrayed{This paper has been accepted for publication at the\\
IEEE Transactions on Pattern Analysis and Machine Intelligence, 2020.
\copyright IEEE}}

\title{Event-based Vision: A Survey}

\begin{document}

\author{Guillermo~Gallego,
Tobi~Delbr\"uck,
Garrick~Orchard,
Chiara~Bartolozzi,
Brian~Taba,
Andrea~Censi,
Stefan~Leutenegger,
Andrew~J.~Davison,
J\"org~Conradt,
Kostas~Daniilidis,
Davide~Scaramuzza%
\thanks{G. Gallego is with the Technische Universit\"at Berlin, Berlin, Germany. 
Tobi Delbruck is with the Dept. of Information Technology and Electrical Engineering, ETH Zurich, at the Inst. of Neuroinformatics, University of Zurich and ETH Zurich, Zurich, Switzerland.
Garrick Orchard is with Intel Labs, CA, USA.
Chiara Bartolozzi is with the Italian institute of Technology, Genoa, Italy.
Brian Taba is with IBM Research, CA, USA.
Andrea Censi is with the Dept. of Mechanical and Process Engineering, ETH Zurich, Switzerland.
Stefan Leutenegger and Andrew Davison are with Imperial College London, London, UK.
J\"org Conradt is with KTH Royal Institute of Technology, Stockholm, Sweden.
Kostas Daniilidis is with University of Pennsylvania, PA, USA.
D. Scaramuzza is with the Dept. of Informatics University of Zurich and Dept. of Neuroinformatics, University of Zurich and ETH Zurich, Switzerland.
}%
\thanks{G. Gallego and D. Scaramuzza were supported by the SNSF-ERC Starting Grant and the Swiss National Science Foundation through the National Center of Competence in Research (NCCR) Robotics.}
\thanks{Contact e-mail: \href{mailto:guillermo.gallego@tu-berlin.de?subject=EventVisionSurveyPaper}{guillermo.gallego@tu-berlin.de}}
}

\markboth{}%
{Gallego \MakeLowercase{\textit{et al.}}: Event-based Vision: A Survey}

\IEEEtitleabstractindextext{%
\begin{abstract}
\input{chapters/abstract.tex}
\end{abstract}

\begin{IEEEkeywords}
Event Cameras, Bio-Inspired Vision, Asynchronous Sensor, Low Latency, High Dynamic Range, Low Power.
\end{IEEEkeywords}}

\maketitle

\IEEEdisplaynontitleabstractindextext

\IEEEpeerreviewmaketitle

\input{content.tex}

\iflongversion
\ifCLASSOPTIONcompsoc
  \section*{Acknowledgments}
\else
  \section*{Acknowledgment}
\fi
The authors would like to thank all the people who contributed to this paper. 
We would like to thank the event camera manufacturers for providing the values in Table~\ref{tab:devices} and for discussing the difficulties in their comparison due to the lack of a common testbed. 
In particular, we thank Hyunsurk Eric Ryu (Samsung Electronics), Chenghan Li (iniVation), Davide Migliore (Prophesee), Marc Osswald (Insightness) and Prof. Chen (CelePixel).
We are also thankful to all members of our research laboratories, for discussion and comments on early versions of this document.
We also thank the editors and anonymous reviewers of IEEE TPAMI for their suggestions, which led us to improve the paper.
\fi

\ifCLASSOPTIONcaptionsoff
  \newpage
\fi

\bibliographystyle{IEEEtran}
%\bibliography{all,event}
\input{main.bbl}

\iflongversion
\else
\input{chapters/biographies.tex}

\fi

\end{document}

%% file: chapters/math_macros.tex
\global\long\def\Lum{L}
\global\long\def\vel{\mathbf{v}}
\global\long\def\prtl#1#2{\frac{\partial#1}{\partial#2}}

\global\long\def\bx{\mathbf{x}}

\global\long\def\pol{p}

%% file: chapters/abstract.tex
Event cameras are bio-inspired sensors that differ from conventional frame cameras:
Instead of capturing images at a fixed rate, they asynchronously measure per-pixel brightness changes, and output a stream of events that encode the time, location and sign of the brightness changes.
Event cameras offer attractive properties compared to traditional cameras:
high temporal resolution (in the order of \si{\micro\second}),
very high dynamic range (\SI{140}{\decibel} vs. \SI{60}{\decibel}), 
low power consumption, and high pixel bandwidth (on the order of \si{\kilo\hertz}) resulting in reduced motion blur.
Hence, event cameras have a large potential for robotics and computer vision in challenging scenarios for traditional cameras, such as low-latency, high speed, and high dynamic range.
However, novel methods are required to process the unconventional output of these sensors in order to unlock their potential.
This paper provides a comprehensive overview of the emerging field of event-based vision, with a focus on the applications and the algorithms developed to unlock the outstanding properties of event cameras.
We present event cameras from their working principle, the actual sensors that are available and the tasks that they have been used for, from low-level vision (feature detection and tracking, optic flow, etc.) to high-level vision (reconstruction, segmentation, recognition).
We also discuss the techniques developed to process events, including learning-based techniques, as well as specialized processors for these novel sensors, such as spiking neural networks.
Additionally, we highlight the challenges that remain to be tackled and the opportunities that lie ahead in the search for a more efficient, bio-inspired way for machines to perceive and interact with the world.

%% file: content.tex
\ifclearsectionlook\cleardoublepage\fi \input{chapters/01_introduction.tex}
\ifclearsectionlook\cleardoublepage\fi \input{chapters/02_working_principle.tex}

\ifclearsectionlook\cleardoublepage\fi \input{chapters/03_event_processing.tex}

\ifclearsectionlook\cleardoublepage\fi \input{chapters/04_algorithms.tex}

\ifclearsectionlook\cleardoublepage\fi \input{chapters/05_systems_demonstrators.tex}
\iflongversion
\ifclearsectionlook\cleardoublepage\fi \input{chapters/06_resources.tex}
\fi
\ifclearsectionlook\cleardoublepage\fi \input{chapters/07_discussion.tex}

\ifclearsectionlook\cleardoublepage\fi \input{chapters/08_conclusion.tex}

%% file: chapters/01_introduction.tex
\section{Introduction and Applications}
\label{sec:introduction}
\IEEEPARstart{``T}{he} 
\emph{brain is imagination, and that was exciting to me;
I wanted to build a chip that could imagine something\footnote{\url{https://youtu.be/FKemf6Idkd0?t=67}}.}''
that is how Misha Mahowald, a graduate student at Caltech in 1986,
started to work with Prof.~Carver Mead on the stereo problem from a joint biological and engineering perspective.
A couple of years later, in 1991, the image of a cat in the cover of Scientific American~\cite{Mahowald91sciamer}, acquired by a novel ``Silicon Retina'' mimicking the neural architecture of the eye, showed a new, powerful way of doing computations, igniting the emerging field of neuromorphic engineering.
Today, we still pursue the same visionary challenge:
understanding how the brain works and building one on a computer chip.
Current efforts include flagship billion-dollar projects, such as the Human Brain Project and the Blue Brain Project in Europe and the U.S. BRAIN (Brain Research through Advancing Innovative Neurotechnologies) Initiative.

This paper provides an overview of the bio-inspired technology of silicon retinas,  or ``event cameras'', such as~\cite{Lichtsteiner08ssc,Posch11ssc,Brandli14ssc,Son17isscc}, with a focus on their application to solve classical as well as new computer vision and robotic tasks.
Sight is, by far, the dominant sense in humans to perceive the world,
and, together with the brain, learn new things.
In recent years, this technology has attracted a lot of attention from academia and industry. 
This is due to the availability of prototype event cameras and the advantages that they offer to tackle problems that are difficult with standard frame-based image sensors %
(that provide stroboscopic synchronous sequences of pictures), 
such as high-speed motion estimation~\cite{Delbruck07iscas,Gallego17pami} or high dynamic range (HDR) imaging~\cite{Rebecq19pami}.

Event cameras are \emph{asynchronous} sensors that pose a \emph{paradigm shift} in the way visual information is acquired.
This is because they sample light based on the scene dynamics, rather than on a clock that has no relation to the viewed scene.
Their advantages are:
very high temporal resolution and low latency (both in the order of microseconds), very high dynamic range (\SI{140}{\decibel} vs. \SI{60}{\decibel} of standard cameras), and low power consumption.
Hence, event cameras have a large potential for robotics and wearable applications in challenging scenarios for standard cameras, such as high speed and high dynamic range.
Although event cameras have become commercially available
only since 2008~\cite{Lichtsteiner08ssc}, the recent body of literature on these new sensors\footnote{\url{https://github.com/uzh-rpg/event-based_vision_resources} \cite{Gallego17resources}} as well as the recent plans for mass production claimed by companies, such as Samsung~\cite{Son17isscc} and Prophesee\footnote{\url{http://rpg.ifi.uzh.ch/ICRA17_event_vision_workshop.html}}, highlight that there is a big commercial interest in exploiting these novel vision sensors for mobile robotic, augmented and virtual reality (AR/VR), and video game applications.
However, because event cameras work in a fundamentally different way from standard cameras,
measuring per-pixel brightness changes (called ``\emph{events}'') asynchronously rather than measuring ``absolute'' brightness at constant rate,
novel methods are required to process their output and unlock their potential.

\textbf{Applications of Event Cameras}:
\label{sec:applications}
Typical scenarios where event cameras offer advantages over other sensing modalities include real-time interaction systems, such as robotics or wearable electronics~\cite{Delbruck16essderc}, where operation under uncontrolled lighting conditions, latency, and power are important~\cite{Liu19msp}.
Event cameras are used for object tracking~\cite{Delbruck13fns,Glover16iros},
surveillance and monitoring~\cite{Litzenberger06itsc},
and object/gesture recognition~\cite{Orchard15pami,Lee14tnnls,Amir17cvpr}.
They are also profitable for depth estimation~\cite{Rogister12tnnls,Rebecq18ijcv},
structured light 3D scanning~\cite{Matsuda15iccp},
optical flow estimation~\cite{Benosman14tnnls,Zhu18rss}, 
HDR image reconstruction~\cite{Cook11ijcnn,Kim14bmvc,Rebecq19pami}
and Simultaneous Localization and Mapping (SLAM)~\cite{Kim16eccv,Rebecq17ral,Rosinol18ral}.
Event-based vision is a growing field of research, and other applications, such as image deblurring~\cite{Pan19cvpr} or star tracking~\cite{Cohen17amos,Chin19cvprw}, will appear as event cameras become widely available~\cite{Gallego17resources}.

\iflongversion
\textbf{Outline}:
The rest of the paper is organized as follows.
Section~\ref{sec:eventCameras} presents event cameras, their working principle and advantages, and the challenges that they pose as novel vision sensors.
Section~\ref{sec:event_processing} discusses several methodologies commonly used to extract information from the event camera output, and discusses the biological inspiration behind some of the approaches.
Section~\ref{sec:algorithms} reviews applications of event cameras, from low-level to high-level vision tasks, and some of the algorithms that have been designed to unlock their potential.
Opportunities for future research and open challenges on each topic are also pointed out.
Section~\ref{sec:systems_and_demonstrators} 
presents neuromorphic processors and embedded systems.
Section~\ref{sec:resources} reviews the software, datasets and simulators to work on event cameras, as well as additional sources of information.
The paper ends with a discussion (Section~\ref{sec:discussion}) 
and conclusions (Section~\ref{sec:conclusion}).
\fi 

%% file: chapters/02_working_principle.tex
\section{Principle of Operation of Event Cameras}
\label{sec:eventCameras}

                                      \input{chapters/020_working_principle_intro.tex}

\ifclearsubseclook\cleardoublepage\fi \input{chapters/021_event_camera_types.tex}
\ifclearsubseclook\cleardoublepage\fi \input{chapters/022_advantages_event_camera.tex}
                                      \input{chapters/023_paradigm_shift_challenges.tex}
\ifclearsubseclook\cleardoublepage\fi \input{chapters/024_event_generation_modeling.tex}

\ifclearsubseclook\cleardoublepage\fi \input{chapters/025_camera_devices.tex}

%% file: chapters/020_working_principle_intro.tex
In contrast to standard cameras, which acquire full images at a rate specified by an external clock (e.g., \SI{30}{fps}), event cameras, such as the Dynamic Vision Sensor (\textbf{DVS}) \cite{Lichtsteiner05wccdais,Lichtsteiner05rme,Lichtsteiner06thesis,Lichtsteiner06isscc,Lichtsteiner08ssc},
respond to \emph{brightness changes} in the scene \emph{asynchronously} and \emph{independently} for every pixel (Fig.~\ref{fig:DVSsummary}b).
Thus, the output of an event camera is a variable data-rate sequence of digital ``events'' or ``spikes'', with each event
representing a change of brightness (log intensity)\footnote{\emph{Brightness} is a perceived quantity; for brevity we use it to refer to log intensity since they correspond closely for uniformly-lighted scenes.} 
of predefined magnitude at a pixel at a particular time\footnote{Nomenclature: 
``Event cameras'' output data-driven events that signal a place and time. 
This nomenclature has evolved over the past decade: 
originally they were known as address-event representation (AER) silicon retinas, and later they became event-based cameras. 
In general, events can signal any kind of information (intensity, local spatial contrast, etc.), but over the last five years or so, the term ``event camera'' has unfortunately become practically synonymous with the particular representation of brightness change output by DVS's.} (Fig.~\ref{fig:DVSsummary}b) (Section~\ref{subsec:event_generation_model}).
This encoding is inspired by the spiking nature of biological visual pathways (Section~\ref{sec:bioinspired_processing}).

Each pixel memorizes the log intensity each time it sends an event,
and continuously monitors for a change of sufficient magnitude from this memorized value (Fig.~\ref{fig:DVSsummary}a).
When the change exceeds a threshold, the camera sends an event,
which is transmitted from the chip with the~$x,y$ location, the time~$t$, and
the 1-bit polarity~$p$ of the change (i.e., brightness increase (``ON'') or decrease (``OFF'')). 
This event output is illustrated in Figs.~\ref{fig:DVSsummary}b, \ref{fig:DVSsummary}e and~\ref{fig:DVSsummary}f.

\input{chapters/fig_DAVIS_summary.tex}

The events are transmitted from the pixel array to periphery and then out of the camera using a shared digital output bus, typically by using address-event representation (\textbf{AER}) readout~\cite{Boahen04tcsi,Liu15book}.
This bus can become saturated, which perturbs the times that events are sent.
Event cameras have readout rates ranging from \SI{2}{\mega\Hz}~\cite{Lichtsteiner08ssc} to \SI{1200}{\mega\Hz}~\cite{Suh20iscas}, depending on the chip and type of hardware interface.

Event cameras are data-driven sensors:
their output depends on the amount of motion or brightness change in the scene.
The faster the motion, the more events per second are generated, since each pixel adapts
its delta modulator sampling rate to the rate of change of the log intensity signal that it monitors.
Events are timestamped with microsecond resolution and are transmitted
with sub-millisecond latency, which make these sensors react quickly to visual stimuli.

The incident light at a pixel is a product of scene illumination and surface reflectance. 
If illumination is approximately constant, a log intensity change signals a reflectance change.
These changes in reflectance are mainly the result of the movement of objects in the field of view.
That is why the DVS brightness change events have a built-in invariance to scene illumination~\cite{Lichtsteiner08ssc}.

\textbf{Comparing Bandwidths of DVS Pixels and Frame-based Camera}:
Although DVS pixels are fast, like any physical transducer, they have a finite bandwidth: 
if the incoming light intensity varies too quickly, the front-end photoreceptor circuits filter out the variations~\cite{Delbruck2020-arxiv-v2e}. 
The rise and fall time that is analogous to the exposure time in standard image sensors is the reciprocal of this bandwidth. 
Fig.~\ref{fig:DVSbandwidth} shows an example of measured DVS pixel frequency response (DVS128 in~\cite{Lichtsteiner08ssc}). 
The measurement setup (Fig.~\ref{fig:DVSbandwidth:setup}) uses a sinusoidally-varying generated signal to measure the response. 
Fig.~\ref{fig:DVSbandwidth:response} shows that, at low frequencies, the DVS pixel produces a certain number of events per cycle. 
Above some cutoff frequency, the variations are filtered out by the photoreceptor dynamics, and thus the number of events per cycle drops. 
This cutoff frequency is a monotonically increasing function of light intensity. 
At the brighter light intensity, the DVS pixel bandwidth is about \SI{3}{\kilo\hertz}, equivalent to an exposure time of about \SI{300}{\micro\second}.
At 1000$\times$ lower intensity, the DVS bandwidth is reduced to about \SI{300}{\hertz}. 
Even when the LED brightness is reduced by a factor of 1000, the frequency response of DVS pixels is ten times higher than the \SI{30}{\hertz} Nyquist frequency from a \SI{60}{fps} image sensor.
Also, the frame-based camera aliases frequencies above the Nyquist frequency back to the baseband, whereas the DVS pixel does not due to the continuous time response.

\input{chapters/fig_DVS_bandwidth.tex}

%% file: chapters/fig_DAVIS_summary.tex
\begin{figure}[t]
\centering
\includegraphics[width=\linewidth]{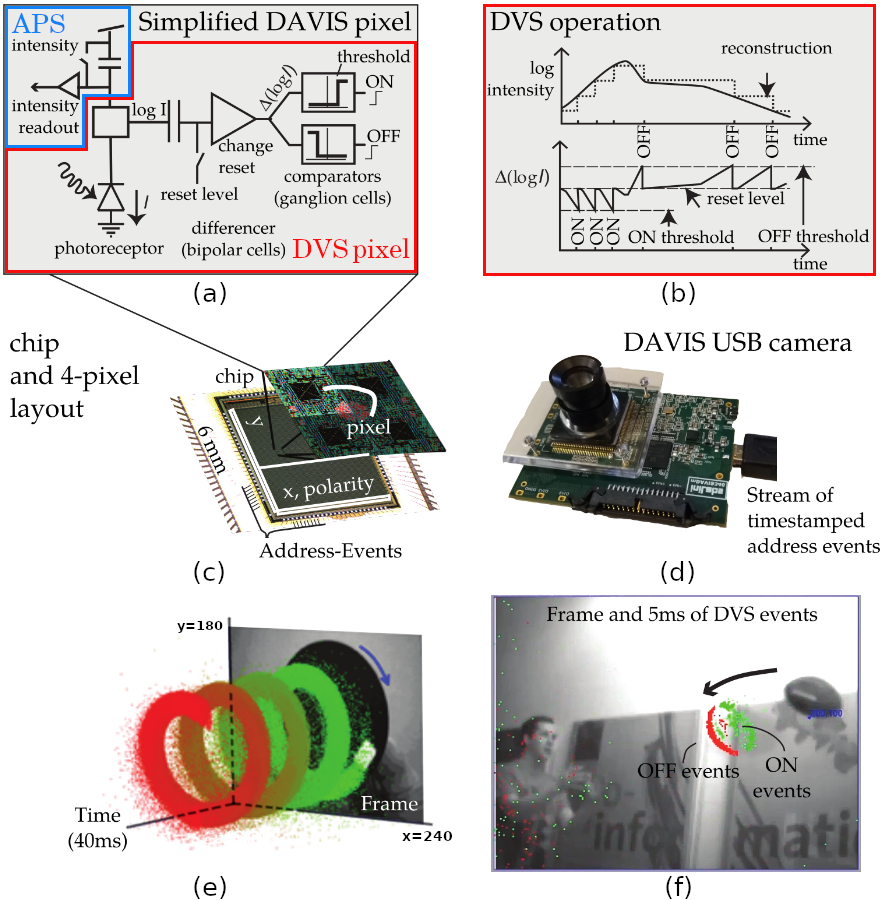}
\vspace{-2ex}
\caption{Summary of the DAVIS camera~\cite{Brandli14ssc}, comprising an event-based dynamic vision sensor (DVS~\cite{Lichtsteiner08ssc}) and a frame-based active pixel sensor (\textbf{APS}) in the same pixel array, sharing the same photodiode in each pixel.
\textbf{(a)} Simplified circuit diagram of the DAVIS pixel (DVS pixel in red, APS pixel in blue).
\textbf{(b)} Schematic of the operation of a DVS pixel, converting light into events.
\textbf{(c)-(d)} Pictures of the DAVIS chip and USB camera.
\textbf{(e)} 
A white square on a rotating black disk viewed by the DAVIS produces grayscale frames and a spiral of events in space-time.
Events in space-time are color-coded, from green (past) to red (present).
\textbf{(f)} Frame and overlaid events of a natural scene;
the frames lag behind the low-latency events (colored according to polarity).
Images adapted from~\cite{Neil17thesis,Brandli14ssc}.
A more in-depth comparison of the DVS, DAVIS and ATIS pixel designs can be found  in~\cite{Posch14ieee}.
\label{fig:DVSsummary}}
\vspace{-1ex}
\end{figure}

%% file: chapters/fig_DVS_bandwidth.tex
\begin{figure}[t]
\centering
\subfloat[\label{fig:DVSbandwidth:setup}Measurement setup]{
\raisebox{0.25\height}{\includegraphics[width=0.45\linewidth]{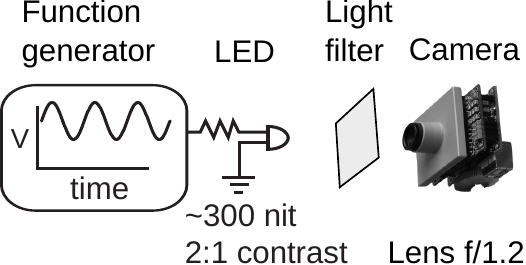}}}\;\;
\subfloat[\label{fig:DVSbandwidth:response}Measured responses for two DC levels of illumination.]{\includegraphics[width=0.44\linewidth]{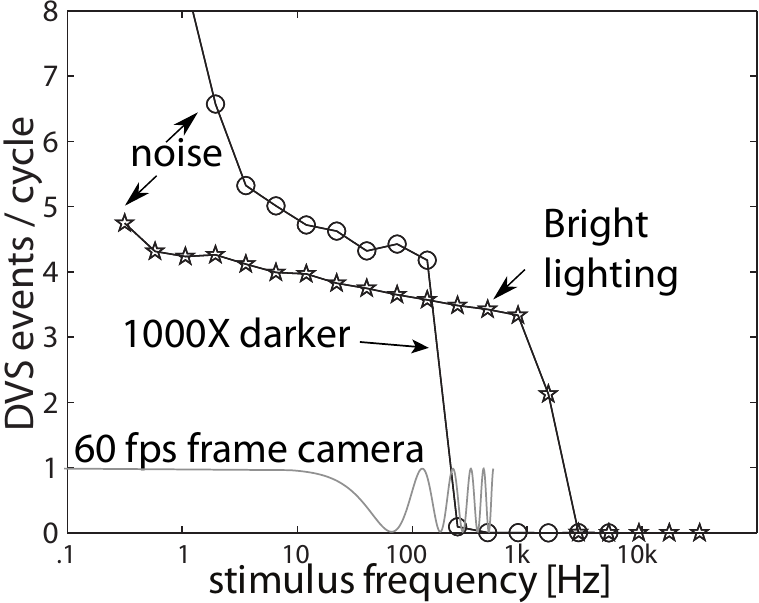}}
\vspace{-1ex}
\caption{``Event transfer function'' from a single DVS pixel in response to sinusoidal LED stimulation.
The background events cause additional ON events at very low frequencies. 
The 60\,fps camera curve shows the transfer function 
including aliasing from frequencies 
above the Nyquist frequency. 
Figure adapted from~\cite{Lichtsteiner08ssc}.}
\label{fig:DVSbandwidth}
\end{figure}

%% file: chapters/021_event_camera_types.tex
\subsection{Event Camera Designs}
\label{subsec:event_camera_types}

This section presents the most common event camera designs.
The actual devices (commercial or prototype cameras such as the DAVIS240) are summarized in Section~\ref{sec:physical_devices}.

The first silicon retina was developed by Mahowald and Mead at Caltech during the period 1986-1992, in Ph.D. thesis work~\cite{Mahowald92thesis} that was awarded the prestigious Clauser prize\footnote{\url{http://www.gradoffice.caltech.edu/current/clauser}}.
Mahowald and Mead's sensor had logarithmic pixels, was modeled after the three-layer Kufler retina, and produced as output spike events using the AER protocol.
However, it suffered from several shortcomings: 
each wire-wrapped retina board required precise adjustment of biasing potentiometers;
there was  considerable mismatch between the responses of different pixels; and pixels were too large to be a device of practical use.
Over the next decade the neuromorphic community developed a series of silicon retinas. 
These developments are summarized in~\cite{Liu15book,Posch14ieee,Dong15sam,Steffen19fnbot}.

The \emph{\textbf{DVS} event camera}~\cite{Lichtsteiner08ssc} had its genesis in a frame-based silicon retina design where the continuous-time photoreceptor was capacitively coupled to a readout circuit that was reset each time the pixel was sampled~\cite{Delbruck91spie}.
More recent event camera technology has been reviewed in the electronics and neuroscience literature~\cite{Liu10nb,Delbruck10iscas,Delbruck12eccvw,Posch14ieee,Liu15book,Delbruck16essderc}. 
Although surprisingly many applications can be solved by only processing DVS events (i.e., brightness changes),
it became clear that some also require some form of static output (i.e., ``absolute'' brightness).
To address this shortcoming, there have been several developments
of cameras that concurrently output dynamic and static information.

The \emph{Asynchronous Time Based Image Sensor} (\textbf{ATIS}) \cite{Posch10isscc,Posch11ssc}
has pixels that contain a DVS subpixel (called change detection CD) that triggers another subpixel to read out the absolute intensity (exposure measurement EM).
The trigger resets a capacitor to a high voltage.
The charge is bled away from this capacitor by another photodiode.
The brighter the light, the faster the capacitor discharges.
The ATIS intensity readout transmits two more events coding the time between crossing two threshold voltages, as in~\cite{Culurciello03}.
This way, only pixels that change provide their new intensity values.
The brighter the illumination, the shorter the time between these two events. 
The ATIS achieves large static dynamic range ($>$\SI{120}{\decibel}).
However, the ATIS has the disadvantage that pixels are at least double the area of DVS pixels.
Also, in dark scenes the time between
the two intensity events can be long and the readout of intensity can be interrupted by
new events (\cite{Orchard14iscas} proposes a workaround to this problem).

The widely-used \emph{Dynamic and Active Pixel Vision Sensor} (\textbf{DAVIS})~\cite{Berner13vlsi,Brandli14ssc} illustrated in~Fig.~\ref{fig:DVSsummary} combines a conventional active pixel sensor (\textbf{APS})~\cite{Fossum97ted} in the same pixel with DVS.
The advantage over ATIS is a much smaller pixel size since the photodiode is shared and the readout circuit only adds about \SI{5}{\percent} to the DVS pixel area.
Intensity (APS) frames can be triggered at a constant frame rate or on demand, by analysis of DVS events, although the latter is seldom exploited\footnote{\url{https://github.com/SensorsINI/jaer/blob/master/src/eu/seebetter/ini/chips/davis/DavisAutoShooter.java}}.
However, the APS readout has limited dynamic range (\SI{55}{dB}) and like a standard camera, it is redundant if the pixels do not change.

Since the ATIS and DAVIS pixel designs include a DVS pixel (change detector) \cite{Posch14ieee} we often use the term ``DVS'' to refer to the binary-polarity event output or circuitry, regardless of whether it is from a DVS, ATIS or DAVIS~design.

%% file: chapters/022_advantages_event_camera.tex
\subsection{Advantages of Event cameras}
\label{sec:advantageseventcameras}
Event cameras offer numerous potential advantages over standard cameras:
\\ \indent
\emph{High Temporal Resolution}:
monitoring of brightness changes is fast, in analog circuitry,
and the read-out of the events is digital, with a \SI{1}{\mega\Hz} clock,
i.e., events are detected and timestamped with microsecond resolution.
Therefore, event cameras can capture very fast motions,
without suffering from motion blur typical of frame-based cameras.
\\ \indent
\emph{Low Latency}: each pixel works independently and there is no need to wait for a global exposure time of the frame: as soon as the change is detected, it is transmitted.
Hence, event cameras have minimal latency: about \SI{10}{\micro\second} on the lab bench, and sub-millisecond in the real world.
\\ \indent
\emph{Low Power}:
Because event cameras transmit only brightness changes, and thus remove redundant data, power is only used to process changing pixels.
At the die level, most cameras use about \SI{10}{\milli\watt}, and there are prototypes that achieve less than \SI{10}{\micro\watt}.
Embedded event-camera systems where the sensor is directly interfaced to a processor have shown system-level power consumption (i.e., sensing plus processing) of \SI{100}{\milli\watt} or less~\cite{Serrano-Gotarredona09tnn,Conradt09iscas,Xu17cvpr,Amir17cvpr}.

\emph{High Dynamic Range (HDR)}.
The very high dynamic range of event cameras ($>$\SI{120}{\decibel}) notably exceeds the \SI{60}{\decibel} of high-quality, frame-based cameras, making them able to acquire information from moonlight to daylight.
It is due to the facts that the photoreceptors of the pixels operate in logarithmic scale
and each pixel works independently, not waiting for a global shutter.
Like biological retinas, DVS pixels can adapt to very dark as well as very bright stimuli.

%% file: chapters/023_paradigm_shift_challenges.tex
\subsection{Challenges Due To The Novel Sensing Paradigm}
\label{subsec:challenges_paradigm_shift}

Event cameras represent a paradigm shift in acquisition of visual information.
Hence, they pose the challenge of designing novel methods (algorithms and hardware) to process the acquired data and extract information from it in order to unlock the advantages of the camera.
Specifically:
\\ \indent
1) \emph{Coping with different space-time output:}
The output of event cameras is fundamentally different from that of standard cameras: events are asynchronous and spatially sparse, whereas images are synchronous and dense.
Hence, frame-based vision algorithms designed for image sequences are not directly applicable to event data.
\\ \indent 2) \emph{Coping with different photometric sensing:}
In contrast to the grayscale information that standard cameras provide,
each event contains binary (increase/decrease) brightness change information. 
Brightness changes depend not only on the scene brightness, but also on the current and past relative motion between the scene and the camera.
\\ \indent 3) \emph{Coping with noise and dynamic effects:}
All vision sensors are noisy because of the inherent shot noise in photons and from transistor circuit noise, 
and they also have non-idealities.
This situation is especially true for event cameras, where the process of quantizing temporal contrast is complex and has not been completely characterized.

Therefore, new methods need to rethink the space-time, photometric and stochastic nature of event data.
This poses the following questions:
What is the best way to extract information from the events relevant for a given task? and 
How can noise and non-ideal effects be modeled to better extract meaningful information from the events?

%% file: chapters/024_event_generation_modeling.tex
\subsection{Event Generation Model}
\label{subsec:event_generation_model}

An event camera~\cite{Lichtsteiner08ssc} has independent pixels that respond to changes in their log photocurrent~$\Lum \doteq \log(I)$ (``brightness'').
Specifically, in a noise-free scenario, an event~$e_k \doteq (\bx_k,t_k,\pol_k)$ is triggered at pixel~$\bx_k \doteq (x_k,y_k)^\top$ and at time~$t_k$ as soon as the brightness increment %
since the last event at the pixel, i.e.
\begin{equation}
\label{eq:BrightnessIncrment}
\Delta \Lum(\bx_k,t_k) \doteq \Lum(\bx_k,t_k) - \Lum(\bx_k,t_k-\Delta t_k),
\end{equation}
reaches a temporal contrast threshold $\pm C$ (Fig.~\ref{fig:DVSsummary}b), i.e.,
\begin{equation}
\label{eq:EventTriggeringCondition}
\Delta \Lum(\bx_k,t_k) = \pol_k \,C,
\end{equation}
where $C > 0$, ~$\Delta t_k$ is the time elapsed since the last event at the same pixel,
and the polarity~$\pol_k \in \{+1,-1\}$ is the sign of the brightness change~\cite{Lichtsteiner08ssc}.

\textbf{The contrast sensitivity} $C$ is determined by the pixel bias currents~\cite{Nozaki17ted,Nozaki18ted}, which set the speed and threshold voltages of the change detector in Fig.~\ref{fig:DVSsummary}
and are generated by an on-chip digitally-programmed bias generator.
The sensitivity~$C$ can be estimated knowing these currents~\cite{Nozaki17ted}.
In practice, positive (``ON'') and negative (``OFF'')
events may be triggered according to different thresholds, $C^+, C^-$.
Typical DVS's \cite{Lichtsteiner08ssc,Son17isscc} can set thresholds between \SIrange{10}{50}{\percent} illumination change. 
The lower limit on~$C$ is determined by noise and pixel-to-pixel mismatch (variability); setting~$C$ too low results in a storm of noise events, starting from pixels with low values of $C$.
Experimental DVS's with higher photoreceptor gain are capable of lower thresholds, e.g., \SI{1}{\percent}~\cite{Serrano-Gotarredona13ijssc,Yang15ssc,Moeys17tbcas};
however these values are only obtained under very bright illumination and ideal conditions.
Fundamentally, the pixel must react to a small change in the photocurrent in spite of the shot noise present in this current.
This shot noise limitation sets the relation between
threshold and speed of the DVS under a particular illumination
and desired detection reliability condition~\cite{Rose73book,Moeys17tbcas}.

\textbf{Events and the Temporal Derivative of Brightness}: Eq.~\eqref{eq:EventTriggeringCondition} states that event camera pixels set a threshold on magnitude of the brightness change since the last event happened.
For a small $\Delta t_k$, such an increment~\eqref{eq:EventTriggeringCondition} can be approximated using Taylor's expansion by
$\Delta\Lum (\bx_k,t_k) \approx \prtl{\Lum}{t}(\bx_k,t_k) \Delta t_k$,
which allows us to interpret the events as providing information about the temporal derivative:
\begin{equation}
\label{eq:TemporalDerivBrightness}
\prtl{\Lum}{t}(\bx_k,t_k) \approx \frac{\pol_k\, C}{\Delta t_k}.
\end{equation}
This is an indirect way of measuring brightness, since with standard cameras we are used to measuring absolute brightness.
Note that DVS events are triggered by a change in brightness magnitude~\eqref{eq:EventTriggeringCondition}, not by the brightness derivative~\eqref{eq:TemporalDerivBrightness} exceeding a threshold.
The above interpretation may be taken into account to design physically-grounded event-based algorithms, such as~\cite{Cook11ijcnn,Kim14bmvc,Gallego17pami,Scheerlinck18accv,Scheerlinck19ral,Gehrig19ijcv,Bryner19icra,Pan19cvpr}, 
as opposed to algorithms that simply process events as a collection of points with vague photometric meaning.

\textbf{Events are Caused by Moving Edges}:
Assuming constant illumination, linearizing~\eqref{eq:EventTriggeringCondition}
and using the brightness constancy assumption one can show that events are caused by moving edges.
For small~$\Delta t$, the intensity increment~\eqref{eq:EventTriggeringCondition} can be approximated by\footnote{Eq.~\eqref{eq:brightnessIncrementLinearized} can be shown~\cite{Gallego15arxiv} by
substituting the brightness constancy assumption (i.e., optical flow constraint)\label{eq_brightness_constancy}
$\prtl{\Lum}{t}(\bx(t),t) + \nabla \Lum(\bx(t),t) \cdot \dot{\bx}(t) = 0,$
with image-point velocity $\vel \equiv \dot{\bx}$,
in Taylor's approximation
$\Delta \Lum (\bx,t) \doteq \Lum (\bx,t) - \Lum (\bx,t - \Delta t)
\approx \prtl{\Lum}{t}(\bx,t) \Delta t$.}:
\begin{equation}
\label{eq:brightnessIncrementLinearized}
\Delta \Lum \approx - \nabla \Lum \cdot \vel \Delta t,
\end{equation}
that is, it is caused by a brightness gradient~$\nabla\Lum(\bx_k,t_k) = (\partial_x \Lum, \partial_y\Lum)^\top$ moving with velocity~$\vel(\bx_k,t_k)$ on the image plane, over a displacement~$\Delta \bx \doteq \vel \Delta t$.
\iflongversion
As the dot product~\eqref{eq:brightnessIncrementLinearized} conveys:
(\emph{i}) if the motion is parallel to the edge, no event is generated since~$\vel \cdot \nabla \Lum = 0$;
(\emph{ii}) if the motion is perpendicular to the edge ($\vel \parallel \nabla \Lum$) events are generated at the highest rate (i.e., minimal time is required to achieve a brightness change of size~$|C|$).
\fi

\input{chapters/table_devices.tex}

\textbf{Probabilistic Event Generation Models}:
Equation~\eqref{eq:EventTriggeringCondition} is an idealized model for the generation of events.
A more realistic model takes into account sensor noise and transistor mismatch, yielding a mixture of frozen and temporally varying stochastic triggering conditions represented by a probability function, which is itself a complex function of local illumination level and sensor operating parameters.
The measurement of such probability density was shown in~\cite{Lichtsteiner08ssc} (for the DVS128), suggesting a normal distribution centered at the contrast threshold $C$. 
The 1$\sigma$ width of the distribution is typically 2-4\% temporal contrast.
This event generation model can be included in emulators~\cite{Katz12iscas} and simulators~\cite{Rebecq18corl} of event cameras,
and in event processing algorithms~\cite{Kim14bmvc,Gallego15arxiv}.
Other probabilistic event generation models have been proposed, such as:
the likelihood of event generation being proportional to the magnitude of the image gradient~\cite{Censi14icra} (for scenes where large intensity gradients are the source of most event data),
or the likelihood being modeled by a mixture distribution to be robust to sensor noise~\cite{Gallego17pami}. 
Future even more realistic models may include the refractory
period (i.e., the duration in time that the pixel ignores log brightness changes after it has generated an event; the larger the refractory period the fewer events are produced by fast moving objects), and bus congestion~\cite{Yang17tcsi}.

\iflongversion
The above event generation models are simple, developed to some extent based on sensor noise characterization.
Just like standard image sensors, 
DVS's also have fixed pattern noise (\textbf{FPN}\footnote{\url{https://en.wikipedia.org/wiki/Fixed-pattern_noise}}), 
but in DVS it manifests as pixel-to-pixel variation
in the event threshold. 
Standard DVS's can achieve 
minimum~$C\approx \pm\SI{15}{\percent}$, 
with a standard deviation of about \SIrange{2.5}{4}{\percent} contrast between pixels~\cite{Lichtsteiner08ssc,Posch11iscas},
and there have been attempts to measure pixelwise thresholds by comparing brightness changes due to DVS events and due to differences of consecutive DAVIS APS frames~\cite{Brandli14iscas}.
However, understanding of \emph{temporal} DVS pixel and readout noise is preliminary~\cite{Lichtsteiner08ssc,Yang14iscas,Yang15ssc,Yang17tcsi},
and noise filtering methods have been developed mainly based on computational efficiency, assuming that events from real objects should be more correlated spatially and temporally than noise events~\cite{Delbruck08issle,Liu15book,Khodamoradi18tetc,Czech16biorob,Padala18fns}.
We are far from having a model that can predict event camera noise statistics under arbitrary illumination and biasing conditions.
Solving this challenge would lead to better estimation methods.
\fi

%% file: chapters/table_devices.tex
\begin{table*}[t!]
\centering\caption{\label{tab:devices}Comparison of commercial or prototype event cameras.
Values are approximate since there is no standard measurement testbed.}
\vspace{-1ex}
\begin{adjustbox}{max width=\textwidth}
\setlength{\tabcolsep}{2pt}
\begin{tabular}{l|l|c|c|c|c|c|c|c|c|c|c|c|c|c}
\toprule
 \multicolumn{2}{l|}{Supplier} & \multicolumn{3}{c|}{iniVation} & \multicolumn{4}{c|}{Prophesee} & \multicolumn{3}{c|}{Samsung} & \multicolumn{2}{c|}{CelePixel} & Insightness\tabularnewline[0.4ex]
 \multicolumn{2}{l|}{Camera model} & DVS128 & DAVIS240 & DAVIS346%
 & ATIS & Gen3 CD & Gen3 ATIS & Gen 4 CD & DVS-Gen2 & DVS-Gen3 & DVS-Gen4 & CeleX-IV & CeleX-V & Rino 3\tabularnewline
\midrule 
 & Year, Reference & 2008~\cite{Lichtsteiner08ssc} & 2014~\cite{Brandli14ssc} & 2017 & 2011~\cite{Posch11ssc} & 2017~\cite{propheseeevk} & 2017~\cite{propheseeevk} & 2020 \cite{Finateu20isscc} & 2017~\cite{Son17isscc} & 2018~\cite{Ryu19cvprw} & 2020~\cite{Suh20iscas} & 2017~\cite{Guo17iscas} & 2019~\cite{Chen19cvprw} & 2018~\cite{Insightness19rino}\tabularnewline
\multirow{16}{*}{\rotatebox[origin=c]{90}{Sensor specifications}} & Resolution (\si{pixels}) & $128\times128$ & $240\times180$ & $346\times260$ & $304\times240$ & $640\times480$ & $480\times360$ & $1280\times720$  & $640\times480$ & $640\times480$ & $1280\times960$ & $768\times640$ & $1280\times800$ & $320\times262$\tabularnewline
 & %
Latency (\si{\micro\second}) & 12$\mu$s @ 1klux & 12$\mu$s @ 1klux & 20 & 3 & 40 - 200 & 40 - 200 & 20 - 150 & 65 - 410 & 50 & 150 & 10 & 8 & 125$\mu$s @ 10lux\tabularnewline
 & Dynamic range (\si{\decibel}) & 120 & 120 & 120 & 143 & $>120$ & $>120$ & $>124$ & 90 & 90 & 100 & 90 & 120 & $>100$\tabularnewline
 & Min. contrast sensitivity (\si{\percent}) & 17 & 11 & 14.3 - 22.5 & 13 & 12 & 12 & 11 & 9 & 15 & 20 & 30 & 10 & 15\tabularnewline
 & Power consumption (\si{\milli\watt}) & 23 & 5 - 14 & 10 - 170 & 50 - 175 & 36 - 95 & 25 - 87 & 32 - 84 & 27 - 50 & 40 & 130 & - & 400 & 20-70\tabularnewline
 & Chip size (\si{\milli\meter\squared}) & 6.3 $\times$ 6 & 5 $\times$ 5 & 8 $\times$ 6 & 9.9 $\times$ 8.2 & 9.6 $\times$ 7.2 & 9.6 $\times$ 7.2 & 6.22 $\times$ 3.5  & 8 $\times$ 5.8 & 8 $\times$ 5.8 & 8.4 $\times$ 7.6 & 15.5 $\times$ 15.8 & 14.3 $\times$ 11.6 & 5.3 $\times$ 5.3\tabularnewline
 & Pixel size (\si{\micro\meter\squared}) & 40 $\times$ 40 & 18.5 $\times$ 18.5 & 18.5 $\times$ 18.5 & 30 $\times$ 30 & 15 $\times$ 15 & 20 $\times$ 20 & 4.86 $\times$ 4.86 & 9 $\times$ 9 & 9 $\times$ 9 & 4.95 $\times$ 4.95 & 18 $\times$ 18 & 9.8 $\times$ 9.8 & 13 $\times$ 13\tabularnewline
 & Fill factor (\si{\percent}) & 8.1 & 22 & 22 & 20 & 25 & 20 & $>77$ & 11 & 12 & 22 & 8.5 & 8 & 22\tabularnewline
 & Supply voltage (\si{\volt}) & 3.3 & 1.8 \& 3.3 & 1.8 \& 3.3 & 1.8 \& 3.3 & 1.8 & 1.8 & 1.1 \& 2.5 & 1.2 \& 2.8 & 1.2 \& 2.8 &  & 1.8 \& 3.3 & 1.2 \& 2.5 & 1.8 \& 3.3\tabularnewline
 & Stationary noise (ev/pix/s) at 25C & 0.05 & 0.1 & 0.1 & - & 0.1 & 0.1 & 0.1 & 0.03 & 0.03 &  & 0.15 & 0.2 & 0.1\tabularnewline
 & CMOS technology (\si{\nano\meter}) & 350 & 180 & 180 & 180 & 180 & 180 & 90 & 90 & 90 & 65/28 & 180 & 65 & 180\tabularnewline
 &  & 2P4M & 1P6M MIM & 1P6M MIM & 1P6M & 1P6M CIS & 1P6M CIS & BI CIS & 1P5M BSI &  &  & 1P6M CIS & CIS & 1P6M CIS\tabularnewline
 \cmidrule{2-15}
 & Grayscale output & no & yes & yes & yes & no & yes & no & no & no & no & yes & yes & yes\tabularnewline
 & Grayscale dynamic range (\si{\decibel}) & NA & 55 & 56.7 & 130 & NA & $>100$ & NA & NA & NA & NA & 90 & 120 & 50\tabularnewline
 & Max. frame rate (fps) & NA & 35 & 40 & NA & NA & NA & NA & NA & NA & NA & 50 & 100 & 30\tabularnewline
\midrule 
\multirow{2}{*}{\rotatebox[origin=c]{90}{Camera}} 
 & Max. Bandwidth (\si{\mega eps}) & 1 & 12 & 12 & - & 66 & 66 & 1066 & 300 & 600 & 1200 & 200 & 140 & 20\tabularnewline
 & Interface & USB 2 & USB 2 & USB 3 &  & USB 3 & USB 3 & USB 3 & USB 2 & USB 3 & USB 3 &  &  & USB 2 \tabularnewline
 & IMU output & no & \SI{1}{\kilo\Hz} & \SI{1}{\kilo\Hz} & no & \SI{1}{\kilo\Hz} & \SI{1}{\kilo\Hz} & no & no & \SI{1}{\kilo\Hz} & no & no & no  & \SI{1}{\kilo\Hz}\tabularnewline
\bottomrule
\end{tabular}
\end{adjustbox}
\vspace{-1ex}
\end{table*}

%% file: chapters/025_camera_devices.tex
\subsection{Event Camera Availability}
\label{sec:physical_devices}

Table~\ref{tab:devices} summarizes the most popular or recent cameras.
The numbers therein are approximate since they were not measured using a common testbed.
Event camera characteristics are considerably different from other CMOS image sensor (\textbf{CIS}) technology, 
and so there is a need for an agreement on standard specifications to be better used by researchers.
As Table~\ref{tab:devices} shows, since the first practical event camera~\cite{Lichtsteiner08ssc} there has been a trend mainly to increase spatial resolution, increase readout speed, and add features, such as:
gray level output (in ATIS and DAVIS), integration with an Inertial Measurement Unit (\textbf{IMU})~\cite{Delbruck14iscas} and multi-camera timestamp synchronization~\cite{Berner06thesis}.
IMUs act as a vestibular sense that may improve camera pose estimation, as in visual-inertial odometry.
Only recently has the focus turned more towards the difficult task of reducing pixel size for economical mass production of sensors with large pixel arrays.
In this respect, 3D wafer stacking fabrication has the biggest impact in reducing pixel size and increasing the fill factor.

\textbf{Pixel Size}:
The most widely used event cameras have quite large pixels:
\SI{40}{\micro\meter} (DVS128), \SI{30}{\micro\meter} (ATIS), \SI{18.5}{\micro\meter} (DAVIS240, DAVIS346) (Table~\ref{tab:devices}).
The smallest published DVS pixel~\cite{Finateu20isscc} is \SI{4.86}{\micro\meter};
while conventional global shutter industrial APS are typically in the range of \SIrange{2}{4}{\micro\meter}.
Low spatial resolution is certainly a limitation for application, although many of the seminal publications are based on the~$128 \times 128$ pixel DVS128~\cite{Lichtsteiner08ssc}.
The DVS with largest published array size has only about \SI{1}{\mega pixel} spatial resolution ($1280 \times 960$ pixels~\cite{Suh20iscas}).
Event camera pixel size has shrunk pretty closely following feature size scaling, which is remarkable considering that a DVS pixel is a mixed-signal circuit, which generally do not scale following technology.
However, achieving even smaller pixels is difficult and may require abandoning the strictly asynchronous circuit design philosophy that the cameras started with~\cite{Inivation20whitepaper}.
Camera cost is constrained by die size (since silicon costs about \$5-\$10\si{/\centi\meter^2} in mass production), and optics (designing new mass production miniaturized optics to fit a different sensor format can cost tens of millions of dollars).

\textbf{Fill Factor}:
A major obstacle for early event camera mass production prospects was the limited fill factor of the pixels (i.e., the ratio of a pixel's light sensitive area to its total area).
Because the pixel circuit is complex, a smaller pixel area can be used for the photodiode that collects light. 
For example, a pixel with \SI{20}{\percent} fill factor throws away 4 out of 5 photons.
Obviously this is not acceptable for optimum performance;
nonetheless, even the earliest event cameras could sense high contrast features under moonlight illumination~\cite{Lichtsteiner08ssc}.
Early CIS sensors dealt with this problem by including microlenses that focused the light onto the pixel photodiode.
What is probably better, however, is to use back-side illumination technology (\textbf{BSI}).
BSI flips the chip so that it is illuminated from the back, so that in principle the entire pixel area can collect photons.
Nearly all smartphone cameras are now back illuminated,
but the additional cost %
of BSI fabrication has meant that only recently BSI event cameras were demonstrated~\cite{Taverni18tcsii,Ryu19cvprw,Finateu20isscc,Suh20iscas}.
BSI also brings problems: light can create additional `parasitic' photocurrents that lead to spurious `leak' events~\cite{Nozaki17ted}.

\textbf{Cost}:
Currently, a practical obstacle to adoption of event camera technology is the high cost of several thousand dollars per camera, similar to the situation with early time of flight, structured lighting and thermal cameras.
The high costs are due to non-recurring engineering costs for the silicon design and fabrication 
(even when much of it is provided by research funding) and the limited samples available from prototype runs.
It is anticipated that this price will drop precipitously once this technology enters mass production, 
as shown by the ``Samsung SmartThings Vision'' consumer-grade home monitoring device: it contains an event camera~\cite{Son17isscc} and sells for 100 dollars.

\iflongversion
\subsubsection*{Advanced Event Cameras}
There are active developments of more advanced event cameras that are only available through scientific collaborations with the developers. 
Next, we discuss issues related to advanced camera developments and the types of new cameras that are being developed.

\textbf{Color}:
Most diurnal animals have some form of color vision, and most conventional cameras offer color sensitivity.
Early attempts at color sensitive event cameras~\cite{Fasnacht07iscas,Berner08iscas,Farian15tbcas} tried to use the ``vertacolor'' principle of splitting colors according to the amount of penetration of the different light wavelengths into silicon,
pioneered by Foveon~\cite{Merrill99patent,Lyon02ist}.
However, it resulted in poor color separation performance.
So far, there are few publications of practical color event cameras,
with either integrated color filter arrays (\textbf{CFA})~\cite{Li15iscas,Moeys17iscas,Scheerlinck19cvprw} 
or color-splitter prisms~\cite{Marcireau18fns};
splitters have a much higher cost than CFA.

\textbf{Higher Contrast Sensitivity}:
Efforts have been made to improve the temporal contrast sensitivity of event cameras,
leading to experimental sensors with higher sensitivity~\cite{Serrano-Gotarredona13ijssc,Yang15ssc,Moeys17tbcas} (down to laboratory condition $\sim\SI{1}{\percent}$).
These sensors are based on variations of the idea of a thermal bolometer~\cite{Posch09jsen},
i.e., increasing the gain before the change detector (Fig.~\ref{fig:DVSsummary}) to reduce the input-referred FPN.
However this intermediate preamplifier requires active gain control to avoid clipping.
Increasing the contrast sensitivity is possible, at the expense of decreasing the dynamic range (e.g.,~\cite{Son17isscc}).
\fi

%% file: chapters/03_event_processing.tex
\section{Event Processing}
\label{sec:event_processing}
\input{chapters/030_processing_overview.tex}

\subsection{Methods for Event Processing}
Event processing systems consist of several stages: pre-processing (input adaptation), core processing (feature extraction and analysis) and post-processing (output creation).
The event representations in Section~\ref{sec:representations} may occur at different stages: 
for example, in~\cite{Gallego17ral} an event packet is used at pre-processing, 
and motion-compensated event images are the internal representation at the core processing stage.
\iflongversion In other cases, the above representations may be used only at pre-processing: 
in \cite{Zhu18rss} events are converted to event images and time surfaces that are then processed by an ANN.
\fi

The methods used to process events are influenced by the choice of representation and hardware platform available. 
These three factors influence each other.
For example, it is natural to use dense representations and design algorithms accordingly that are executed on standard processors (e.g., CPUs or GPUs).
At the same time, it is also natural to process events one-by-one on SNNs (Section~\ref{sec:bioinspired_processing}) that are implemented on neuromorphic hardware (Section~\ref{sec:hardware}), 
in search for more efficient and low-latency solutions.
Major exponents of event-by-event methods are filters (deterministic or probabilistic) and SNNs.
For events processed in packets there are also many methods: hand-crafted feature extractors, deep neural networks (\textbf{DNNs}), etc.
Next, we review some of the most common methods.

\input{chapters/031_processing_event_by_event.tex}

\input{chapters/032_processing_groups.tex}

\input{chapters/033_processing_bioinspired.tex}

%% file: chapters/030_processing_overview.tex
\input{chapters/fig_representatiaons.tex}

One of the key questions of the paradigm shift posed by event cameras is how to extract meaningful information from the event data to fulfill a given task.
This is a very broad question, since the answer is application dependent, and it drives the algorithmic design of the task solver.

Event cameras acquire information in an asynchronous and sparse way, with high temporal resolution and low latency. 
Hence, the temporal aspect, specially latency, plays an essential role in the way events are processed.
Depending on how many events are processed simultaneously, two categories of algorithms can be distinguished:
(\emph{i}) methods that operate on an \emph{event-by-event basis},
where the state of the system (the estimated unknowns) can change upon the arrival of a single event,
thus achieving minimum latency,
and (\emph{ii}) methods that operate on \emph{groups or packets of events}, which introduce some latency.
Discounting latency considerations, methods based on groups (i.e., temporal windows) of events can still provide a state update upon the arrival of each event if the window slides by one event.
Hence, the distinction between both categories is more subtle:
an event alone does not provide enough information for estimation, and so additional information, in the form of past events or extra knowledge, is needed.
We review this categorization.

Orthogonally, depending on how events are processed, we can distinguish between model-based approaches and model-free (i.e., data-driven, machine learning) approaches.
Assuming events are processed in an optimization framework, another classification concerns the type of objective or loss function used: geometric- vs. temporal- vs. 
photometric-based (e.g., a function of the event polarity or the event activity).
Each category presents methods with advantages and disadvantages
and current research focuses on exploring the possibilities that each method can offer.

\input{chapters/030_representations.tex}

%% file: chapters/fig_representatiaons.tex
\iffalse
\begin{figure}[t]
    \centering
    \includegraphics[width=0.8\linewidth]{images/representations/event_tensors_crop.png}
    \vspace{-2ex}
    \caption{Different ways to convert events into more familiar representations, suitable for processing with modern learning architectures~\cite{Gehrig19iccv}.}
    \label{fig:representations}
\end{figure}
\fi

\global\long\def\represheight{2.19cm}
\begin{figure*}[t]
    \centering
    \subfloat[]{\includegraphics[trim={1.9cm 1cm 2.2cm 2cm},clip,height=\represheight]{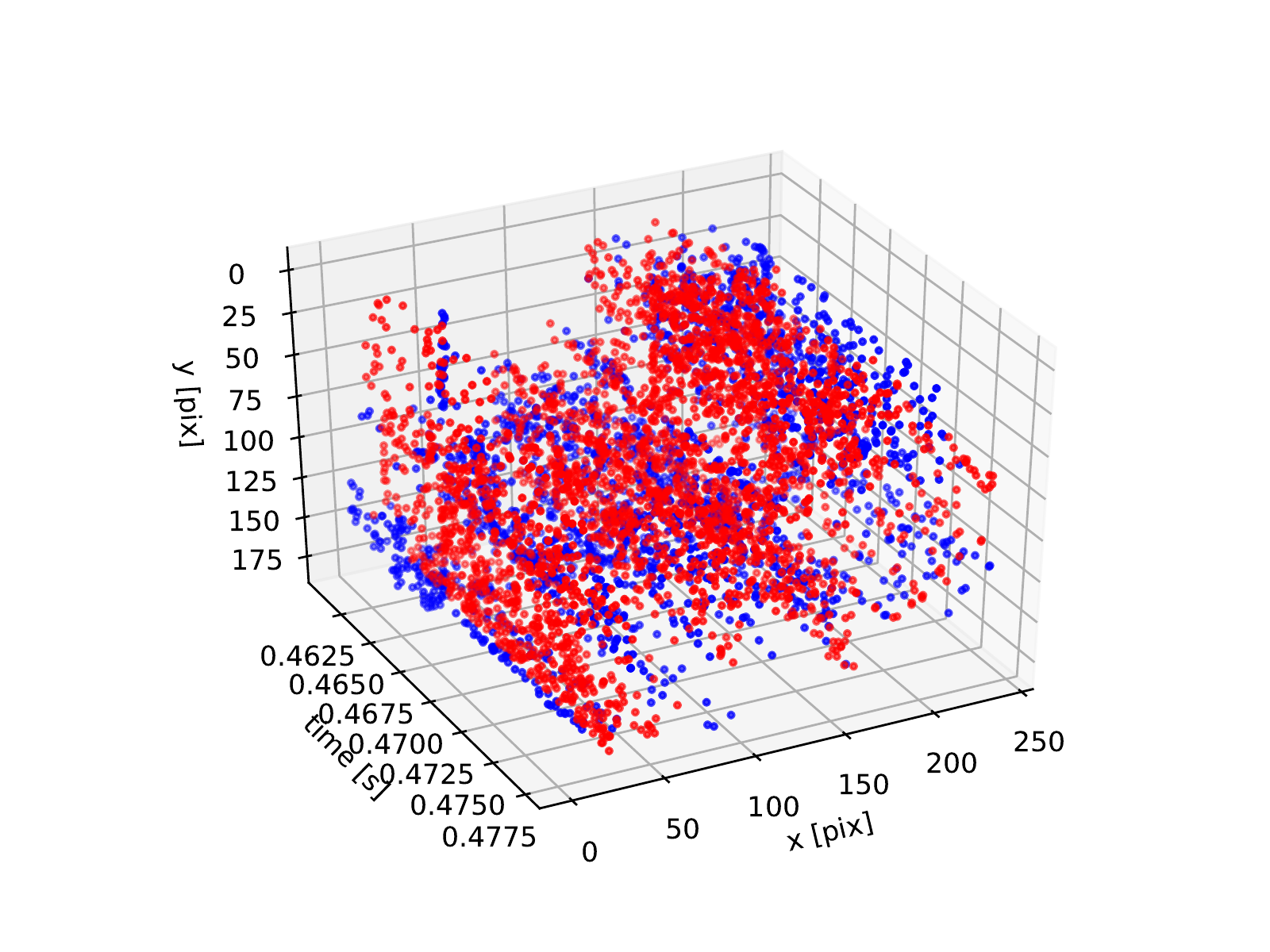}}
    \subfloat[]{\frame{\includegraphics[height=\represheight]{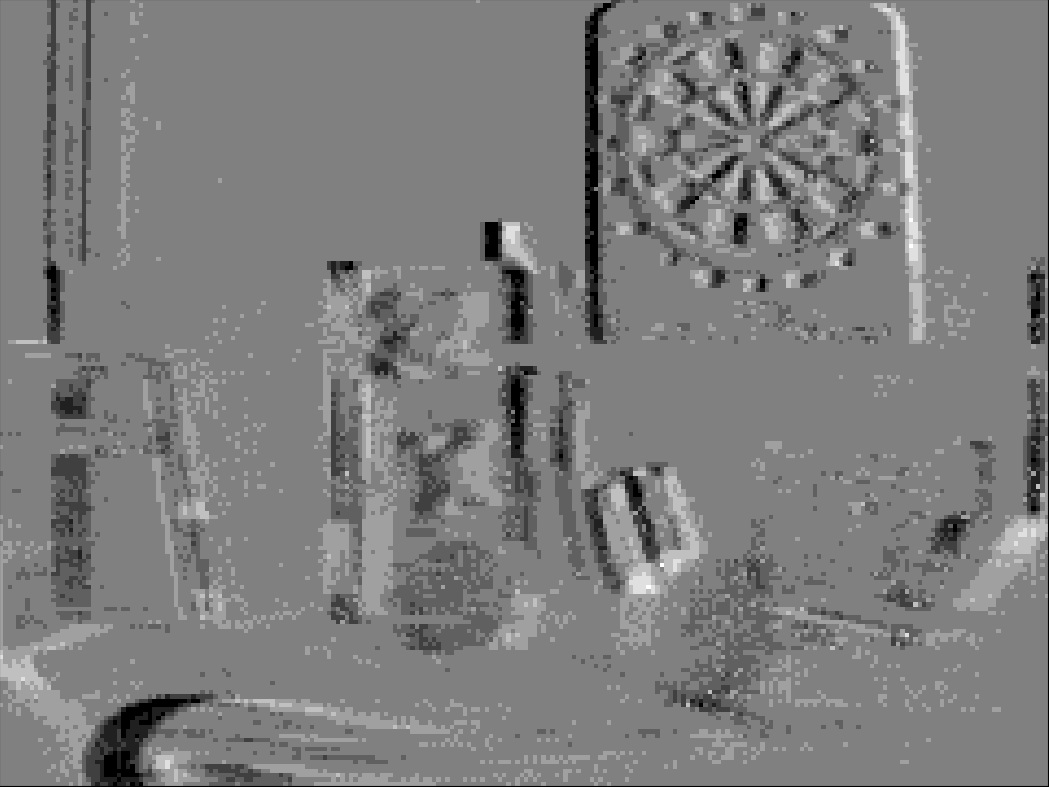}}}\,
    \subfloat[]{\frame{\includegraphics[height=\represheight]{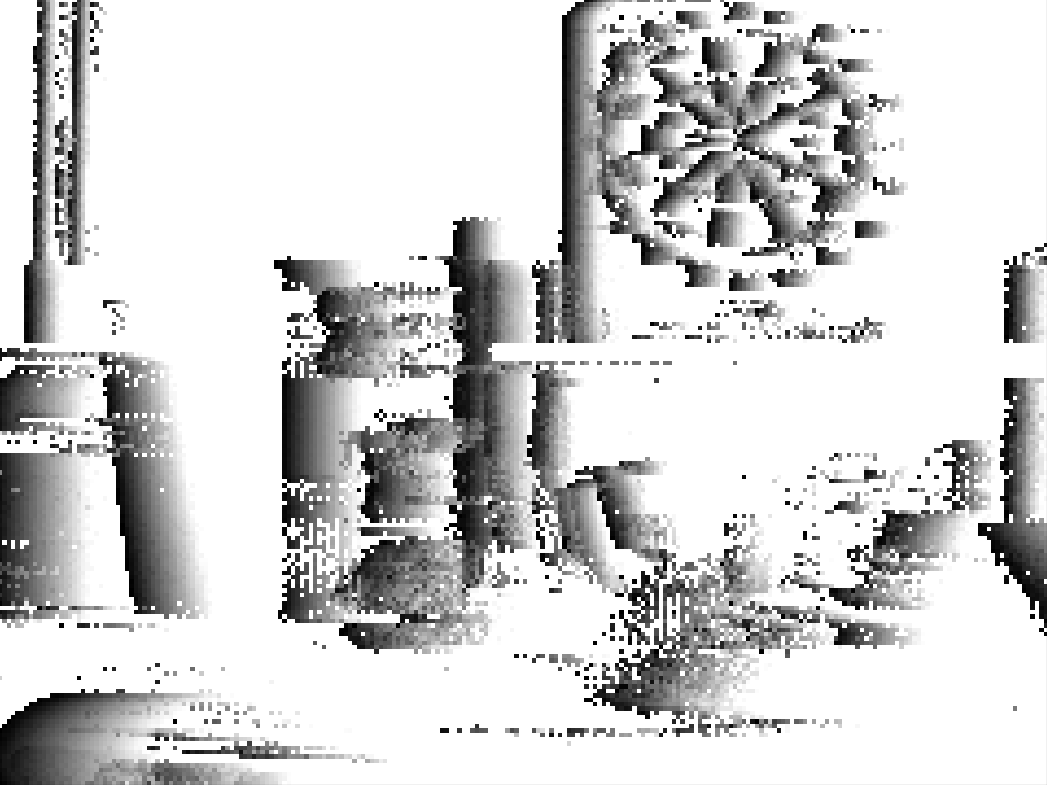}}}
    \subfloat[]{\includegraphics[trim={1.9cm 1cm 2.2cm 2cm},clip,height=\represheight]{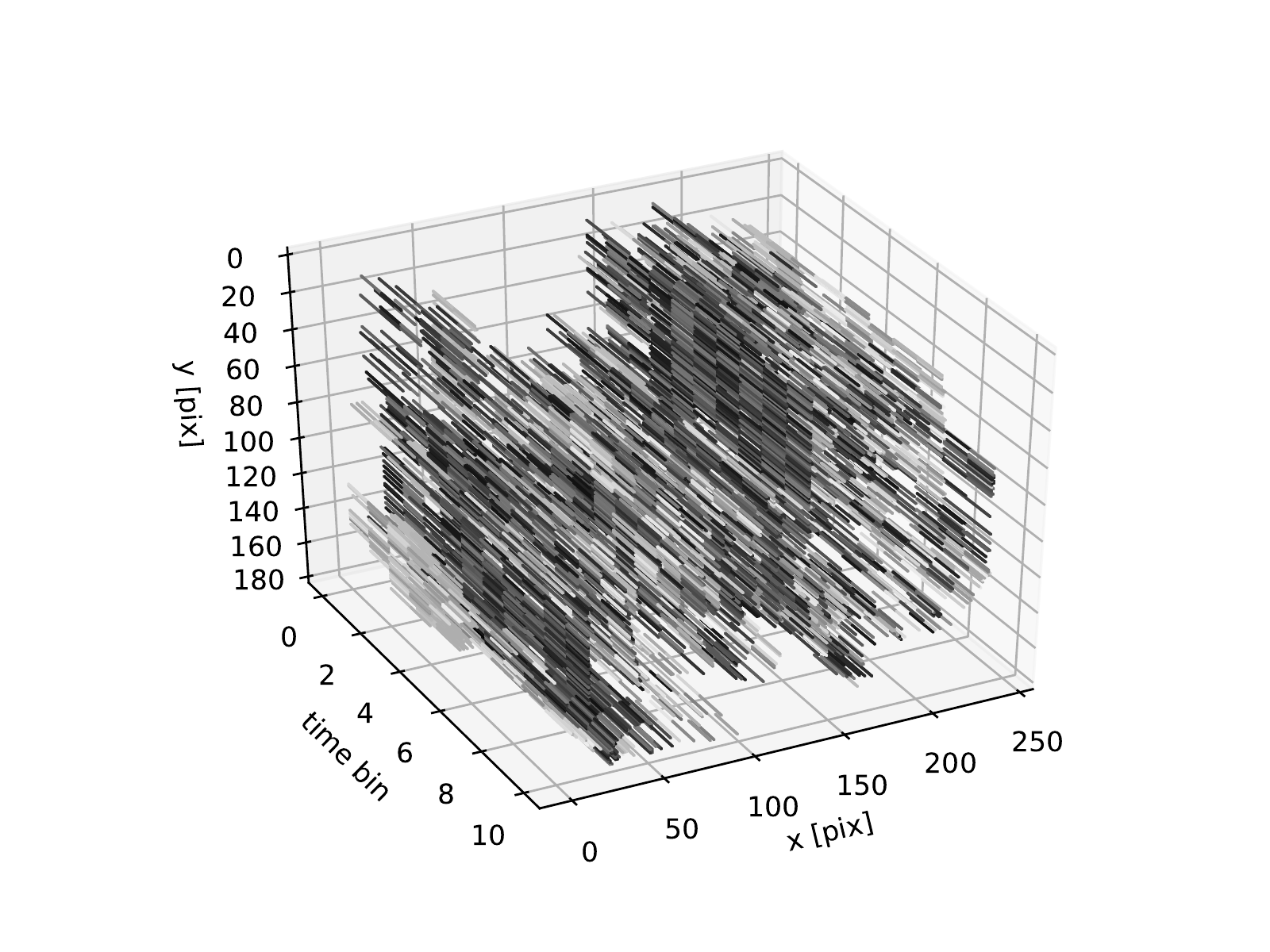}}
    \subfloat[]{\frame{\includegraphics[height=\represheight]{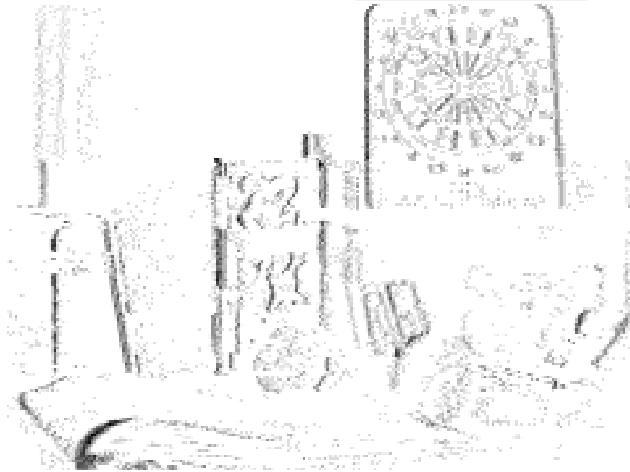}}}\,
    \subfloat[]{\frame{\includegraphics[height=\represheight]{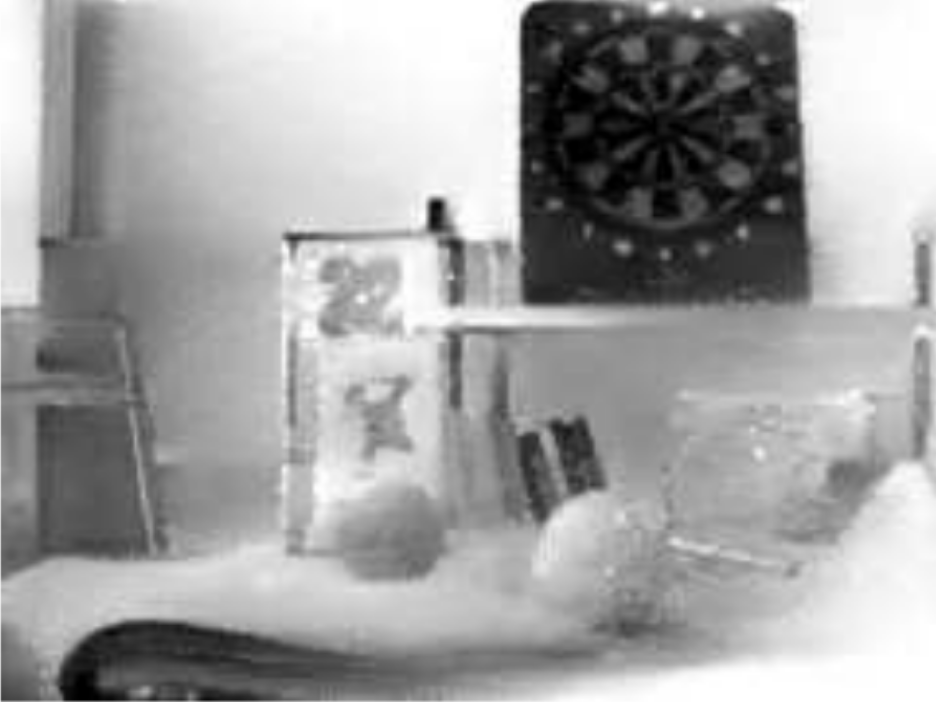}}}
    \vspace{-1ex}
    \caption{Several event representations (Section~\ref{sec:representations}) of the \emph{slider\_depth} sequence~\cite{Mueggler17ijrr}. 
    (\textbf{a}) Events in space time, colored according to polarity (positive in blue, negative in red).
    (\textbf{b}) Event frame (brightness increment image $\Delta \Lum(\bx)$). 
    (\textbf{c}) Time surface with last timestamp per pixel (darker pixels indicate recent time), only for negative events.
    (\textbf{d}) Interpolated voxel-grid ($240\times 180\times 10$ voxels), colored according to polarity, from dark (negative) to bright (positive).
    (\textbf{e}) Motion-compensated event image~\cite{Gallego19cvpr} (sharp edges obtained by event accumulation are darker than pixels with no events, in white).
    (\textbf{f}) Reconstructed intensity image by~\cite{Rebecq19pami}.
	Grid-like representations are compatible with conventional computer vision methods~\cite{Gehrig19iccv}.}
    \label{fig:representations}
    \vspace{-1ex}
\end{figure*}

%% file: chapters/030_representations.tex
\subsection{Event Representations}
\label{sec:representations}

Events are processed and often transformed into alternative representations (Fig.~\ref{fig:representations}) that facilitate the extraction of meaningful information (``features'') to solve a given task.
Here we review popular representations of event data.
Several of them arise from the need to aggregate the little information conveyed by individual events in the absence of additional knowledge.
Some representations are simple, hand-crafted data transformations whereas others are more elaborate.

\textbf{Individual events} $e_k \doteq (\bx_k,t_k,\pol_k)$ are used by event-by-event processing methods, such as probabilistic filters and Spiking Neural Networks (\textbf{SNNs}) (Section~\ref{sec:bioinspired_processing}).
The filter or SNN has additional information, built up from past events or given by additional knowledge, 
that is fused with the incoming event asynchronously to produce an output.
Examples include: \cite{Weikersdorfer12robio,Kim14bmvc,Gallego17pami,Scheerlinck18accv,Paredes19pami}.

\textbf{Event packet}: Events $\mathcal{E}\doteq\{e_k\}_{k=1}^{N_e}$ in a spatio-temporal neighborhood are processed together to produce an output.
Precise timestamp and polarity information is retained by this representation.
Choosing the appropriate packet size $N_e$ is critical to satisfy the assumptions of the algorithm (e.g., constant motion speed during the span of the packet), which varies with the task.
Examples are \cite{Rogister12tnnls,Reinbacher17iccp,Rebecq18ijcv,Mueggler18tro}.

\textbf{Event frame/image or 2D histogram}:
The events in a spatio-temporal neighborhood are converted in a simple way (e.g., by counting events or accumulating polarity pixel-wise) into an image (2D grid) that can be fed to image-based computer vision algorithms.
Some algorithms may work in spite of the different statistics of event frames and natural images.
Such histograms can provide a natural activity-driven sample rate; see ~\cite{Liu18bmvc} for methods to accumulate such frames for computing flow.
However, this practice is not ideal in the event-based paradigm because it quantizes event timestamps, can discard sparsity (but see~\cite{Aimar19tnnls}), and the resulting images are highly sensitive to the number of events used.
Nevertheless the high impact of event frames in the literature \cite{Kogler09icvs,Cook11ijcnn,Liu18bmvc,Rebecq17ral,Maqueda18cvpr,Gehrig19ijcv} is clear because
(\emph{i}) they are a simple way to convert an unfamiliar event stream into a familiar 2D representation containing spatial information about scene edges, which are the most informative regions in natural images, 
(\emph{ii}) they inform not only about the presence of events but also about their absence (which is informative), 
(\emph{iii}) they have an intuitive interpretation (e.g., an edge map, a brightness increment image)
and 
(\emph{iv}) they are the data structure compatible with conventional computer vision.

\textbf{Time surface (TS)}: A TS is a 2D map where each pixel stores a single time value (e.g., the timestamp of the last event at that pixel~\cite{Delbruck08issle,Lagorce17pami}).
Thus events are converted into an image whose ``intensity'' is a function of the motion history at that location, %
with larger values corresponding to a more recent motion.
TSs are called Motion History Images in classical computer vision~\cite{Ahad12mva}.
They explicitly expose the rich temporal information of the events and can be updated asynchronously.
Using an exponential kernel, TSs emphasize recent events over past events.
To achieve invariance to motion speed, normalization is proposed~\cite{Alzugaray18ral,Manderscheid19cvpr}.
Compared to other grid-like representations of events, TSs highly compress information as they only keep one timestamp per pixel, thus their effectiveness degrades on textured scenes, in which pixels spike frequently.
To make TSs less sensitive to noise, each pixel value may be computed by filtering the events in a space-time window~\cite{Sironi18cvpr}.
More examples include \cite{Benosman14tnnls,Vasco16iros,Mueggler17bmvc,Zhou18eccv}.

\textbf{Voxel Grid}: is a space-time (3D) histogram of events, where each voxel represents a particular pixel and time interval.
This representation preserves better the temporal information of the events by avoiding to collapse them on a 2D grid (Fig.~\ref{fig:representations}).
If polarity is used the voxel grid is an intuitive discretization of a scalar field (polarity $p(x,y,t)$ or brightness variation $\partial\Lum(x,y,t)/\partial t$) defined on the image plane, with absence of events marked by zero polarity.
Each event's polarity may be accumulated on a voxel~\cite{Bardow16cvpr,Mostafavi19cvpr} or 
spread among its closest voxels using a kernel \cite{Zhu19cvpr,Rebecq19cvpr,Rebecq19pami}.
Both schemes quantize event timestamps but the latter (interpolated voxel grid) provides sub-voxel accuracy.

\textbf{3D point set}: Events in a spatio-temporal neighborhood are treated as points in 3D space, $(x_k,y_k,t_k)\!\in\! \mathbb{R}^3$. 
Thus the temporal dimension becomes a geometric one.
It is a sparse representation, and is used on point-based geometric processing methods, such as plane fitting~\cite{Benosman14tnnls} or PointNet~\cite{Sekikawa19cvpr}. %

\textbf{Point sets on image plane}: Events are treated as an evolving set of 2D points on the image plane.
It is a popular representation among early shape tracking methods based on mean-shift or ICP \cite{Litzenberger06dspws,Ni12tro,Ni15neco,Tedaldi16ebccsp,Kueng16iros}, where events provide the only data needed to track edge patterns. %

\textbf{Motion-compensated event image} \cite{Gallego17ral,Gallego18cvpr}: is a representation that depends not only on events but also on motion hypothesis.
The idea of motion compensation is that, as an edge moves on the image plane, it triggers events on the pixels it traverses; the motion of the edge can be estimated by warping the events to a reference time and maximizing their alignment, producing a sharp image (i.e., histogram) of warped events (IWE) \cite{Gallego18cvpr}.
Hence, this representation (IWE) suggests a criterion to measure how well events fit a candidate motion:
the sharper the edges produced by warping events, the better the fit~\cite{Gallego19cvpr}.
Moreover, the resulting motion-compensated images have an intuitive meaning (i.e., the edge patterns causing the events)
and provide a more familiar representation of visual information than the events.
In a sense, motion compensation reveals a hidden (``motion-invariant'') map of edges in the event stream.
The images may be useful for further processing, such as feature tracking \cite{Rebecq17bmvc,Gehrig19ijcv}.
There are motion-compensated versions of point sets~\cite{Zhu17icra,Zhu17cvpr} and time surfaces~\cite{Mitrokhin18iros,Mitrokhin19iros}.

\textbf{Reconstructed images}: Brightness images obtained by image reconstruction (Section~\ref{sec:imagereconstruction}) can be interpreted as a more motion-invariant representation than event frames or TSs, and be used for inference~\cite{Rebecq19pami} yielding first-rate results.

\iflongversion
\emph{Event polarity} may be considered in two ways:
processing positive and negative events separately and merging results (e.g., using TSs \cite{Lagorce17pami}),
or processing them together in a common representation (e.g., brightness increments images~\cite{Gehrig19ijcv}), where polarity is often aggregated among neighboring events.
Event polarity depends on motion direction, hence it is a nuisance for tasks that should be independent of motion, such as object recognition 
(to mitigate this, training data from multiple motion directions should be available).
For motion estimation tasks, polarity may be useful, especially to detect abrupt changes of direction.
\fi

A general framework for converting event data into some of the above grid-based representations is presented in~\cite{Gehrig19iccv}. 
It also studies how the choice of representation passed to an artificial neural network (\textbf{ANN}) affects task performance and consequently proposes to automatically learn the representation that maximizes such performance.

%% file: chapters/031_processing_event_by_event.tex
\label{sec:event_by_event}

\textbf{Event-by-event--based Methods}:
Deterministic filters, such as (space-time) convolutions and activity filters %
have been used for noise reduction, feature extraction~\cite{Brosch15fns}, image reconstruction~\cite{Reinbacher16bmvc,Scheerlinck18accv} and brightness filtering \cite{Scheerlinck19ral}, among other applications.
Probabilistic filters (Bayesian methods), such as Kalman- and particle filters have been used for pose tracking in SLAM systems~\cite{Weikersdorfer12robio,Censi14icra,Kim14bmvc,Kim16eccv,Gallego17pami}.
These methods rely on the availability of additional information (typically ``appearance'' information, e.g., grayscale images or a map of the scene), which may be provided by past events or by additional sensors.
Then, each incoming event is compared against such information and the resulting mismatch provides innovation to update the filter state.
Filters are a dominant class of methods for event-by-event processing because they naturally (\emph{i}) handle asynchronous data, thus providing minimum processing latency, preserving the sensor's characteristics,
and (\emph{ii}) aggregate information from multiple small sources (e.g., events).

The other dominant class of methods takes the form of a multi-layer ANN (whether spiking or not) containing many parameters which must be computed from the event data.
Networks trained with unsupervised learning typically act as feature extractors for a classifier (e.g., SVM), which still requires some labeled data for training \cite{Orchard15pami,Akolkar15icra,Lagorce17pami}.
If enough labeled data is available, supervised learning methods such as backpropagation can be used to train a network without the need for a separate classifier. 
Many approaches use packets of events during training (deep learning on frames), and later convert the trained network to an SNN that processes data event-by-event \cite{PerezCarrasco13pami,OConnor13fns,Diehl15ijcnn,Esser16pnas,Rueckauer17fns}.
Event-by-event model-free methods have mostly been applied to classify objects~\cite{PerezCarrasco13pami,OConnor13fns,Orchard15pami,Lagorce17pami} or actions~\cite{Lee14tnnls,Amir17cvpr,Shrestha18nips}, and have targeted embedded applications~\cite{PerezCarrasco13pami}, often using custom SNN hardware~\cite{Orchard15pami,Amir17cvpr} (Section~\ref{sec:hardware}).
SNNs trained with deep learning typically provide higher accuracy than those relying on unsupervised learning for feature extraction, but there is growing interest in finding efficient ways to implement supervised learning directly in SNNs~\cite{Lee16fns,Shrestha18nips} and in embedded devices~\cite{neftci18iscience}.

%% file: chapters/032_processing_groups.tex
\label{sec:groups_events}
\textbf{Methods for Groups of Events}:
Because each event carries little information and is subject to noise, 
several events are often processed together to yield a sufficient signal-to-noise ratio for the problem considered.
Methods for groups of events use the above representations (event packet, event frame, etc.) to gather the information contained in the events in order to estimate the problem unknowns, usually without requiring additional data.
Hence, events are processed differently depending on their representation.

Many representations just perform data pre-processing to enable the re-utilization of image-based computer vision tools.
In this respect, \emph{event frames} are a practical representation that has been used by multiple methods on various tasks.
In~\cite{Kogler09icvs,Kogler11book} event frames allow to re-utilize traditional stereo methods, providing modest results.
They also provide an adaptive frame rate signal that is profitable for camera pose estimation \cite{Rebecq17ral} (by image alignment) or optical flow computation \cite{Liu18bmvc} (by block matching).
Event frames are also a simple yet effective input for image-based learning methods (DNNs, SVMs, Random Forests) \cite{Li16bics,Maqueda18cvpr,Nguyen19cvprw,Zhu18rss}.
Few works design algorithms taking into account their photometric meaning \eqref{eq:brightnessIncrementLinearized}.
This was done in \cite{Cook11ijcnn}, showing that such a simple representation allows to jointly compute several visual quantities of interest (optical flow, brightness, etc.).
Intensity increment images~\eqref{eq:brightnessIncrementLinearized} are also used for feature tracking~\cite{Gehrig19ijcv}, image deblurring~\cite{Pan19cvpr} or camera tracking~\cite{Bryner19icra}.

Because \emph{time surfaces} (TSs) are sensitive to scene edges and the direction of motion they have been utilized for many tasks involving motion analysis and shape recognition.
For example, fitting local planes to the TS yields optical flow information~\cite{Benosman14tnnls,Mueggler15icra}.
TSs are used as building blocks of hierarchical feature extractors, similar to neural networks, 
that aggregate information from successively larger space-time neighborhoods 
and is then passed to a classifier for recognition \cite{Lagorce17pami,Sironi18cvpr}.
TSs provide proxy intensity images for matching in stereo methods \cite{Ieng18fnins,Zhou18eccv}, where the photometric matching criterion becomes temporal: matching pixels based on event concurrence and similarity of event timestamps across image planes.
Recently, TSs have been probed as input to convolutional ANNs (\textbf{CNNs}) to compute optical flow~\cite{Zhu18rss}, where the network acts both as feature extractor and velocity regressor.
TSs are popular for corner detection using adaptations of image-based methods (Harris, FAST) \cite{Vasco16iros,Mueggler17bmvc,Alzugaray18ral} or new learning-based ones~\cite{Manderscheid19cvpr}.
However, their performance degrades on highly textured scenes~\cite{Mueggler17bmvc} due to the ``motion overwriting'' problem~\cite{Ahad12mva}.

Methods working on \emph{voxel grids} include variational optimization and ANNs (e.g., DNNs).
They require more memory and often more computations than methods working on lower dimensional representations but are able to provide better results because temporal information is better preserved.
In these methods voxel grids are used as an internal representation~\cite{Bardow16cvpr} (e.g., to compute optical flow)
or as the multichannel input/output of a DNN \cite{Zhu19cvpr,Rebecq19cvpr}.
Thus, voxel grids are processed by means of convolutions \cite{Zhu19cvpr,Rebecq19cvpr} or the operations derived from the optimality conditions of an objective function~\cite{Bardow16cvpr}.

Once events have been converted to grid-like representations, countless tools from conventional vision can be applied to extract information: from feature extractors (e.g., CNNs) to similarity metrics (e.g., cross-correlation) that measure the goodness of fit or consistency between data and task-model hypothesis (the degree of event alignment, etc.).
Such metrics are used as objective functions for classification (SVMs, CNNs), clustering, data association, motion estimation, etc.
In the neuroscience literature there are efforts to design metrics that act directly on spikes (e.g., event stream), to avoid the issues that arise due to data conversion.

\emph{Deep learning methods} for groups of events consist of a deep neural network (DNN).
Sample applications include classification~\cite{Moeys16ebccsp,Lungu17iscas}, 
image reconstruction~\cite{Rebecq19pami,Mostafavi19cvpr}, 
steering angle prediction~\cite{Binas17icml,Maqueda18cvpr}, 
and estimation of optical flow~\cite{Zhu18rss,Ye18arxiv,Zhu19cvpr}, depth \cite{Ye18arxiv} or ego-motion~\cite{Zhu19cvpr}.
These methods differentiate themselves mainly in the representation of the input and in the loss functions optimized during training.
Several representations have been used, such as event images~\cite{Maqueda18cvpr,Nguyen19cvprw},
TSs~\cite{Zhu18rss,Ye18arxiv,Mitrokhin19iros}, voxel grids \cite{Zhu19cvpr,Rebecq19cvpr} 
or point sets~\cite{Sekikawa19cvpr} (Section~\ref{sec:representations}).
While loss functions in classification tasks use manually annotated labels,
networks for regression tasks from events may be supervised by a third party ground truth (e.g., a pose)~\cite{Maqueda18cvpr,Nguyen19cvprw} or by an associated grayscale image~\cite{Zhu18rss} to measure photoconsistency, or be completely unsupervised (depending only on the training input events)~\cite{Zhu19cvpr,Ye18arxiv}. 
Loss functions for unsupervised learning from events are studied in~\cite{Gallego19cvpr}.
In terms of architecture, most networks have an encoder-decoder structure, as in Fig.~\ref{fig:architecture}. 
Such a structure allows the use of convolutions only, thus minimizing the number of network weights. 
Moreover, a loss function can be applied at every spatial scale of the decoder. 

\input{chapters/fig_voxelgrid_and_CNN.tex}

Finally, \emph{motion compensation} is a technique to estimate the parameters of the motion that best fits a group of events.
It has a continuous-time warping model that allows to exploit the fine temporal resolution of events (Section~\ref{sec:representations}),
and hence departs from conventional image-based algorithms.
Motion compensation can be used to estimate 
ego-motion \cite{Gallego17ral,Gallego18cvpr},
optical flow \cite{Zhu17icra,Stoffregen17acra,Gallego18cvpr,Zhu19cvpr},
depth \cite{Gallego18cvpr,Rebecq18ijcv,Gallego19cvpr}, 
motion segmentation \cite{Stoffregen17acra,Mitrokhin18iros,Stoffregen19iccv}
or feature motion for VIO \cite{Rebecq17bmvc,Zhu17cvpr}.
The technique in~\cite{Mueggler18tro} also has a continuous-time motion model, albeit not used for motion compensation but rather to fuse event data with IMU data.
To find the parameters of the continuous-time motion models~\cite{Gallego19cvpr,Mueggler18tro}, standard optimization methods, e.g., conjugate gradient or Gauss-Newton, may be applied.

The \emph{number of events per group} (i.e., size of the spatio-temporal neighborhood) is an important hyper-parameter of many methods.
It highly depends on the processing algorithm and the available resources, and accepts multiple selection strategies~\cite{Gallego17ral,Liu18bmvc,Liu19msp,Mostafavi19cvpr}, 
such as constant number of events, constant observation time (i.e., constant frame rate), or more adaptive ones (thresholding the number of events in regions of the image plane)~\cite{Liu18bmvc}.
Utilizing a constant number of events fits naturally with the camera's output rate but it does not account for spatial variations of the rate.
A constant frame rate selects a varying number of events, which may be too few or too many, depending on the scene.
Criteria more adapted to the scene dynamics (in time and space) are often preferred but nontrivial to design.

%% file: chapters/fig_voxelgrid_and_CNN.tex
\begin{figure}[t]
    \centering
    \includegraphics[width=0.9\linewidth]{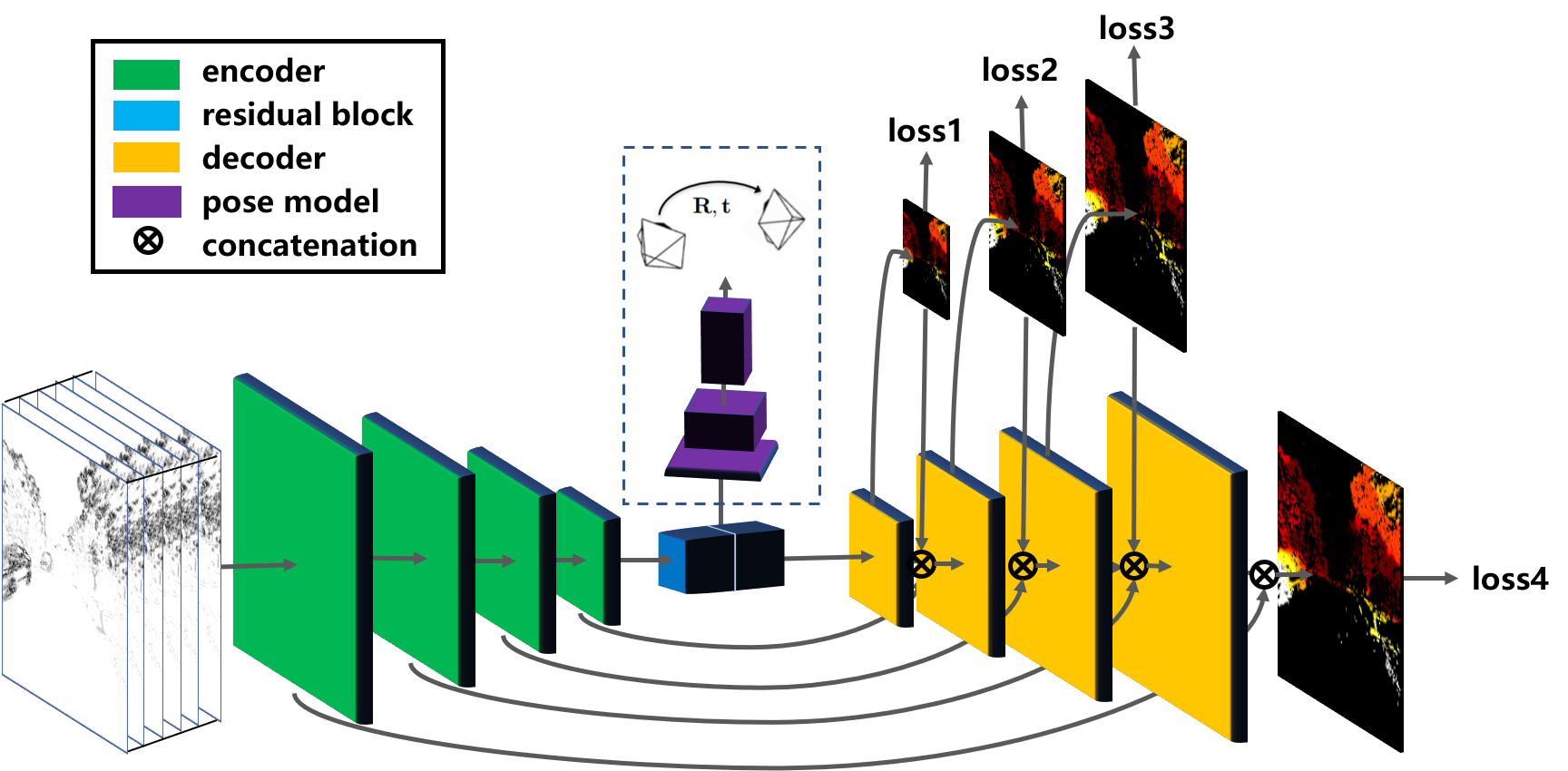}
    \caption{Events in a space-time volume are converted into an interpolated voxel grid (left) that is fed to a DNN to compute optical flow and ego-motion in an unsupervised manner~\cite{Zhu19cvpr}. 
    Thus, modern tensor-based DNN architectures are re-utilized using novel loss functions (e.g., motion compensation) adapted to event data.}
    \label{fig:architecture}
    \vspace{-1ex}
\end{figure}

%% file: chapters/033_processing_bioinspired.tex
\subsection{Biologically Inspired Visual Processing}
\label{sec:bioinspired_processing}

Biological principles and computational primitives drive the design of event camera pixels and some of the event-processing algorithms (and hardware), such as Spiking Neural Networks (\textbf{SNNs}).

\textbf{Visual pathways}:
The DVS~\cite{Lichtsteiner08ssc} was inspired by the function of biological visual pathways, which have ``transient'' pathways dedicated to processing dynamic visual information in the so-called ``where'' pathway\iflongversion \footnote{\url{https://en.wikipedia.org/wiki/Two-streams_hypothesis}}.
\else
.
\fi 
Animals ranging from insects to humans all have these transient pathways. 
In humans, the transient pathway occupies about \SI{30}{\percent} of the visual system. 
It starts with transient ganglion cells, which are mostly found in retina outside the fovea. 
It continues with magno layers of the thalamus and particular sublayers of area V1. 
It then continues to area MT and MST, which are part of the dorsal pathway where many motion selective cells are found~\cite{Liu10nb}.
The DVS corresponds to the part of the transient pathway(s) up to retinal ganglion cells. 
Similarly, the grayscale (EM) events of the ATIS correspond to the ``sustained'' or ``what'' pathway through the parvo layers of the brain~\cite{Posch14ieee,Steffen19fnbot}.

\textbf{Event processing by SNNs}:
Artificial neurons, such as Leaky-Integrate and Fire or Adaptive Exponential, 
are computational primitives inspired in neurons found in the mammalian's visual cortex.
They are the basic building blocks of artificial SNNs.
A neuron receives input spikes (``events'') from a small region of the visual space (a receptive field), 
which modify its internal state (membrane potential) and produce an output spike (action potential) when the state surpasses a threshold.
Neurons are connected in a hierarchical way, forming an SNN.
Spikes may be produced by pixels of the event camera or by neurons of the SNN.
Information travels along the hierarchy, from the event camera pixels to the first layers of the SNN and then through to higher (deeper) layers.
Most first layer receptive fields are based on Difference of Gaussians (selective to center-surround contrast), Gabor filters (selective to oriented edges), and their combinations. The receptive fields become increasingly more complex as information travels deeper into the network.
In ANNs, the computation performed by inner layers is approximated as a convolution.
One common approach in artificial SNNs is to assume that a neuron will not generate any output spikes if it has not received any input spikes from the preceding SNN layer. This assumption allows computation to be skipped for such neurons. %
The result of this visual processing is %
almost simultaneous with the stimulus presentation~\cite{CamunasMesa14biocas},
which is very different from traditional CNNs, 
where convolution is computed simultaneously at all locations at fixed time intervals.

\textbf{Tasks}:
Bio-inspired models have been adopted for several low-level visual tasks.
For example, event-based \emph{optical flow} can be estimated by using spatio-temporally oriented filters~\cite{Delbruck08issle,Orchard13biocas,Brosch15fns} that mimic the working principle of receptive fields in the primary visual cortex~\cite{Chicca06iscas,DeValois00visres}.
The same type of oriented filters have been used to implement a spike-based model of \emph{selective attention}~\cite{Rea13fns} based on the biological proposal from~\cite{Itti01nature}.
Bio-inspired models from binocular vision, such as recurrent lateral connectivity and excitatory-inhibitory neural connections~\cite{Marr76Science}, 
have been used to solve the event-based \emph{stereo} correspondence problem~\cite{Mahowald92thesis,Mahowald94book,Osswald17srep,Dikov17cbbs,Piatkowska13iccvw} 
or to control binocular vergence on humanoid robots~\cite{Vasco16humanoids}.
The visual cortex has also inspired the hierarchical feature extraction model proposed in~\cite{Riesenhuber99nature},
which has been implemented in SNNs and used for \emph{object recognition}.
The performance of such networks improves the better they extract information from the precise timing of the spikes~\cite{Akolkar15neco}.
Early networks were hand-crafted (e.g., Gabor filters)~\cite{Serrano-Gotarredona09tnn}, 
but recent efforts let the network build receptive fields through brain-inspired learning, such as Spike-Timing Dependent Plasticity (\textbf{STDP}), 
yielding better recognition rates~\cite{Akolkar15icra}.
This research is complemented by approaches where more computationally inspired types of supervised learning, such as back-propagation, are used in deep networks to efficiently implement spiking deep convolutional networks~\cite{CamunasMesa14fns,Milde17arxiv,Lee16fns,Stromatias17fns,Neftci17fnins}.
The advantages of the above methods over their traditional vision counterparts are lower latency and higher efficiency.

\iflongversion
To build small, efficient and reactive computational systems, \emph{insect vision} is also a source of inspiration for event-based processing.
To this end, systems for fast and efficient obstacle avoidance and target acquisition in small robots have been developed~\cite{Milde18neco,Blum17rss,Salt17arxiv} based on models of neurons driven by DVS output that respond to looming objects and trigger escape reflexes.
\fi

%% file: chapters/04_algorithms.tex
\section{Algorithms / Applications}
\label{sec:algorithms}
							          \input{chapters/040_algorithm_overview.tex}
								      \input{chapters/041_feature_detection_tracking.tex}
\ifclearsubseclook\cleardoublepage\fi \input{chapters/042_oflow.tex}

\ifclearsubseclook\cleardoublepage\fi \input{chapters/043_depth.tex}
\ifclearsubseclook\cleardoublepage\fi \input{chapters/044_slam.tex}
\iflongversion
\ifclearsubseclook\cleardoublepage\fi \input{chapters/045_visual_inertial.tex}
\fi
\ifclearsubseclook\cleardoublepage\fi \input{chapters/046_image_reconstruction.tex}
\ifclearsubseclook\cleardoublepage\fi \input{chapters/047_segmentation.tex}

\ifclearsubseclook\cleardoublepage\fi \input{chapters/048_recognition.tex}
\ifclearsubseclook\cleardoublepage\fi \input{chapters/049_control.tex}

%% file: chapters/040_algorithm_overview.tex
In this section, we review several works on event-based vision, presented according to the task addressed.
We start with low-level vision on the image plane, such as feature detection, tracking, and optical flow estimation.
Then, we discuss tasks that pertain to the 3D structure of the scene, such as depth estimation, visual odometry (\textbf{VO}) and historically related subjects, e.g., intensity image reconstruction.
Finally, we consider motion segmentation, recognition and coupling perception with control.

%% file: chapters/041_feature_detection_tracking.tex
\subsection{Feature Detection and Tracking}
\label{sec:feature_detection}
\label{sec:feature_tracking}

Feature detection and tracking on the image plane are fundamental building blocks of many vision tasks such as visual odometry, object segmentation and scene understanding.
Event cameras make it possible to track asynchronously, adapted to the dynamics of the scene and with low latency, high dynamic range and low power (Section~\ref{sec:advantageseventcameras}).
Thus, they allow to track in the ``blind'' time between the frames of a standard camera.
To do so, the methods developed need to deal with the unique space-time and photometric characteristics of the visual signal: events report only brightness changes, asynchronously (Section~\ref{subsec:challenges_paradigm_shift}).

\textbf{Challenges}:
Since events represent brightness changes, which depend on motion direction, one of the main challenges of feature detection and tracking with event cameras is overcoming the variation of scene appearance caused by such motion dependency (Fig.~\ref{fig:dataassoc}).
Tracking requires the establishment of correspondences between events (or features built from the events) at different times (i.e., data association), which is difficult due to the varying appearance.
The second main challenge consists of dealing with sensor noise and possible event clutter caused by the camera motion.

\input{chapters/fig_data_association.tex}

\textbf{Literature Review}:
Early event-based feature methods were very simple and focused on demonstrating the low-latency and low-processing requirements of event-driven vision systems.
Hence they assumed a stationary camera scenario and tracked moving objects as clustered \emph{blob-like sources of events}~\cite{Litzenberger06dspws,Litzenberger06itsc,Delbruck07iscas,Drazen11fluids,Delbruck13fns}, circles~\cite{Ni12jm} or lines~\cite{Conradt09iscas}.
Only pixels that generated events needed to be processed.
Simple Gaussian correlation filters sufficed to detect blobs of events, 
which could be modeled by Gaussian Mixtures~\cite{Piatkowska12cvprw}.
For tracking, each incoming event was associated to the nearest existing blob/feature 
and used to asynchronously update its parameters (location, size, etc.).
Circles~\cite{Ni12jm} and lines~\cite{Conradt09iscas} were treated as blobs in the Hough transform space.
These methods were used in traffic monitoring and surveillance~\cite{Litzenberger06dspws,Litzenberger06itsc,Piatkowska12cvprw},
high-speed robotic tracking~\cite{Delbruck07iscas,Delbruck13fns} and particle tracking in fluids~\cite{Drazen11fluids} or microrobotics~\cite{Ni12jm}.
However, they worked only for a limited class of object~shapes.

Tracking of more complex, high-contrast \emph{user-defined shapes} has been demonstrated using event-by-event adaptations of the Iterative Closest Point (ICP) algorithm~\cite{Ni12tro},
gradient descent~\cite{Ni15neco}, Mean-shift and Monte-Carlo methods~\cite{Lagorce15tnnls}, or particle filtering~\cite{Glover17iros}.
The iterative methods in~\cite{Ni12tro,Ni15neco} used a nearest-neighbor strategy to associate incoming events to the target shape and update its transformation parameters, showing very high-speed tracking (\SI{200}{\kilo\Hz} equivalent frame rate).
Other works~\cite{Lagorce15tnnls} handled geometric transformations of the target shape (\emph{aka} ``kernel'') by matching events against a pool of rotated and scaled versions of it.
The predefined kernels tracked the object without overlapping themselves due to a built-in repulsion mechanism.
Complex objects, such as faces or human bodies, have been tracked with part-based shape models~\cite{Valeiras15tnnls},
where objects are represented as a set of basic elements linked by springs~\cite{Fischler73tc}.
The part trackers simply follow incoming blobs of events generated by ellipse-like shapes,
and the elastic energy of this virtual mechanical system provides a quality criterion for tracking.
In most tracking methods events are treated as individual points (without polarity) and update the system's state asynchronously, with minimal latency.
The performance of the methods strongly depends on the tuning of several model parameters, 
which is done experimentally according to the object to track~\cite{Lagorce15tnnls,Valeiras15tnnls}.

The previous methods require a priori knowledge or user input to determine the objects to track.
This restriction is valid for scenarios like tracking cars on a highway or balls approaching a goal, where knowing the objects greatly simplifies the computations.
But when the space of objects becomes larger, methods to determine more \emph{realistic features} become necessary.
The features proposed in \cite{Tedaldi16ebccsp,Zhu17icra} consist of local edge patterns that are represented as point sets.
Incoming events are registered to them by means of some form of ICP.
Other methods~\cite{Rebecq17bmvc,Rosinol18ral} proposed to re-utilize well-known feature detectors~\cite{Harris88} and trackers~\cite{Lucas81ijcai} on patches of motion-compensated event images (Section~\ref{sec:representations}), providing good results.
All these methods allowed to track features for cameras moving in natural scenes, hence enabling ego-motion estimation in realistic scenarios~\cite{Kueng16iros,Zhu17cvpr,Rebecq17bmvc}.
Features built from motion-compensated events (in image form~\cite{Rebecq17bmvc} or point-set form~\cite{Zhu17icra}) provide a useful representation of edge patterns.
However, they depend on motion direction, and, therefore, trackers suffer from drift as event appearance changes over time~\cite{Gehrig19ijcv}.
To track with no drift, motion-invariant features are needed.

\textbf{Combining Events and Frames}:
Data association (Fig.~\ref{fig:dataassoc}) simplifies if the absolute intensity of the pattern to be tracked (Fig.~\ref{fig:dataassoc:preview}, i.e., a motion-invariant representation or ``map'' of the feature) is available.
This is the approach followed by works that leverage the strengths of a combined frame- and event-based sensor (\`a la DAVIS~\cite{Brandli14ssc}).
The algorithms in \cite{Tedaldi16ebccsp,Kueng16iros,Gehrig19ijcv} automatically detect arbitrary edge patterns (features) on the frames and track them asynchronously with events.
The feature location is given by the Harris corner detector~\cite{Harris88} and the feature descriptor is given by the edge pattern around the corner: 
\cite{Tedaldi16ebccsp,Kueng16iros} convert Canny edges to point sets used as templates for ICP tracking, thus they assume events are mostly triggered at strong edges.
In contrast, the edge pattern in \cite{Gehrig19ijcv} is given by the frame intensities, and tracking consists of finding the motion parameters that minimize the photometric error between the events and their frame prediction using a generative model~\eqref{eq:brightnessIncrementLinearized}.
A comparison of five feature trackers is provided in \cite{Gehrig19ijcv}, 
showing that the generative model is most accurate, with sub-pixel performance, albeit it is computationally expensive.
Finally, \cite{Gehrig19ijcv} also shows the interesting fact that an event-based sensor suffices: 
frames can be replaced by images reconstructed from events (Section~\ref{sec:imagereconstruction}) and still achieve similar detection and tracking results.

\textbf{Corner Detection and Tracking}:
Since event cameras naturally respond to edges in the scene, they shorten the detection of lower-level primitives such as keypoints or ``corners''.
Such primitives identify pixels of interest around which local features can be extracted without suffering from the aperture problem, and therefore provide reliable tracking information.
The method in~\cite{Clady15nn} computes corners as the intersection of two moving edges, which are obtained by fitting planes in the space-time stream of events.
To deal with event noise, least-squares is supplemented by a sampling technique similar to RANSAC.
This method of fitting planes locally to time surfaces has also been profitable to estimate optical flow~\cite{Benosman14tnnls} and ``event lifetime''~\cite{Mueggler15icra}, which are obtained from the coefficients of the planes.
Recently, extensions of popular frame-based keypoint detectors, such as Harris~\cite{Harris88} and FAST~\cite{Rosten06eccv}, have been developed for event cameras~\cite{Vasco16iros,Mueggler17bmvc,Alzugaray18ral}, 
by operating on time surfaces (TSs) as if they were natural intensity images.
In~\cite{Vasco16iros} the TS is binarized before applying the derivative filters of Harris' detector.
To speed up detection, \cite{Mueggler17bmvc} replaces the derivative filters with pixelwise comparisons on two concentric circles of the TS around the current event. 
Moving corners produce local TSs with two clearly separated regions: recent vs. old events. 
Hence, corners are obtained by searching for arcs of contiguous pixels with higher TS values than the rest.
The method in~\cite{Alzugaray18ral} improves the detector in~\cite{Mueggler17bmvc} and proposes a strategy to track the corners.
Assuming corners follow continuous trajectories on the image plane and the detected event corners are accurate, these are threaded by proximity along trajectories, following a tree-based hypothesis graph.
The above TS-based hand-crafted corner detectors suffer from variations of the TS due to changes in motion direction.
To overcome them, \cite{Manderscheid19cvpr} proposes a data-driven method to learn the TS appearance of intensity-image corners. %
To this end, a grayscale input (from DAVIS or ATIS camera) provides the supervisory signal to label the corners.
As a trade-off between accuracy and speed, a random forest classifier is used.
Event corners find multiple applications, such as visual odometry or ego-motion segmentation~\cite{Vasco17icar}; yet there are only a few demonstrations.

\textbf{Opportunities}:
In spite of the abundance of detection and tracking methods, they are rarely evaluated on common datasets for performance comparison.
Establishing benchmark datasets~\cite{Hu16fns} and evaluation procedures will foster progress in this and other topics.
Also, in most algorithms, parameters are defined experimentally according to the tracking target. 
It would be desirable to have adaptive parameter tuning to increase the range of operation of the trackers.
Learning-based feature detection and tracking methods also offer considerable room for research.

%% file: chapters/fig_data_association.tex
\begin{figure}
\vspace{-2ex}
\centering
\subfloat[]{\includegraphics[trim={60px 50px 0 0},clip,width=0.31\linewidth]{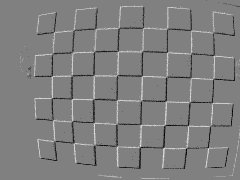}\label{fig:dataassoc:diagonal}}\;
\subfloat[]{\includegraphics[trim={58px 50px 2px 0},clip,width=0.31\linewidth]{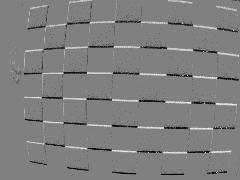}\label{fig:dataassoc:horiz}}\;
\subfloat[]{\includegraphics[trim={40px 50px 20px 0},clip,width=0.31\linewidth]{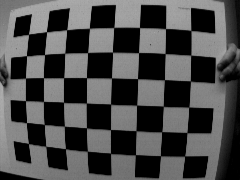}\label{fig:dataassoc:preview}}
\vspace{-1ex}
\caption{The challenge of data association.
Panels~(a) and~(b) show events from a scene (c) under two different motion directions: (a)~diagonal and (b)~up-down.
Intensity increment images~(a) and~(b) are obtained by accumulating event polarities over a short time interval: pixels that do not change intensity are represented in gray, whereas pixels that increased or decreased intensity are represented in bright and dark, respectively.
Clearly, it is not easy to establish event correspondences between~(a) and~(b) due to the changing appearance of the edge patterns in (c) with respect to the motion.
Image adapted from~\cite{Gehrig19ijcv}.
\label{fig:dataassoc}}
\vspace{-1ex}
\end{figure}

%% file: chapters/042_oflow.tex
\vspace{-0.65ex}
\subsection{Optical Flow Estimation}
\label{sec:oflow}

\input{chapters/table_optical_flow.tex}

Optical flow estimation is the problem of computing the velocity of objects on the image plane without knowledge about the scene geometry or motion. 
The problem is ill-posed and thus requires regularization to become tractable.

Event-based optical flow estimation is challenging because of the unfamiliar way in which events encode visual information (Section~\ref{sec:eventCameras}).
In conventional cameras optical flow is obtained by analyzing two consecutive images.
These provide spatial and temporal derivatives that are substituted in the brightness constancy assumption (p.~\pageref{eq_brightness_constancy}), 
which together with smoothness assumptions provide enough equations to solve for the flow at each image pixel.
In contrast, events provide neither absolute brightness nor spatially continuous data.
Each event does not carry enough information to determine flow, and so events need to be aggregated to produce an estimate,
which leads to the unusual question of where in the $x$-$y$-$t$-space of the image plane spanned by the events is flow computed.
Ideally one would like to know the flow field over the whole space, which deems computationally expensive. %
In practice, optical flow is computed only at specific points: at the event locations, or at images with artificially-chosen times.
Nevertheless, computing flow from events is attractive 
because they represent edges, which are the parts of the scene where flow estimation is less ambiguous, 
and because their fine timing information allows measuring high speed flow~\cite{Liu19msp}. %
Finally, another challenge is to design a flow estimation algorithm that is biologically plausible, 
i.e., compatible with what is known from neuroscience about early processing in the primate visual cortex, 
and that can be implemented efficiently in neuromorphic processors.

\textbf{Literature Review}: Table~\ref{tab:optical_flow_classification} lists some event-based optical flow methods, 
categorized according to different criteria.
Early works~\cite{Benosman12nn} tried to adapt classical approaches in computer vision to event-based data (Fig.~\ref{fig:oflow:flying:eklt}).
These are based on the brightness constancy assumption~\cite{Lucas81ijcai}, 
and discussion focused on whether events carried enough information to estimate flow with such approaches~\cite{Brosch15fns}.
Events allow to estimate the temporal derivative of brightness~\eqref{eq:TemporalDerivBrightness}, 
and so additional assumptions were needed to approximate the spatial derivative $\nabla \Lum$ in order to apply such classical methods~\cite{Lucas81ijcai}.
However, due to the potentially very small number of events generated at each pixel as an edge crosses over it, 
it is difficult to estimate derivatives ($\nabla \Lum, \partial \Lum/\partial t$) reliably~\cite{Brosch15fns}, 
which leads gradient-based methods like~\cite{Benosman12nn} to inconclusive flow estimates.
Approaches that consider the local distribution of events in the $x$-$y$-$t$-space, as in~\cite{Benosman14tnnls}, are more robust and therefore preferred.

The method in \cite{Benosman14tnnls} reasons about the local distribution of events geometrically, in terms of time surfaces and planar approximations.
As an edge moves it produces events that resemble points on a surface in space-time (the time surface, Section~\ref{sec:event_processing}).
The surface slopes in the $x$-$t$ and $y$-$t$ cross sections encode the edge motion, 
thus optical flow is estimated by fitting planes to the surface and reading the slopes from the plane coefficients.
In spite of providing only normal flow (i.e., the component of the optical flow perpendicular to the edge), 
the method works even in the case of only a few generated events.
Of course, the goodness of fit depends on the size of the spatio-temporal neighborhood (this remark generalizes to other methods).
If the neighborhood is too small then the plane fit may become arbitrary. %
If the neighborhood is too large then the event stream may not be well approximated by a local plane.

\input{chapters/fig_optical_flow.tex}

A hierarchical architecture for optical flow estimation building on experimental findings of the primate visual system is proposed in \cite{Brosch15fns}.
It applies a set of spatio-temporal filters on the event stream to yield selectivity to different motion speeds and directions (\`a la Gabor filters) while maintaining the sparse representation of events. 
Such filters are formally equivalent to spatio-temporal correlation detectors.
Other biologically-inspired methods \cite{Orchard13biocas,Paredes19pami} can also be interpreted as filter banks sampling the event stream along different spatio-temporal orientations; 
\cite{Orchard13biocas} and \cite{Brosch15fns} define hand-crafted filters, whereas \cite{Paredes19pami} learns them from event data using a novel STDP rule.
The SNN in \cite{Orchard13biocas} detects motion patterns by delaying events through synaptic connections and employing neurons as coincidence detectors.
Its neurons are sensitive to 8 speeds and 8 directions (i.e., 64 velocities) over receptive fields of $5\times 5$ pixels.
These methods are implementable in neuromorphic hardware, offering low-power, efficient computations.

\iflongversion
The method in~\cite{Barranco15iwann} targets the issue of flow estimation at textured regions.
It converts events into event frames and applies a Gabor filter bank.
Then, assuming constancy of the (spatio-temporal) edges of the event frame, optical flow is given by the phase gradient~\cite{Fleet90ijcv} of the filter bank output.
Due to the aperture problem, only the velocity in the direction of the spatial phase gradient can be computed.
\fi

Methods like~\cite{Bardow16cvpr,Cook11ijcnn} estimate optical flow jointly with other quantities, notably image intensity, 
so that the quantities involved bring in well-known equations and boost each other towards convergence.
Knowing image intensity, or equivalently ($\nabla \Lum, \partial \Lum/\partial t$), 
is desirable since it can be used on the brightness constancy law to provide constraints on the optical flow.
In this respect, \cite{Bardow16cvpr} combines multiple equations (\eqref{eq:EventTriggeringCondition}, brightness constancy, smoothness priors, etc.) as penalty terms into an objective function that is optimized via calculus of variations.
The method finds the optical flow and image intensity on the image plane that minimizes the objective function, i.e., that best explains the distribution of events in the $x$-$y$-$t$-space (using a voxel grid).
Thus, it outputs a dense flow (i.e., flow at every pixel).
Flow vectors at pixels where no events were produced (i.e., regions of homogeneous brightness) are due to the smoothness priors, thus they are less reliable than those computed at pixels where events were triggered (i.e., at edges).

The method in \cite{Liu18bmvc} estimates optical flow by computing event frames (Section~\ref{sec:event_processing}) at an adaptive rate and applying video coding techniques (block matching).
It can be interpreted as finding the optical flow vector that best matches the distributions of events within two cuboids (collapsed into event frames).
Thus, the optical flow problem is posed as that of finding event correspondences, i.e., events triggered by the same scene point (at different times).
The method defines two sets of events (``blocks'') and a similarity metric to compare them. 
It is assumed that the appearance of event frames do not change significantly for short times and hence simple metrics, such as sum of absolute distances, suffice to compare them.
The method can be implemented in FPGA, trading off efficiency for accuracy.

The framework in~\cite{Gallego18cvpr,Stoffregen17acra,Gallego19cvpr} 
computes optical flow by maximizing the sharpness of image patches obtained by warping cuboids of events, 
producing motion-compensated images (Section~\ref{sec:event_processing}).
It can be interpreted as applying an adaptive filter to the events,
where the filter coefficients define the spatio-temporal direction that maximizes the filter's response. 
Motion compensation was also used to compute flow in \cite{Zhu17icra}, albeit using point sets.

Recently, deep learning methods have emerged~\cite{Zhu18rss,Zhu19cvpr,Ye18arxiv}.
These are based on the availability of large amounts of event data paired with an ANN.
In \cite{Zhu18rss}, an encoder-decoder CNN is trained using a self-supervised scheme to estimate dense optical flow. 
The loss function measures the error between DAVIS grayscale images aligned using the flow produced by the network.
The trained network is able to accurately predict optical flow from events only, passed as time surfaces and event frames.
The work~\cite{Ye18arxiv} presents the first monocular ANN architecture %
to estimate dense optical flow, depth and ego-motion (i.e., learning structure from motion) from events only.
The input to the ANN consists of events over multiple time slices, given as event frames and time surfaces with average timestamps.
This reduces event noise and preserves the structure of the event stream better than~\cite{Zhu18rss}. %
The network is trained unsupervised, measuring the photometric error between the events in neighboring time slices aligned using the estimated flow.
Later, \cite{Zhu18rss} was extended to unsupervised learning of flow and ego-motion in~\cite{Zhu19cvpr} using a motion-compensation loss function in terms of time surfaces.

\textbf{Evaluation}:
Optical flow estimation is computationally expensive. 
Some methods~\cite{Bardow16cvpr,Zhu18rss,Ye18arxiv,Zhu19cvpr} require a GPU, 
while other approaches are more lightweight~\cite{Liu18bmvc}, albeit not as accurate.
Few algorithms~\cite{Orchard13biocas,Benosman14tnnls,Brosch15fns,Liu18bmvc} have been pushed to hardware logic circuits that offload CPU and minimize latency.
The review~\cite{Rueckauer16fns} compared some early event-based optical flow methods~\cite{Delbruck08issle,Benosman12nn,Benosman14tnnls}, 
but only on flow fields generated by a rotating camera, i.e., lacking motion parallax and occlusion. %
For newer methods, there are multiple trade offs (accuracy vs. efficiency vs. latency) that have not been properly quantified yet.

\textbf{Opportunities}:
Comprehensive datasets with accurate ground truth optical flow in multiple scenarios (varying texture, speed, parallax, occlusions, illumination, etc.)
and a common evaluation methodology would be essential to assess progress and reproducibility in this paramount low-level vision task.
Providing ground truth \emph{event-based} optical flow in real scenes is challenging,
especially for moving objects not conforming to the motion field induced by the camera's ego-motion.
A thorough quantitative comparison of existing event-based optical flow methods would help identify key ideas to develop improved methods.

%% file: chapters/table_optical_flow.tex
\begin{table}
\centering
\caption{\label{tab:optical_flow_classification}Classification of several optical flow methods according to their output and design.
Some methods provide full motion flow (F) whereas others only its component normal to the local brightness edge (N).
The output may be a dense (D) flow field (i.e., optical flow for every pixel at some time) or sparse (S) (i.e., flow computed at selected pixels).
According to their design, methods may be model-based or model-free (Artificial Neural Network - ANN),
and neuro-biologically inspired or not.}
\vspace{-1ex}
\begin{adjustbox}{max width=\columnwidth}\setlength{\tabcolsep}{6pt}
\begin{tabular}{lcccc}
\toprule
\textbf{Reference} & \textbf{N/F?}  & \textbf{S/D?} & \textbf{Model?} & \textbf{Bio?}\\
\midrule
Delbruck \cite{Delbruck08issle,Rueckauer16fns} & Normal & Sparse & Model & Yes\\
Benosman et al. \cite{Benosman12nn,Rueckauer16fns} & Full & Sparse & Model & No\\
Orchard et al. \cite{Orchard13biocas} 		& Full & Sparse & ANN & Yes\\
Benosman et al. \cite{Benosman14tnnls,Rueckauer16fns} & Normal & Sparse & Model & No\\
Barranco et al. \cite{Barranco14ieee} 		& Normal & Sparse & Model & No\\
\iflongversion
Barranco et al. \cite{Barranco15iwann} 		& Normal & Sparse & Model & No\\ %
Conradt et al. \cite{Conradt15robio} 		& Normal & Sparse & Model & No\\ %
\fi
\iflongversion
Brosch et al. \cite{Brosch15fns,Tschechne14annpr} & Normal & Sparse & Model & Yes\\
\else
Brosch et al. \cite{Brosch15fns} & Normal & Sparse & Model & Yes\\
\fi
Bardow et al. \cite{Bardow16cvpr} 			& Full & Dense & Model & No\\
Liu et al. \cite{Liu18bmvc} 		& Full & Sparse & Model & No\\
Gallego\cite{Gallego18cvpr}, Stoffregen\cite{Stoffregen17acra} & Full & Sparse & Model & No\\
Haessig et al. \cite{Haessig18tbcas} 		& Normal & Sparse & ANN & Yes\\
Zhu et al. \cite{Zhu18rss,Zhu19cvpr}       & Full & Dense & ANN & No\\
Ye et al. \cite{Ye18arxiv}	& Full & Dense & ANN & No\\
Paredes-Vall\'es \cite{Paredes19pami} 	    & Full & Sparse & ANN & Yes\\
\bottomrule
\end{tabular}
\end{adjustbox}
\vspace{-1ex}
\end{table}

%% file: chapters/fig_optical_flow.tex
\global\long\def\flowheight{2.12cm}
\begin{figure}[t]
\centering
  \subfloat[]{\includegraphics[trim={1cm 1cm 1cm 0.5cm},clip,height=\flowheight]{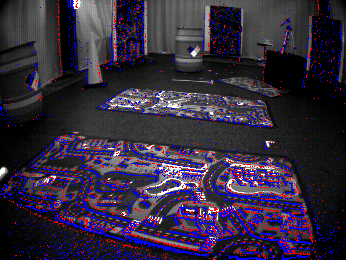}
    \label{fig:oflow:flying:events}}
  \subfloat[]{\includegraphics[trim={1cm 1cm 1cm 0.5cm},clip,height=\flowheight]{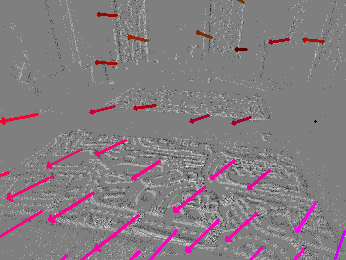}
    \label{fig:oflow:flying:eklt}}
  \subfloat[]{\includegraphics[height=\flowheight]{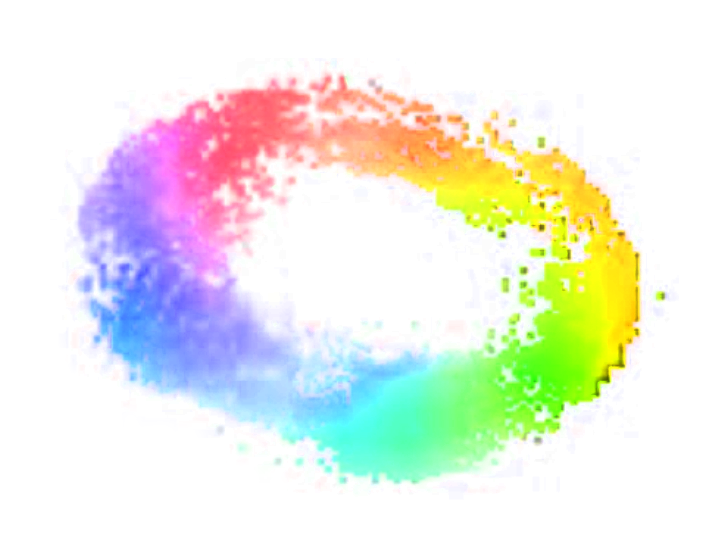}
  \label{fig:oflow:figetspinner}}
\vspace{-1ex}
\caption{\label{fig:oflow:flying}Two optical flow estimation examples.
(a) and (b): \emph{indoor flying} scene~\cite{Zhu18ral}. 
In (a), events (polarity shown in red/blue) are overlaid on a grayscale frame from a DAVIS. 
(b) shows the sparse optical flow (colored according to magnitude and direction) computed using~\cite{Lucas81ijcai} on brightness increment images. 
(c) A different scene: dense optical flow of a fidget spinner spinning at \SI{750}{\degree/\second} in a dark environment~\cite{Zhu19cvpr}. 
Events enable the estimation of optical flow in challenging scenarios.}
\vspace{-1ex}
\end{figure}

%% file: chapters/043_depth.tex
\subsection{3D reconstruction. Monocular and Stereo}
\label{sec:depth}
Depth estimation with event cameras is a broad field. 
It can be divided according to the considered scenario and camera setup or motion, which determine the problem assumptions.

\textbf{Instantaneous Stereo}:
Most works on depth estimation with event cameras target the problem of ``instantaneous'' stereo, i.e.,
3D reconstruction using events on a very short time (ideally on a per-event basis) 
from two or more synchronized cameras that are rigidly attached.
Being synchronized, the events from different image planes share a common clock.
These works follow the classical two-step stereo solution:
first solve the event correspondence problem across image planes (i.e., epipolar matching) and then triangulate the location of the 3D point~\cite{Hartley03book}.
The main challenge is finding correspondences between events; %
it is the computationally intensive step.
Events are matched 
(\emph{i}) using traditional stereo metrics (e.g., normalized cross-correlation) on event frames~\cite{Schraml10iscas,Kogler11book} or time surfaces \cite{Ieng18fnins} (Section~\ref{sec:event_processing}),
and/or (\emph{ii}) by exploiting simultaneity and temporal correlations of the events across sensors~\cite{Kogler11isvc,Lee12iscas,Ieng18fnins}.
These approaches are \emph{local}, matching events by comparing their neighborhoods since events cannot be matched based on individual timestamps \cite{Piatkowska14msci,CamunasMesa14fns}.
Additional constraints, such as the epipolar constraint~\cite{Benosman11tnn}, ordering, uniqueness, edge orientation and polarity may be used to reduce matching ambiguities and false correspondences, thus improving depth estimation~\cite{Rogister12tnnls,Carneiro13nn,CamunasMesa14fns}.
Event matching can also be done by comparing local context descriptors~\cite{Zou16icip,Zou17bmvc} of the
spatial distribution of events on both stereo image planes.

\emph{Global} approaches produce better depth estimates (i.e., less sensitive to ambiguities) than local approaches by considering additional regularity constraints.
In this category, we find extensions of Marr and Poggio's cooperative stereo algorithm~\cite{Marr76Science} for the case of event cameras \cite{Mahowald92thesis,Piatkowska13iccvw,Firouzi16npl,Osswald17srep,Dikov17cbbs}.
These approaches consist of a network of disparity sensitive neurons that receive events from both cameras and perform various operations (amplification, inhibition) that implement matching constraints (uniqueness, continuity) to extract disparities.
They use not only the temporal similarity to match events but also their spatio-temporal neighborhoods, with iterative nonlinear operations that result in an overall globally-optimal solution.
A discussion of cooperative stereo is provided in~\cite{Steffen19fnbot}.
Also in this category are~\cite{Kogler14jei,Xie17fns,Xie18ijars}, which use Belief Propagation on a Markov Random Field or semiglobal matching~\cite{Hirschmuller08pami} to improve stereo matching.
These methods are primarily based on optimization, trying to define a well-behaved energy function whose minimizer is the correct correspondence map.
The energy function incorporates regularity constraints, which enforce coupling of correspondences at neighboring points and therefore make the solution map less sensitive to ambiguities than local methods, at the expense of computational effort.
A table comparing different stereo methods is provided in~\cite{Andreopoulos18cvpr};
however, it should be interpreted with caution since the methods were not benchmarked on the same dataset.

Recently, brute-force space-sweeping using dedicated hardware (a GPU) has been proposed~\cite{Zhu18eccv}.
The method is based on ideas similar to~\cite{Rebecq18ijcv,Gallego18cvpr}:
the correct depth manifests as ``in focus'' voxels of displaced events
in the Disparity Space Image~\cite{Szeliski10book,Rebecq18ijcv}.
In contrast, other approaches pair event cameras with neuromorphic processors (Section~\ref{sec:hardware}) to produce fully event-based low-power (\SI{100}{\milli\watt}), high-speed stereo systems~\cite{Dikov17cbbs,Andreopoulos18cvpr}. %
There is an efficiency vs.~accuracy trade-off that has not been quantified yet.

Most of the methods above are demonstrated in scenes with static cameras and few moving objects, so that correspondences are easy to find due to uncluttered event data.
Event matching happens with low latency, at high rate ($\sim$\SI{1}{\kilo\Hz}) and consuming little power, which shows that event cameras are promising for high-speed 3D reconstructions of moving objects or in uncluttered scenes.

\iflongversion
\textbf{Multi-Perspective Panoramas}:
Some works~\cite{Schraml16tie,Schraml15cvpr} also target the problem of instantaneous stereo
(depth maps produced using events over very short time intervals),
but using two non-simultaneous event cameras.
These methods exploit a constrained hardware setup (two rotating event cameras with known motion) to either (\emph{i}) recover intensity images on which conventional stereo is applied~\cite{Schraml16tie}
or (\emph{ii}) match events using temporal metrics~\cite{Schraml15cvpr}.
\fi

\input{chapters/fig_depth_EMVS.tex}

\textbf{Monocular Depth Estimation}:
Depth estimation with a single event camera has been shown in~\cite{Rebecq18ijcv,Kim16eccv,Gallego18cvpr}.
It is a significantly different problem from previous ones because temporal correlation between events across multiple image planes cannot be exploited.
These methods recover a semi-dense 3D reconstruction of the scene (i.e., 3D edge map) by integrating information from the events of a moving camera over time,
and therefore require knowledge of camera motion.
Hence they do not pursue instantaneous depth estimation, but rather depth estimation for SLAM~\cite{Cadena16tro}.

The method in~\cite{Kim16eccv} is part of a pipeline that uses three filters operating in parallel to jointly estimate the motion of the event camera, a 3D map of the scene, and the intensity image.
Their depth estimation approach requires using an additional quantity---the intensity image---to solve for data association.
In contrast, \cite{Rebecq18ijcv} (Fig.~\ref{fig:depth:building}) proposes a space-sweep method that leverages the sparsity of the event stream to perform 3D reconstruction without having to establish event matches or recover the intensity images.
It back-projects events into space, creating a ray density volume~\cite{Collins96cvpr}, and then finds scene structure as local maxima of ray density.
It is computationally efficient and used for VO in~\cite{Rebecq17ral}.

\iflongversion
\textbf{Stereo Depth for SLAM}:
Recently, inspired by work in small-baseline multi-view stereo~\cite{Newcombe11iccv}, a stereo depth estimation method for SLAM has been presented~\cite{Zhou18eccv}.
It obtains a semi-dense 3D reconstruction of the scene by optimizing the local spatio-temporal consistency of events across image planes using time surfaces.
It does not follow the classical paradigm of event matching plus triangulation~\cite{Ieng18fnins}, but rather a forward-projection approach that enables depth estimation without establishing event correspondences explicitly. 
The method opens the door for bringing the advantages of event cameras to event-based stereo SLAM applications such as self-driving cars.

\textbf{Depth Estimation using Structured Light}:
All the above 3D reconstruction methods are passive, i.e., do not interfere with the scene.
In contrast, there are some works on event-based active 3D reconstruction, based on emitting light onto the scene and measuring reflection with event cameras~\cite{Brandli13fns,Matsuda15iccp,Martel18iscas}.
For example, \cite{Brandli13fns} combines a DVS with a pulsed line laser to allow fast terrain reconstruction, in the style of a 3D line scanner.
Motion Contrast 3D scanning~\cite{Matsuda15iccp} is a structured light technique that simultaneously achieves high resolution, high speed and robust performance in challenging 3D scanning environments (e.g., strong illumination, or highly reflective and moving surfaces).
Active systems with pulsed lasers exploit the high temporal resolution and redundancy suppression of event cameras, but they are application specific and may not be safe (depending on the power of the laser needed to scan far away objects).
\fi

\textbf{Opportunities}:
Although there are many methods for event-based depth estimation, it is difficult to compare their performance since they are not evaluated on the same dataset.
In this sense, it would be desirable to
(\emph{i}) provide a comprehensive dataset and testbed for event-based depth evaluation
and (\emph{ii}) benchmark many existing methods on the dataset, to be able to compare their performance.

%% file: chapters/fig_depth_EMVS.tex
\iffalse
\begin{figure}[t]
\centering
\global\long\def\heightMonoDepth{2.9cm}
\subfloat[]{\includegraphics[height=\heightMonoDepth]{images/slam/building_preview.jpg}
  \label{fig:slam:building:preview}}\;
\subfloat[]{\includegraphics[height=\heightMonoDepth]{images/slam/building_events.png}
  \label{fig:slam:building:events}}\\[0.5ex]
\subfloat[]{\includegraphics[height=\heightMonoDepth]{images/slam/building_depth_white.png}
  \label{fig:slam:building:depth}}\;
\subfloat[]{\includegraphics[height=\heightMonoDepth]{images/slam/building_pcl.png}
  \label{fig:slam:building:pointcloud}}
\caption{Example of monocular depth estimation with a hand-held event camera.
(a)~Scene,
(b)~events (positive and negative) in a short time interval,
(c)~semi-dense depth map, pseudo-colored from red (close) to blue (far),
(d)~3D point cloud.
Image courtesy of~\cite{Rebecq18ijcv}.}
\label{fig:depth:building}
\end{figure}
\fi

\begin{figure}[t]
\centering
\global\long\def\heightMonoDepth{3cm}
\subfloat[]{\includegraphics[height=\heightMonoDepth]{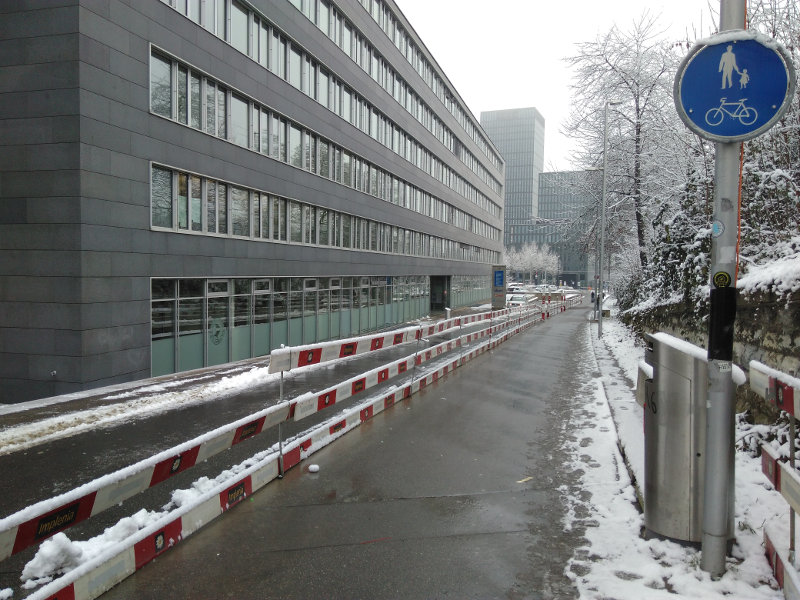}
  \label{fig:slam:building:preview}}\;\;
  \subfloat[]{\includegraphics[height=\heightMonoDepth]{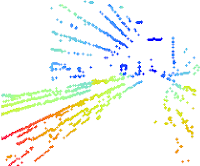}
  \label{fig:slam:building:depth}}
  \vspace{-1ex}
\caption{Example of monocular depth estimation with a hand-held event camera.
(a)~Scene,
(b)~semi-dense depth map, pseudo-colored from red (close) to blue (far).
Image courtesy of~\cite{Rebecq18ijcv}.}
\label{fig:depth:building}
\iflongversion
\vspace{-1.7ex}
\else
\vspace{-1ex}
\fi
\end{figure}

%% file: chapters/044_slam.tex
\subsection{Pose Estimation and SLAM}
\label{sec:slam}

Addressing the Simultaneous Localization and Mapping (SLAM) problem with event cameras has been difficult because most methods and concepts developed for conventional cameras (feature detection, matching, iterative image alignment, etc.) are not applicable or were not available; 
events are fundamentally different from images.
The challenge is therefore to design new SLAM techniques that are able to unlock the camera's advantages (Sections \ref{subsec:challenges_paradigm_shift} and \ref{sec:advantageseventcameras}), showing their usefulness to tackle difficult scenarios for current frame-based cameras.
Historically, the design goal of such techniques has focused on preserving the low-latency nature of the data, i.e., being able to produce a state estimate for every incoming event (Section~\ref{sec:event_processing}). 
However, each event does not contain enough information to estimate the state from scratch (e.g., the six degrees of freedom (DOF) pose of a calibrated camera),
and so the goal becomes that each event be able to asynchronously update the state of the system.
Probabilistic (Bayesian) filters~\cite{Thrun05book} are popular in event-based SLAM~\cite{Weikersdorfer13icvs,Censi14icra,Kim14bmvc,Gallego17pami} because they naturally fit with this description.
Their main adaptation for event cameras consists of designing sensible likelihood functions based on the event generation process (Section~\ref{subsec:event_generation_model}).

Since events are caused by the apparent motion of intensity edges, the majority of maps emerging from SLAM systems naturally consist only of scene edges, i.e., semi-dense maps (Fig.~\ref{fig:slam:dynamic} and~\cite{Rebecq18ijcv}).
However, note that an event camera does not directly measure intensity gradients but only temporal changes~\eqref{eq:EventTriggeringCondition}, 
and so the presence, orientation and strength of edges (on the image plane and in 3D) must be estimated together with the camera's motion.
The strength of the intensity gradient at a scene point is correlated with the firing rate of events corresponding to that point, 
and it enables reliable tracking~\cite{Reinbacher17iccp}.
Edge information for tracking may also be obtained from gradients of brightness maps~\cite{Kim14bmvc,Kim16eccv,Gallego17pami} used in generative models (Section~\ref{subsec:event_generation_model}).

The event-based SLAM problem in its most general setting (\mbox{6-DOF} motion and natural 3D scenes) is a challenging problem that has been addressed step-by-step in scenarios with increasing complexity.
Three complexity axes can be identified:
dimensionality of the problem, type of motion and type of scene.
The literature is dominated by methods that address the localization subproblem first (i.e., motion estimation)
because it has fewer degrees of freedom to estimate. %
Regarding the type of motion, solutions for constrained motions, such as rotational %
or planar %
(both being \mbox{3-DOF}), have been investigated before addressing the most complex case of a freely moving camera (\mbox{6-DOF})%
.
Solutions for artificial scenes in terms of photometry (high contrast) and/or structure (line-based or 2D maps) %
have been proposed before focusing on the most difficult case: natural scenes (3D and with arbitrary photometric variations). %
Some proposed solutions require additional sensing (e.g., \mbox{RGB-D}) to reduce the complexity of the problem.
This, however, introduces some of the bottlenecks present in frame-based systems (e.g., latency and motion blur).
Table~\ref{tab:slam} classifies the related work using these complexity axes.

\input{chapters/table_related_work_slam.tex}

\iflongversion
\textbf{Camera Tracking Methods}:
Pose tracking with an event camera was first presented in~\cite{Weikersdorfer12robio}.
It proposed a particle filter to track the motion of a ground robot that was viewing a flat scene, which was parallel to the plane of motion and consisted of artificial B\&W line patterns.
The main innovation was the design of the likelihood function to quantify the event generation probability given the robot's pose and scene map. 
The function was based on the reprojection error between the event's location and the nearest edge in the map.
In~\cite{Censi14icra}, a standard grayscale camera was attached to a DVS to estimate, using a Bayesian filter, the small displacement between the current event and the previous frame of the standard camera.
The system was developed for planar motion and B\&W scenes.
The likelihood function was proportional to the strength of the gradient of the map at the event's location (Section~\ref{subsec:event_generation_model}).
In \cite{Milford15rssw}, pose tracking under a non-holonomic and planar motion was proposed, supporting loop closure and topologically-correct trajectories. It converted events into frames and used traditional method SeqSLAM.

Estimation of the 3D orientation of an event camera has been addressed in~\cite{Kim14bmvc,Gallego17ral,Reinbacher17iccp}.
The rotational motion of the camera was tracked by~\cite{Kim14bmvc} using a particle filter, whose likelihood function was approximately Gaussian centered at a contrast sensitivity $C\approx 0.2$ (Section~\ref{subsec:event_generation_model}).
The work in \cite{Reinbacher17iccp} estimated the camera's rotation by minimization of a photometric error at the event locations; it used a map of event probabilities that represented the strength of the scene edges~\cite{Weikersdorfer13icvs}.
Finally, the motion-compensation optimization framework (Section~\ref{sec:event_processing}) was introduced in~\cite{Gallego17ral} to estimate the angular velocity of the camera rather than its absolute rotation.
These systems are restricted to rotational motions, and, thus, do not account for translation or depth.
Nevertheless they inspire ideas to solve more complex problems, such as~\cite{Kim16eccv,Rebecq17bmvc,Gallego18cvpr}.

Camera tracking in \mbox{6-DOF} is the most challenging one.
An event-based algorithm to track the pose of a DVS during very high-speed motion was presented in~\cite{Mueggler14iros}.
This was a hand-crafted method developed for artificial, B\&W line-based maps, rather than a probabilistic filter.
It assumed that events were generated only by one of the map lines (the one closest to the event), which were tracked and intersected to provide points for PnP methods (camera resectioning).
A continuous-time formulation and extension of such method was given in~\cite{Mueggler15rss}.
It computed the camera trajectory, rather than individual poses, by non-linear optimization of the event-to-line reprojection error.
By contrast, \cite{Gallego17pami,Bryner19icra} showed 6-DOF high-speed tracking capabilities in natural scenes.
They used a generative model: \cite{Gallego17pami} proposed a probabilistic filter with a robust likelihood function comprising mixture densities (Section~\ref{subsec:event_generation_model}), whereas \cite{Bryner19icra} pursued non-linear optimization of the photometric error between brightness increment images (Section~\ref{sec:representations}) and their prediction given the scene map.
The latter gave slightly better results.
\fi %

\textbf{Tracking and Mapping}:
Let us focus on methods that address the tracking-and-mapping problem.
Cook et al.~\cite{Cook11ijcnn} proposed a generic message-passing algorithm within an interacting network to jointly estimate ego-motion, image intensity and optical flow from events.
However, the system was restricted to rotational motion. %
Joint estimation is appealing because it allows to employ as many equations as possible relating the variables (e.g., \eqref{eq:brightnessIncrementLinearized} and rotational prior) in the hope of finding a better solution to the problem.

An event-based 2D SLAM system was presented in~\cite{Weikersdorfer13icvs} by extension of~\cite{Weikersdorfer12robio}, and thus it was restricted to planar motion and high-contrast scenes.
The method used a particle filter for tracking, with the event likelihood function inversely related to the the reprojection error of the event with respect to the map. 
The map of scene edges was concurrently built; it consisted of an occupancy map~\cite{Thrun05book}, with each pixel representing the probability that the pixel triggered events.
The method was extended to 3D in~\cite{Weikersdorfer14icra}, but it relied on an external \mbox{RGB-D} sensor attached to the event camera for depth estimation.
The depth sensor introduced bottlenecks, which deprived the system of the low latency and high-speed advantages of event cameras.

\input{chapters/fig_SLAM_TRO.tex}

The filter-based approach in~\cite{Kim14bmvc} showed how to simultaneously track the 3D orientation of a rotating event camera and create high-resolution panoramas of natural scenes.
It operated probabilistic filters in parallel for both subtasks.
A panoramic gradient was built using per-pixel Kalman filters, each one estimating the orientation and strength of the scene edge at its location.
This gradient map was then upgraded to an absolute intensity one with super-resolution and HDR properties by Poisson integration.
SLAM during rotational motion was also presented in~\cite{Reinbacher17iccp}, where camera tracking was performed by minimization of a photometric error at the event locations given a probabilistic edge map.
The map was simultaneously built, and each map point represented the probability of events being generated at that location~\cite{Weikersdorfer13icvs}.
Hence it was a panoramic occupancy map measuring the strength of the scene edges.

Recently, solutions to the full problem of event-based 3D SLAM for \mbox{6-DOF} motions and natural scenes, not relying on additional sensing, have been proposed~\cite{Kim16eccv,Rebecq17ral} (Table~\ref{tab:slam}).
The approach in~\cite{Kim16eccv} extends~\cite{Kim14bmvc} and consists of three interleaved probabilistic filters to perform pose tracking as well as depth and intensity estimation.
\sle{However, it suffers from limited robustness (especially during initialization) due to the assumption of uncorrelated depth, intensity gradient, and camera motion. Furthermore, it} is computationally intensive, requiring a GPU for real-time operation.
In contrast, the semi-dense approach in~\cite{Rebecq17ral} shows that intensity reconstruction is not needed for depth estimation or pose tracking.
The approach has a geometric foundation: it performs space sweeping for 3D reconstruction~\cite{Rebecq18ijcv} and edge-map alignment (non-linear optimization with few events per frame) for pose tracking.
The resulting SLAM system runs in real-time on a CPU.

Trading off latency for efficiency, probabilistic filters~\cite{Weikersdorfer13icvs,Kim14bmvc,Kim16eccv} can operate on small groups of events.
Other approaches are natively designed for groups, based for example on non-linear optimization~\cite{Rebecq17ral,Gallego17ral,Gallego18cvpr}, and run in real time on the CPU.
Processing multiple events simultaneously is also beneficial to reduce noise.

\textbf{Opportunities}:
The above-mentioned SLAM methods lack loop-closure capabilities to reduce drift.
Currently, the scales of the scenes on which event-based SLAM has been demonstrated are considerably smaller than those of frame-based SLAM.
However, trying to match both scales may not be a sensible goal since event cameras may not be used to tackle the same problems as standard cameras; 
both sensors are complementary, as argued in~\cite{Censi14icra,Gallego17pami,Rosinol18ral,Gehrig19ijcv}.
Stereo event-based SLAM~\cite{Zhou20arxiv} is an unexplored topic, as well as designing more accurate, efficient and robust methods than the existing monocular ones.
Robustness of SLAM systems can be improved by sensor fusion with IMUs~\cite{Cadena16tro,Rosinol18ral}.

%% file: chapters/table_related_work_slam.tex
\global\long\def\TwoD{2D}
\global\long\def\ThreeD{3D}
\global\long\def\ThreeDOF{3}
\global\long\def\SixDOF{6}
\global\long\def\BlackWhite{B\&W}
\global\long\def\BlackWhiteLines{B\&W, lines}
\global\long\def\Natural{natural}

\begin{table}[t]
\centering
\caption{\label{tab:slam}Event-based methods for pose tracking and/or mapping with an event camera.
The type of motion is noted with labels ``2D'' (3-DOF motions, e.g., planar or rotational) and ``3D'' (free 6-DOF motion in 3D space). 
Columns indicate whether the method performs tracking (``Track'') and depth estimation (``Depth'') using only events (``Event''), 
the type of scene considered (``Scene''), 
and any additional requirements.
Only \cite{Kim16eccv,Rebecq17ral} address the most general scenario using only events.}
\vspace{-1ex}
\begin{adjustbox}{max width=\columnwidth}
\setlength{\tabcolsep}{1.5pt}
\begin{tabular}{lcccccl}
\toprule
\textbf{Reference} & \textbf{Dim} & \textbf{Track} & \textbf{Depth} & \textbf{Scene} & \textbf{Event} & \textbf{Additional requirements}\\
\midrule
Cook \cite{Cook11ijcnn} & \TwoD & \cmark & \xmark & \Natural & \cmark & rotational motion only
\\
Weikersdorfer \cite{Weikersdorfer13icvs} & \TwoD & \cmark & \xmark & \BlackWhite & \cmark & scene parallel to motion\\
Kim \cite{Kim14bmvc} & \TwoD & \cmark & \xmark & \Natural & \cmark & rotational motion only \\
Gallego \cite{Gallego17ral} & \TwoD & \cmark & \xmark & \Natural & \cmark & rotational motion only \\
Reinbacher \cite{Reinbacher17iccp} & \TwoD & \cmark & \xmark & \Natural & \cmark & rotational motion only \\
Censi \cite{Censi14icra} & \ThreeD & \cmark & \xmark & \BlackWhite & \xmark & attached depth sensor\\
Weikersdorfer \cite{Weikersdorfer14icra} & \ThreeD & \cmark & \cmark & \Natural & \xmark & attached RGB-D sensor\\
Mueggler \cite{Mueggler14iros} & \ThreeD & \cmark & \xmark & \BlackWhite & \cmark & 3D map of lines\\
Gallego \cite{Gallego17pami} & \ThreeD & \cmark & \xmark & \Natural & \xmark & 3D map of the scene
\\
Rebecq \cite{Rebecq18ijcv} & \ThreeD & \xmark & \cmark & \Natural & \cmark & %
pose information
\\
Kueng \cite{Kueng16iros} & \ThreeD & \cmark & \cmark & \Natural & \xmark & intensity images\\
Kim \cite{Kim16eccv} & \ThreeD & \cmark & \cmark & \Natural & \cmark & image reconstruction\\ 
Rebecq \cite{Rebecq17ral} & \ThreeD & \cmark & \cmark  & \Natural & \cmark & $-$\\ 
\bottomrule
\end{tabular}
\end{adjustbox}
\vspace{-1ex}
\end{table}

%% file: chapters/fig_SLAM_TRO.tex
\begin{figure}[t]
\centering
\setlength{\tabcolsep}{2pt}
\begin{tabular}{ll}
\raisebox{.3\height}{\includegraphics[width=0.4\linewidth]{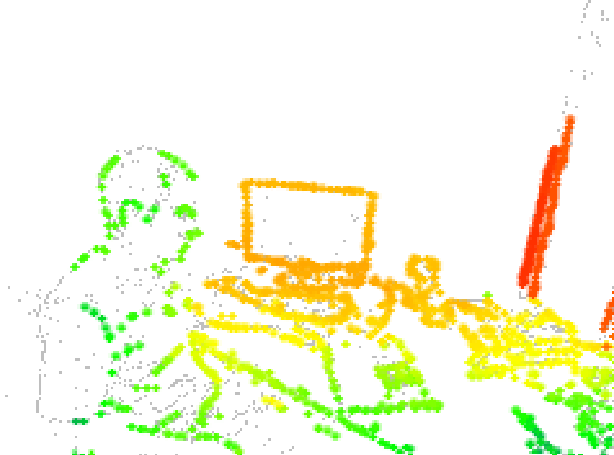}}
\label{fig:slam:dynamic:tracking}
& \includegraphics[trim={1cm 0.1cm 0 1.65cm},clip,width=0.54\linewidth]{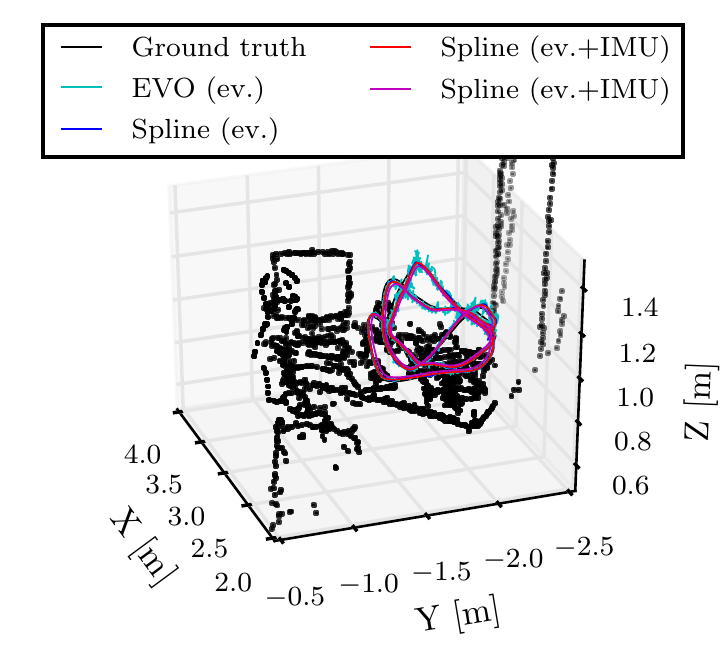}
\label{fig:slam:dynamic:3d}\\[-0.8ex]
{\small (a) Events \& projected map.} 
& {\small (b) Camera trajectory \& 3D map.}\\
\end{tabular}
\vspace{-1.5ex}
\caption{Event-based SLAM.
(a) Reconstructed scene from~\cite{Mueggler17ijrr}, with the reprojected semi-dense map colored according to depth and overlaid on the events (in gray), showing the good alignment between the map and the events.
(b) Estimated camera trajectory (several methods) and semi-dense 3D map (i.e., point cloud).
Image courtesy of~\cite{Mueggler18tro}.}
\label{fig:slam:dynamic}
\vspace{-1ex}
\end{figure}

%% file: chapters/045_visual_inertial.tex
\subsection{Visual-Inertial Odometry (VIO)}
\label{sec:vio}

The robustness of event-based visual odometry and SLAM systems can be improved by combining an event camera with an inertial measurement unit (IMU) rigidly attached.
VIO is qualitatively different from VO: a VO system may lose track (i.e., fail to produce an output pose). 
In contrast, a VIO system does not ``fail'' (there is always an output pose), it drifts.
For this and other reasons, some event cameras have an integrated IMU (see Table~\ref{tab:devices}).

A key issue in VIO is how to temporally fuse data from the synchronous, high-rate IMU (e.g., \SI{1}{\kilo\hertz}) and the asynchronous event camera. 
There are three options in the literature: (i) use an asynchronous probabilistic filter~\cite{Zhu17cvpr},
(ii) use pre-integration theory~\cite{Forster17troOnmanifold} to convert IMU data into lower-rate pieces of information at desired times where events are collected~\cite{Zhu17cvpr,Rebecq17bmvc,Rosinol18ral}
or (iii) consider a continuous-time framework so that both sensor measurements are referred to a common temporal axis that is also used to model the solution (i.e., the camera poses)~\cite{Mueggler18tro}.

\textbf{Feature-based VIO}:
The majority of existing event-based VIO systems are ``feature-based'', consisting of two stages: first, features tracks are extracted from the events (Section~\ref{sec:feature_tracking}), and then these point trajectories on the image plane are fused with IMU data using modern VIO algorithms, such as~\cite{Mourikis07icra,Leutenegger15ijrr,Forster17troOnmanifold}.
That is, a front-end converts the photometric information conveyed by the events into geometric information that is then processed by highly optimized geometric VIO pipelines.
For example, \cite{Zhu17cvpr} tracked features using~\cite{Zhu17icra}, and combined them with IMU data by means of a Kalman filter~\cite{Mourikis07icra}.
Recently, \cite{Rebecq17bmvc} proposed to synthesize motion-compensated event images from spatio-temporal windows of events (Section~\ref{sec:event_processing})
and then detect-and-track features using classical methods~\cite{Rosten06eccv,Lucas81ijcai}.
Feature tracks were fused with inertial data by means of keyframe-based nonlinear optimization~\cite{Leutenegger15ijrr}
to recover the camera trajectory and a sparse map of 3D landmarks.
Later, the nonlinear optimization in~\cite{Rebecq17bmvc} was extended to also fuse intensity frames~\cite{Rosinol18ral},
and was demonstrated on a resource-constrained platform (quadrotor), enabling it to fly in low light and HDR scenarios by exploiting the advantages of event cameras.
The above methods are benchmarked on the 6-DOF motion dataset~\cite{Mueggler17ijrr}, and each method outperforms its predecessor.

\textbf{Reprojection-error--based VIO}:
The work in~\cite{Mueggler18tro} presents a different approach,
fusing events and inertial data using a continuous-time framework~\cite{PatronPerez15ijcv}.
As opposed to the above-mentioned feature-based methods, it optimizes a combined objective with inertial- and event-reprojection error terms over a segment of the camera trajectory, in the style of visual-inertial bundle adjustment.

\textbf{Opportunities}:
The above works adapt state-of-the-art VIO methods for conventional cameras by first converting events into geometric information.
However, it should be possible to avoid this conversion step and directly recover the camera motion and scene structure from the events, as suggested by~\cite{Gallego18cvpr}; 
for example, by optimizing a function with photometric (i.e., event firing rate~\cite{Rebecq17ral}) and inertial error terms, akin to VI-DSO~\cite{vonStumberg18icra} for standard cameras.

Stereo event-based VIO is an unexplored topic, and it would be interesting to see how ideas from event-based depth estimation can be combined with SLAM and VIO.

Also to be explored are learning-based approaches to tackle all of the above problems.
Currently, literature is dominated by model-based methods, but, as it happened in frame-based vision, we anticipate that learning-based methods will also play a major role in event-based processing.
Some works in this direction are~\cite{Lagorce17pami,Maqueda18cvpr,Sironi18cvpr,Zhu19cvpr}.

%% file: chapters/046_image_reconstruction.tex
\subsection{Image Reconstruction}
\label{sec:imagereconstruction}

Events represent brightness changes, and so, in ideal conditions (noise-free scenario, perfect sensor response, etc.) integration of the events yields ``absolute'' brightness.
This is intuitive, since events are just a non-redundant (i.e., ``compressed'') per-pixel way of encoding the visual content in the scene.
Event integration or, more generically, image reconstruction (Fig.~\ref{fig:image_reconstruction}) can be interpreted as ``decompressing'' the visual data encoded in the event stream.
Due to the very high temporal resolution of the events, brightness images can be reconstructed at very high frame rate (e.g., \SIrange{2}{5}{\kilo\Hz}~\cite{Barua16wacv,Rebecq19pami}) or even continuously in~time~\cite{Scheerlinck18accv}.

As the literature reveals, the insight about image reconstruction from events is that it requires regularization. 
Event cameras have independent pixels that report brightness changes, and, consequently, per-pixel integration of such changes during a time interval only produces brightness increment images.
To recover the absolute brightness at the end of the interval, 
an offset image (i.e., the brightness image at the start of the interval) would need to be added to the increment~\cite{Mueggler17ijrr,Brandli14iscas}.
Surprisingly, some works have used spatial and/or temporal smoothing \cite{Reinbacher16bmvc,Barua16wacv,Munda18ijcv,Scheerlinck18accv} to reconstruct brightness starting from a zero initial condition, i.e., without knowledge of the offset image.
Other forms of regularization, using learned features from natural scenes~\cite{Barua16wacv,Rebecq19cvpr,Mostafavi19cvpr,Rebecq19pami} are also effective.

\input{chapters/fig_image_reconstruction.tex}

\textbf{Literature Review}:
Image reconstruction from events was first established in~\cite{Cook11ijcnn} under rotational camera motion and static scene assumptions.
These assumptions together with the brightness constancy \eqref{eq:brightnessIncrementLinearized} were used in a message-passing algorithm between pixels in a network of visual maps to jointly estimate several quantities, such as scene brightness.
Also under the above motion and scene assumptions, \cite{Kim14bmvc} showed how to reconstruct high-resolution panoramas from the events, and they popularized the idea of event-based HDR image reconstruction.
Each pixel of the panoramic image used a Kalman filter to estimate the brightness gradient (based on~\eqref{eq:brightnessIncrementLinearized}), which was then integrated using Poisson reconstruction to yield absolute brightness.
The method in~\cite{Belbachir14cvprw} exploited the constrained motion of a platform rotating around a single axis to reconstruct images that were then used for stereo depth estimation.

Motion restrictions were then replaced by regularizing assumptions to enable image reconstruction for generic motions and scenes \cite{Bardow16cvpr}.
In this work, image brightness and optical flow were simultaneously estimated using a variational framework that contained several penalty terms (on data fitting~\eqref{eq:BrightnessIncrment} and smoothness of the solution) to best explain a space-time volume of events discretized as a voxel grid. 
This method was the first to show reconstructed video from events in dynamic scenes.
Later \cite{Reinbacher16bmvc,Munda18ijcv,Barua16wacv} showed that image reconstruction was possible even without having to estimate motion.
This was done using a variational image denoising approach based on time surfaces~\cite{Reinbacher16bmvc,Munda18ijcv} or using sparse signal processing with a patch-based learned dictionary that mapped events to image gradients, which were then Poisson-integrated \cite{Barua16wacv}.
Concurrently, the VO methods in~\cite{Kim16eccv,Rebecq17ral} extended the image reconstruction technique in~\cite{Kim14bmvc} to 6-DOF camera motions by using the computed scene depth and poses:
\cite{Kim16eccv} used a robust variational regularizer to reduce noise and improve contrast of the reconstructed image,
whereas \cite{Rebecq17ral} showed image reconstruction as an ancillary result, since it was not needed to achieve VO.
Recently, \cite{Scheerlinck18accv} proposed a temporal smoothing filter for image reconstruction and for continuously fusing events and frames. 
The filter acted independently on every pixel, thus showing that no spatial regularization on the image plane was needed to recover brightness, although it naturally reduced noise and artefacts at the expense of sacrificing some real detail.
More recently, \cite{Rebecq19cvpr,Rebecq19pami} has presented a deep learning approach that achieves considerable gains over previous methods and mitigates visual artefacts. 
Reflecting back on earlier works, the motion restrictions or hand-crafted regularizers that enabled image reconstruction have been replaced by perceptual, data-driven priors from natural scenes that consequently produced more natural-looking images.
Note that image reconstruction methods used in VO or SLAM~\cite{Cook11ijcnn,Kim14bmvc,Kim16eccv} assume static scenes, 
whereas methods with weak or no motion assumptions \cite{Bardow16cvpr,Reinbacher16bmvc,Barua16wacv,Munda18ijcv,Scheerlinck18accv,Rebecq19cvpr,Rebecq19pami} are naturally used to reconstruct videos of arbitrary (e.g., dynamic) scenes.

Besides image reconstruction from events, another category of methods tackles the problem of fusing events and frames (e.g., from the DAVIS~\cite{Brandli14ssc}),
thus enhancing the brightness information from the frames with high temporal resolution and HDR properties of events~\cite{Brandli14iscas,Scheerlinck18accv,Pan19cvpr}.
These methods also do not rely on motion knowledge and are ultimately based on \eqref{eq:EventTriggeringCondition}.
The method in \cite{Brandli14iscas} performs direct event integration between frames, pixel-wise.
However, the fused brightness becomes quickly corrupted by event noise (due to non-ideal effects, sensitivity mismatch, missing events, etc.), and so fusion is reset with every incoming frame.
To mitigate noise, events and frames are fused in \cite{Scheerlinck18accv} using a per-pixel, temporal complementary filter that is high-pass in the events and low-pass in the frames.
It is an efficient solution that takes into account the complementary sensing modality of events and frames: 
frames carry slow-varying brightness information (i.e., low temporal frequency), whereas events carry ``change'' information (i.e., high frequency).
The fusion method in~\cite{Pan19cvpr} exploits the high temporal resolution of the events to additionally remove motion blur from the frames, producing high frame-rate, sharp video from a single blurry frame and events.
It is based on a double integral model (one integral to recover brightness and another one to remove blur) within an optimization framework.
A limitation of the above methods is that they still suffer from artefacts due to event noise. 
These might be mitigated if combined with learning-based approaches~\cite{Rebecq19pami}.

\iflongversion
\textbf{Image quality}:
The quality of the reconstructed image is directly affected by noise in the contrast threshold (Section~\ref{subsec:event_generation_model}),
which changes per pixel (due to manufacturing mismatch) and also due to dynamical effects (incident light, time, etc.)~\cite{Brandli14iscas}.
Image quality has been quantified in several works~\cite{Munda18ijcv,Scheerlinck18accv,Mostafavi19cvpr,Rebecq19cvpr,Rebecq19pami} 
and is also affected by the spatial resolution of the sensor.
\fi

\textbf{Applications}:
Image reconstruction implies that, in principle, it is possible to convert the events into brightness images and then apply mature computer vision algorithms~\cite{Rebecq19cvpr,Rebecq19pami,Scheerlinck19cvprw}.
This can have a high impact on both, event- and frame-based communities.
The resulting images capture high-speed motions and HDR scenes, which may be beneficial in some applications, 
but it comes at the expense of computational cost, latency and power consumption.

Despite image reconstruction having been useful to support tasks such as recognition~\cite{Barua16wacv}, SLAM~\cite{Kim16eccv} or optical flow estimation~\cite{Bardow16cvpr},
there are also works in the literature, such as~\cite{Gallego18cvpr,Sironi18cvpr,Ye18arxiv,Zhu19cvpr}, showing that it is not needed to fulfill such tasks.
One of the most valuable aspects of image reconstruction is that it provides scene representations (e.g., appearance maps~\cite{Kim14bmvc,Gallego17pami}) that are more \emph{invariant} to motion than events and also facilitate establishing event correspondences, which is one of the biggest challenges of some event data processing tasks, such as feature tracking~\cite{Gehrig19ijcv}.

%% file: chapters/fig_image_reconstruction.tex
\iffalse
\begin{figure}[t]
\centering
\includegraphics[trim={0 2cm 0 2cm},clip,width=\columnwidth]{images/hdr_comparison.pdf}
\caption{Image reconstruction example.
Camera pointing at the Sun, in front of a traffic sign.
Left: view from a standard camera, showing severe under-exposure on the foreground.
Middle: frame from the DAVIS~\cite{Brandli14ssc}, showing severe under- and over-exposed areas.
Right: HDR image reconstructed from the events.
Image courtesy of~\cite{Rebecq17ral}.
}
\label{fig:image_reconstruction}
\vspace{-1ex}
\end{figure}

\global\long\def\heighttunnel{2.15cm}
\begin{figure}[t]
\centering
\includegraphics[height=\heighttunnel]{images/rpg/tunnel/events.png} 
\includegraphics[height=\heighttunnel]{images/rpg/tunnel/HUAWEI.png} 
\includegraphics[height=\heighttunnel]{images/rpg/tunnel/E2VID.png}
\caption{Image Reconstruction example.
Car driving out of a tunnel.
The frames from the consumer camera (Huawei P20 Pro) (Middle) suffer from under- or over-exposure, 
while the events (Left) capture a broader dynamic range of the scene, 
which is recovered by image reconstruction methods (Right).
Images courtesy of~\cite{Rebecq19pami}.
}
\label{fig:image_reconstruction}
\vspace{-1ex}
\end{figure}
\fi

\global\long\def\heighttunnel{2.15cm}
\begin{figure}[t]
\centering
\subfloat[Tunnel scene]{\includegraphics[height=\heighttunnel]{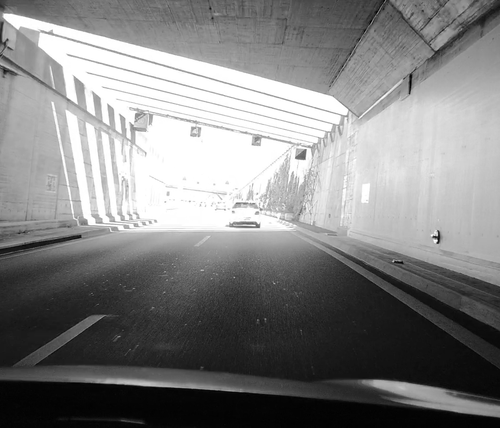} 
\includegraphics[trim={0cm 0cm 2cm 0cm},clip,height=\heighttunnel]{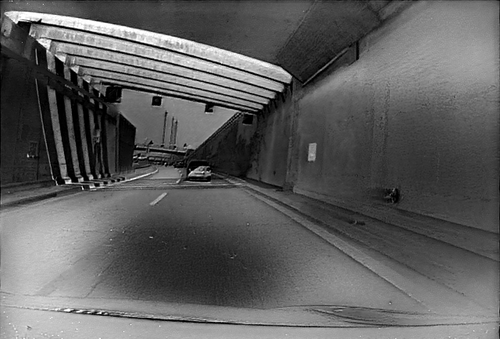}}\;\;\;
\subfloat[Exploding mug]{\includegraphics[height=\heighttunnel]{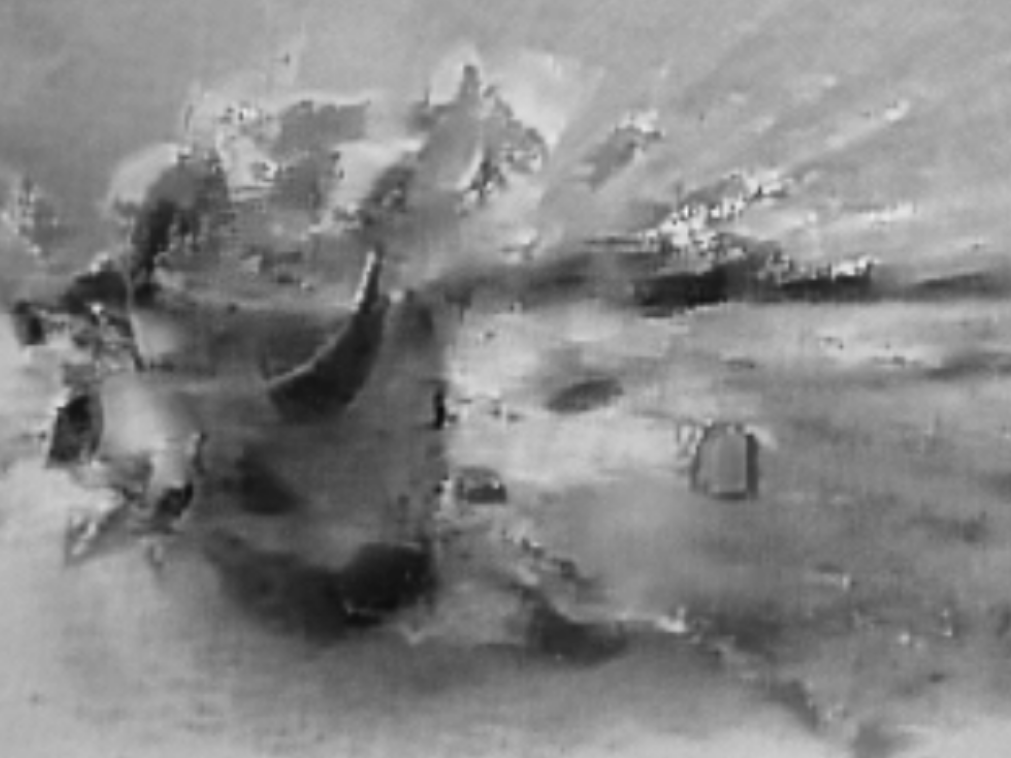}}
\caption{Image Reconstruction.
In the scenario of a car driving out of a tunnel the frames from a consumer camera (Huawei P20 Pro) (Left) suffer from under- or over-exposure, 
while events capture a broader dynamic range of the scene, which is recovered by image reconstruction methods (Middle).
Events also enable the reconstruction of high-speed scenes, such as a exploding mug (Right).
Images courtesy of~\cite{Rebecq19pami,Scheerlinck20wacv}.}
\label{fig:image_reconstruction}
\end{figure}

%% file: chapters/047_segmentation.tex
\subsection{Motion Segmentation}
\label{sec:motionsegmentation}

Segmentation of moving objects viewed by a stationary event camera is simple because events are solely imputable to the motion of the objects (assuming constant illumination) \cite{Litzenberger06dspws,Lagorce15tnnls,Ni15neco}.
The challenges arise in the scenario of a moving camera because events are triggered everywhere on the image plane~\cite{Glover16iros,Mitrokhin18iros,Stoffregen19iccv} (Fig.~\ref{fig:motionsegmentation}), produced by moving objects and the static scene (whose apparent motion is induced by the camera's ego-motion) and the goal is to infer this causal classification for each event.
However, each event carries very little information, and therefore it is challenging to perform the mentioned per-event classification.

\input{chapters/fig_segmentation_iCub.tex}

Overcoming these challenges has been done by tackling segmentation scenarios of increasing complexity, 
obtained by reducing the amount of additional information given to solve the problem.
Such additional information adopts the form of known object shape or known motion, 
i.e., the algorithm knows ``what object to look for'' or ``what type of motion it expects'' and objects are segmented by detecting (in-)consistency with respect to the expectation.
The less additional information is provided, the more unsupervised the problem becomes (e.g., clustering).
In such a case, segmentation is enabled by the key insight that moving objects produce distinctive traces of events on the image plane and it is possible to infer the trajectories of the objects that generate those traces, yielding the segmented objects~\cite{Stoffregen19iccv}. 
Like clustering, this is a joint optimization problem in the motion parameters of the objects (i.e., the ``clusters'') and the event-object associations (i.e., the segmentation).

\textbf{Literature Review}:
Considering known object shape, \cite{Glover16iros} presents a method to detect and track a circle in the presence of event clutter caused by the moving camera.
It is based on the Hough transform using optical flow information extracted from temporal windows of events.
The method was extended in~\cite{Glover17iros} using a particle filter to improve tracking robustness:
\cb{the duration of the observation window was dynamically selected to accommodate for sudden motion changes due to accelerations of the object.}
More generic object shapes were detected and tracked by~\cite{Vasco17icar} using event corners (Section~\ref{sec:feature_tracking}) as geometric primitives. 
In this method, additional knowledge of the robot joints controlling the camera motion was required.

Segmentation has been addressed by \cite{Stoffregen17acra,Mitrokhin18iros,Stoffregen19iccv} under mild assumptions leveraging the idea of motion-compensated event images~\cite{Gallego17ral} (Section~\ref{sec:event_processing}).
Essentially this technique associates events that produce sharp edges when warped according to a motion hypothesis.
The simplest hypothesis is a linear motion model (i.e., constant optical flow), 
yet it is sufficiently expressive: for short times, scenes may be described as collections of objects producing events that fit different linear motion models.
Such a scene description is what the cited segmentation algorithms seek for.
Specifically, the method in \cite{Stoffregen17acra} first fits a linear motion-compensation model to the dominant events, then removes these and fits another linear model to the remaining events, greedily.
Thus, it clusters events according to optical flow, yielding motion-compensated images with sharp object contours.
Similarly, \cite{Mitrokhin18iros} detects moving objects in clutter by fitting a motion-compensation model to the dominant events (i.e., the background) and detecting inconsistencies with respect to it (i.e., the objects).
They test the method in challenging scenarios inaccessible to standard cameras (HDR, high-speed) and release a dataset.
The work in~\cite{Stoffregen19iccv} proposes an iterative clustering algorithm that jointly estimates the event-object associations (i.e., segmentation) and the motion parameters of the objects (i.e., clusters) that produce sharpest motion-compensated event images.
It allows for general parametric motion models~\cite{Gallego18cvpr} to describe each object
and produces better results than greedy methods~\cite{Mitrokhin18iros,Stoffregen17acra}.
In~\cite{Mitrokhin19iros} a learning-based approach for segmentation using motion-compensation is proposed: 
ANNs are used to estimate depth, ego-motion, segmentation masks of independently moving objects and object 3D velocities.
An event-based dataset is provided for supervised learning, which includes accurate pixel-wise motion masks of 3D-scanned objects that are reliable even in poor lighting conditions and during fast motion.

Segmentation is a paramount topic in frame-based vision, yet it is rather unexplored in event-based vision.
As more complex scenes are addressed and more advanced event-based vision techniques are developed, 
more works targeting this challenging problem are expected to appear.

%% file: chapters/fig_segmentation_iCub.tex
\begin{figure}[t]
\centering
\includegraphics[width=\linewidth]{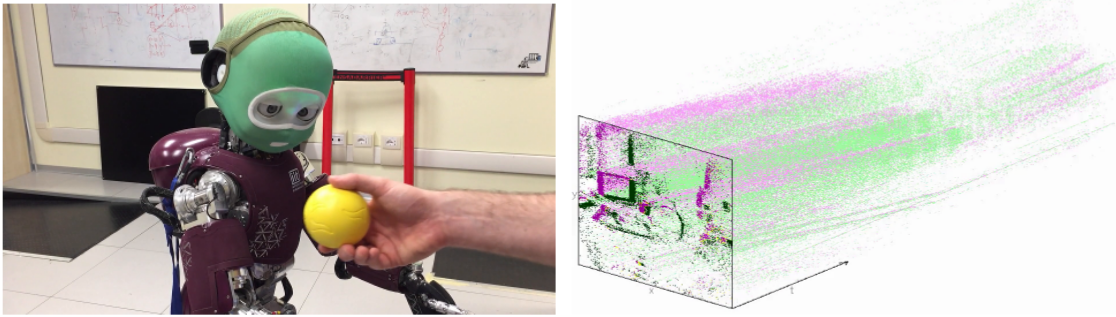}
\caption{The iCub humanoid robot from IIT has two event cameras in the eyes.
Here, it segments and tracks a ball under event clutter produced by the motion of the head.
Right: space-time visualization of the events on the image frame, colored according to polarity (positive in green, negative in red).
Image courtesy of~\cite{Glover17iros}.
\label{fig:motionsegmentation}}
\end{figure}

%% file: chapters/048_recognition.tex
\subsection{Recognition}
\label{sec:recognition}

\textbf{Algorithms}:
Recognition algorithms for event cameras have grown in complexity,
from template matching of simple shapes to 
classifying arbitrary edge patterns using either traditional machine learning
on hand-crafted features or modern deep learning methods.
This evolution aims at endowing recognition systems with more expressibility (i.e., approximation capacity) and robustness to data distortions.

Early research with event-based sensors began with tracking a moving object using a static sensor.
An event-driven update of the position of a model of the object shape was used to detect and track objects with a known simple shape, such as a  blob~\cite{Delbruck07iscas}, circle~\cite{Serrano-Gotarredona09tnn,Wiesmann12cvprw} or line~\cite{Conradt09iscas}.
Simple shapes can also be detected by matching against a predefined template, which removes the need to describe the geometry of the object.
This \emph{template matching} approach was implemented using convolutions in early hardware~\cite{Serrano-Gotarredona09tnn}.

For more complex objects, templates can be used to match low level features instead of the entire object, after which a \emph{classifier} can be used to make a decision based on the distribution of features observed~\cite{Lagorce17pami}.
Nearest Neighbor classifiers are typically used, with distances calculated in feature space.
Accuracy can be improved by increasing feature invariance, which can be achieved using a hierarchical model where feature complexity increases in each layer.
With a good choice of features, only the final classifier needs to be retrained when switching tasks. 
This leads to the problem of selecting which features to use.
Hand-crafted orientation features were used in early works, but far better results are obtained by learning the features from the data itself.
In the simplest case, each template can be obtained from an individual sample, but such templates are sensitive to noise in the sample data~\cite{Orchard15pami}.
One may follow a generative approach, learning features that enable to accurately reconstruct the input, as was done in~\cite{OConnor13fns} with a Deep Belief Network (DBN).
More recent work obtains features by unsupervised learning, clustering the event data and using the center of each cluster as a feature~\cite{Lagorce17pami}.
During inference, each event is associated to its closest feature, 
and a classifier operates on the distributions of features observed.
With the rise of \emph{deep learning} in frame-based computer vision, many have sought to leverage deep learning tools for event-based recognition, using back-propagation to learn features.
This approach has the advantage of not requiring a separate classifier at the output, but the disadvantage of requiring far more labeled data for training.
Image recognition with events also suffers from the practical problem of the availability of training data in the event domain~\cite{Gehrig20cvpr}.
In \cite{Frazzoli-RSS-19} the authors use \emph{wormhole learning}, a semi-supervised approach in which, starting from a detector in the RGB domain, one is able to train a detector in the event domain; moreover, in a second phase the teacher becomes the student, and some of the illumination invariance of the event sensor is transferred to the RGB-only detector.

\input{chapters/fig_recognition.tex}

Most learning-based approaches convert events/spikes into (dense) tensors, a convenient representation for image-based hierarchical models, e.g., ANNs (Fig~\ref{fig:recognition:TN_DAVIS}).
There are different ways the value of each tensor element can be computed
(Section~\ref{sec:representations}).
Simple methods use the time surfaces, or event histogram frames.
A more robust method uses time surfaces with exponential decay~\cite{Lagorce17pami} or with average timestamps~\cite{Sironi18cvpr}.
Image reconstruction methods (Section~\ref{sec:imagereconstruction}) may also be used.
Some recognition approaches rely on converting spikes to frames during inference~\cite{Moeys16ebccsp,Barua16wacv}, while others convert the trained ANN to an SNN which can operate directly on the event data~\cite{PerezCarrasco13pami}.
Similar ideas can be applied for tasks other than recognition~\cite{Maqueda18cvpr,Zhu18rss}.
As neuromorphic hardware advances (Section~\ref{sec:hardware}), there is increasing interest in learning directly in SNNs~\cite{Lee16fns} or even directly in the neuromorphic hardware itself~\cite{neftci18iscience}.

\textbf{Tasks}:
Early tasks focused on detecting the presence of a simple shape (such as a circle) from a static sensor~\cite{Delbruck07iscas,Wiesmann12cvprw,Serrano-Gotarredona09tnn}, but soon progressed to the classification of more complex shapes, such as card pips \cite{PerezCarrasco13pami} (Fig.~\ref{fig:recognition:poker_dvs}), block letters~\cite{Orchard15pami} and faces~\cite{Lagorce17pami,Barua16wacv}.
A popular task throughout has been the classification of hand-written digits. Inspired by the role it has played in frame-based computer vision, a few event-based MNIST datasets have been generated from the original MNIST dataset~\cite{Orchard15fns,Serrano-Gotarredona13ijssc}.
These datasets remain a good test for algorithm development, with many algorithms now achieving over \SI{98}{\percent} accuracy on the task \cite{Lee16fns,Neil16iscas,Wu18fns,Yousefzadeh18tbcas,Shrestha18nips,Sironi18cvpr}, but few would propose digit recognition as a strength of event-based vision.
More difficult tasks involve either more difficult objects, such as the Caltech-101 and Caltech-256 datasets (both of which are still considered easy by computer vision) or more difficult scenarios, such as recognition from on-board a moving vehicle~\cite{Sironi18cvpr}.
Very few works tackle these tasks so far, and those that do typically fall back on generating event frames and processing them using a traditional deep learning framework.

A key challenge for recognition is that event cameras respond to relative motion in the scene (Section~\ref{subsec:challenges_paradigm_shift}), and thus require either the object or the camera to be moving.
It is therefore unlikely that event cameras will be a strong choice for recognizing static or slow moving objects, 
although little has been done to combine the advantages of frame- and event-based cameras for these applications.
The event-based appearance of an object is highly dependent on the above-mentioned relative motion (Fig.~\ref{fig:dataassoc}), thus tight control of the camera motion could be used to aid recognition~\cite{Orchard15fns}.

Since the camera responds to dynamic signals, obvious applications include recognizing objects by the way they move~\cite{Clady17fns}, or recognizing dynamic scenes such as gestures or actions~\cite{Lee14tnnls,Amir17cvpr}.
These tasks are typically more challenging than static object recognition because they include a time dimension, but this is exactly where event cameras excel.

\textbf{Opportunities}:
Event cameras exhibit many alluring properties, but event-based recognition has a long way to go if it is to compete with modern frame-based approaches.
While it is important to compare event- and frame-based methods, one must remember that each sensor has its own strengths.
The ideal acquisition scenario for a frame-based sensor consists of both the sensor and object being static, which is the worst possible scenario for event cameras.
For event-based recognition to find widespread adoption, it will need to find applications which play to its strengths.
Such applications are unlikely to be similar to well established computer vision recognition tasks which play to the frame-based sensor's strengths.
Instead, such applications are likely to involve resource constrained recognition of dynamic sequences, or recognition from on-board a moving platform.
Finding and demonstrating the use of event-based sensors in such applications remains an open challenge. %

Although event-based datasets have improved in quality in recent years, there is still room for improvement.
\iflongversion 
Much more data is being collected, but annotation remains challenging.
There is not yet an agreed upon or standard tool or format for annotations.
Many event-based datasets are derived from frame-based vision.
While these datasets have played an important role in the field, they inherently play to the strengths of frame-based vision and are thus unlikely to give rise to new event-based sensor applications.
\fi
Data collection and annotation is a tiresome and thankless task, but developing an easy-to-use pipeline for collecting and annotating event-based data would be a significant contribution to the field, especially if the tools can mature to the stage where the task can be outsourced to laymen.

%% file: chapters/fig_recognition.tex
\begin{figure}
\centering
\global\long\def\heightfigrecog{2.7cm}
\subfloat[Event camera and IBM TrueNorth]{\includegraphics[height=\heightfigrecog]{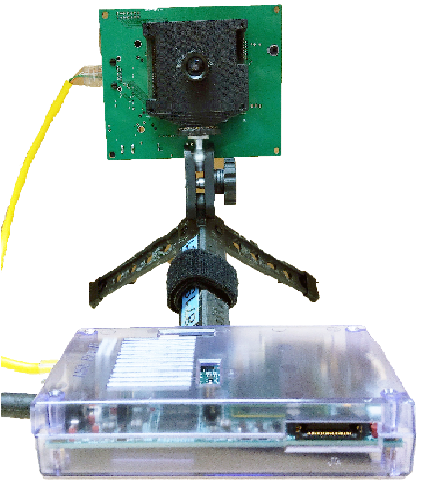}\label{fig:recognition:TN_davis_system}
  \includegraphics[trim={5cm 0cm 0cm 1cm},clip,height=\heightfigrecog]{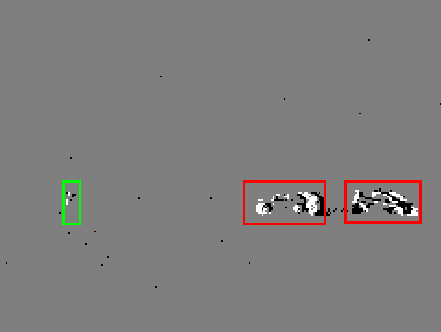}\label{fig:recognition:TN_davis_result}}\;\;
\subfloat[Poker-DVS]{\includegraphics[height=\heightfigrecog]{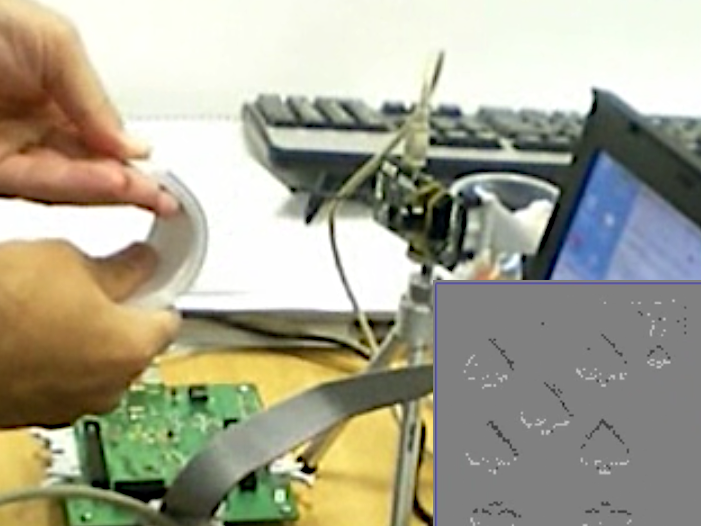}\label{fig:recognition:poker_dvs}} 
\caption{Recognition of moving objects. 
(a) A DAVIS240C sensor with FPGA attached tracks and sends regions of events to IBM's TrueNorth NS1e evaluation platform for classification.
Results on a street scene show red boxes around tracked and classified cars.\label{fig:recognition:TN_DAVIS}
(b) In~\cite{PerezCarrasco13pami} very high speed object recognition (browsing a full deck of 52 cards in just \SI{0.65}{\second}) was illustrated with event-driven convolutional neural networks.}
\end{figure}

%% file: chapters/049_control.tex
\subsection{Neuromorphic Control}
\label{sec:control}
In living creatures, most information processing happens through spike-based representation: 
spikes encode the sensory data; spikes perform the computation; and spikes transmit actuator ``commands''. 
Therefore, biology shows that the event-based paradigm is, in principle, applicable not just to perception and inference, but also to control.

\input{chapters/fig_control}

\textbf{Neuromorphic-vision-driven Control Architecture}:
In this type of architecture (Fig.~\ref{fig:arch}), there is a neuromorphic sensor, an event-based estimator, and a traditional controller. 
The estimator computes a state, and the controller computes the control based on the provided state. 
The controller is not aware of the asynchronicity of the architecture.

Neuromorphic-vision-driven control architectures have been demonstrated since the early days of neuromorphic cameras, and they have proved the two advantages of low latency and computational efficiency.
The earliest demonstrators were the spike-based convolutional target tracking demo in the CAVIAR project~\cite{Serrano-Gotarredona09tnn} and
the ``robot goalie'' described in \cite{Delbruck07iscas,Delbruck13fns}. 
Another early example was the pencil-balancing robot~\cite{Conradt09iscas}.
In that demonstrator two DVS's observed a pencil as inverted pendulum placed on a small movable cart. 
The pencil's state in 3D was estimated in below \SI{1}{\milli\second} latency. 
A simple hand tuned PID controller kept the pencil balanced upright.
It was also demonstrated on an embedded system, thereby establishing the ability to run on severely constrained computing resources.

\textbf{Event-based Control Theory}:\label{sub:control}
Event-based techniques can be motivated from the perspective of control and decision theory. 
Using a biological metaphor, event-based control can be understood as a form of what economics calls \emph{rational inattention}~\cite{sim03}: 
more information allows for better decisions, but if there are costs associated to obtaining or processing the information, it is rational to take decisions with only partial information available.

In event-based control, the control signal is changed asynchronously~\cite{miskowicz2018event}. 
There are several variations of the concept depending on how the ``control events'' are generated.
One important distinction is between \emph{event-triggered control} and
\emph{self-triggered control}~\cite{Tabuada2012}. 
In \emph{event-based control} the events are generated
``exogenously'' based on certain condition; 
for example, a ``recompute control'' request might be triggered when the trajectory's tracking error exceeds a threshold.
In \emph{self-triggered control}, the controller decides by itself when is the
next time it should be called based on the situation. 
For example, a controller might decide to ``sleep'' for longer if the state is near the target, or to recompute the control signal sooner if it is required. 

The advantages of event-based control are usually justified considering a trade-off between computation / communication cost and control performance.
The basic consideration is that, while the best control performance is obtained by recomputing the control infinitely often (for an infinite cost), there are strongly diminishing returns.
A solid principle of control theory is that the control frequency depends on the time constant of the plant and the sensor: 
it does not make sense to change the control much quicker than the new incoming information or the speed of the actuators.
This motivates choosing control frequencies that are comparable with the plant dynamics and adapt to the situation.
For example, one can show that an event-triggered controller achieves the same performance with a fraction of the computation; or, conversely, a better performance with the same amount of computation. 
In some cases (scalar linear Gaussian) these trade-offs can be obtained in closed form~\cite{Astrom2008,Wang2014}. 
(Analogously, certain trade-offs can be obtained in closed form for perception~\cite{CensiMFS15}.)

Unfortunately, the large literature in event-based control is of restricted utility for the embodied neuromorphic setting. 
Beyond the superficial similarity of dealing with ``events'' the settings are quite different. 
For example, in network-based control, one deals with typically low-dimensional states and occasional events---the focus is on making the most of each single event. 
By contrast, for an autonomous vehicle equipped with event cameras, the problem is typically how to find useful signals in potentially millions of events per second.
Particularizing the event-based control theory to the neuromorphic case is a relatively young avenue of research~\cite{Mueller2015,Singh2016,Censi2015,Mueller2015a}.
The challenges lie in handling the non-linearities typical of the vision modality, which prevents clean closed-form results.

\textbf{Open questions in Neuromorphic Control}:
Finally, we describe some of open problems in this topic.

\emph{Task-driven sensing}:
In animals, perception has value because it is followed by action, and the information collected is \emph{actionable information} that helps with the task.
A significant advance would be the ability for a controller to modulate the sensing process based on the task and the context. 
In current hardware there is limited software-modulated control for the sensing processing, though it is possible to modulate some of the hardware biases.
Integration with region-of-interest mechanisms, heterogeneous camera bias settings, etc. would provide additional flexibility and more computationally efficient control.

\emph{Thinking fast and slow}:
Existing research has focused on obtaining low-latency control, but there has been little work on how to integrate this sensorimotor level into the rest of an agent's cognitive architecture. 
Using again a bio-inspired metaphor, and following Kahneman~\cite{kahneman2011thinking}, the fast/instinctive/``emotional'' system must be integrated with the slower/deliberative system.

%% file: chapters/fig_control.tex
\begin{figure}[t]
\centering
\subfloat[]{\includegraphics[scale=0.68]{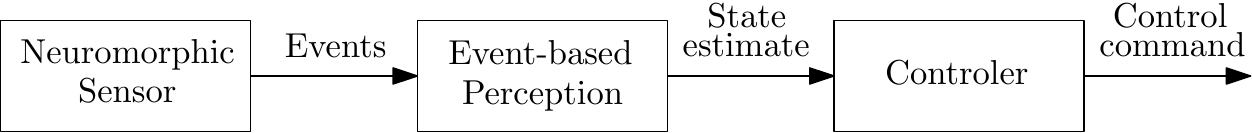}}\\%[-0.1ex]
\subfloat[]{\includegraphics[scale=0.68]{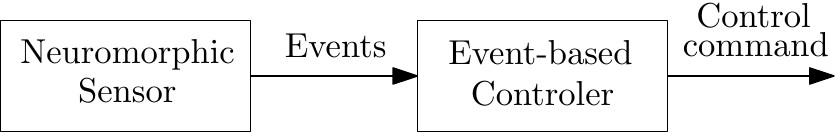}}\\%[-0.1ex]
\subfloat[]{\includegraphics[scale=0.68]{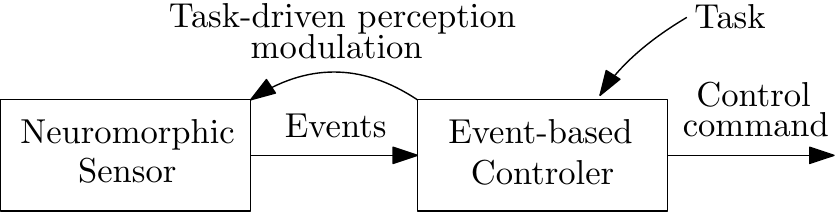}}

\caption{\label{fig:arch}
Control architectures based on neuromorphic events.
In a neuromorphic-vision-driven control architecture (a), a neuromorphic sensor produces events, an event-based perception system produces state estimates, and a traditional controller is called asynchronously to compute the control signal. 
In a native neuromorphic-based architecture (b), the events generate directly changes in control. 
Finally, (c) shows an architecture in which the task informs the events that are generated.}
\end{figure}

%% file: chapters/05_systems_demonstrators.tex
\section{Event-based Systems and Applications}
\label{sec:systems_and_demonstrators}

\input{chapters/051_neuromorphic_hw.tex}

\subsection{Applications in Real-Time On-Board Robotics}
\label{sec:embedded_robotics}
\input{chapters/fig_embedded_appls.tex}

As event-based vision sensors often produce significantly less data per time interval compared to traditional cameras, multiple applications can be envisioned where extracting relevant vision information can happen in real-time within a simple computing system directly connected to the sensor, avoiding USB connection. 
Fig~\ref{fig:aplications:eDVS128_Pushbot} shows an example of such, where a dual-core ARM micro controller running at \SI{200}{\mega\hertz} with \SI{136}{\kilo\byte} on-board SRAM fetches and processes events in real-time. 
The combined embedded system of sensor and micro controller here operate a simple wheeled robot in tasks such as line following, active and passive object tracking, distance estimation, and simple mapping~\cite{Waniek15robio}.

A different example of near-sensor processing (``edge computing'') is the Speck SoC\footnote{\url{https://www.speck.ai/}}, which combines a DVS and the Dynap-se neuromorphic CNN processor. 
Its peak power consumption is less than \SI{1}{\milli\watt} and latency is less than \SI{30}{\milli\second}.
Application domains are low-power, continuous object detection, surveillance, and automotive systems.

Event cameras have also been used on-board quadrotors with limited computational resources, both for autonomous landing~\cite{Hordijk17jfr} or flight~\cite{Rosinol18ral} (Fig.~\ref{fig:aplications:USLAM_drone}), in challenging scenes.

%% file: chapters/051_neuromorphic_hw.tex
\subsection{Neuromorphic Computing}
\label{sec:hardware}
Neuromorphic engineering tries to capture some of the unparalleled computational power and efficiency of the brain by mimicking its structure and function. 
Typically this results in a massively parallel hardware accelerator for SNNs (Section~\ref{sec:bioinspired_processing}), which is how we will define a neuromorphic processor. 
Since the neuron spikes within such a processor are inherently asynchronous, a neuromorphic processor is the best computational partner for an event camera.
Neuromorphic processors act on the events injected by the event camera directly, without conversion, and offer better data-processing locality (spatially and temporally) than standard architectures such as CPUs,
yielding low power and low latency computer vision systems.

Neuromorphic processors may be categorized by their neuron model implementations (Table~\ref{tab:processors}), which are broadly divided between analog neurons (Neurogrid, BrainScaleS, ROLLS, \mbox{DYNAP-se}), 
digital neurons (TrueNorth, Loihi, ODIN) 
and software neurons (SpiNNaker). 
Some architectures also support on-chip learning (Loihi, ODIN, \mbox{DYNAP-le}).
When evaluating a neuromorphic processor for an event-based vision system, the following criteria should be considered in addition to the processor's functionality and performance:
(\emph{i})
the software development ecosystem: a minimal toolchain includes an API to compose and train a network, 
a compiler to prepare the network for the hardware, 
and a runtime library to deploy the network in hardware,
(\emph{ii})
event-based vision systems typically require that a processor be available as a standalone system suitable for mobile applications, and not just hosted in a remote server,
(\emph{iii})
the availability of neuromorphic processors.

Several developments are necessary to enable a more widespread use of these processors, such as:
(\emph{i}) developing a more user-friendly ecosystem (an easier way to program the desired method for deployment in hardware), 
(\emph{ii}) enabling more processing capabilities of the hardware platform, 
(\emph{iii}) increasing the availability of devices beyond early access programs targeted at selected partners.

\input{chapters/table_processors.tex}

\label{sec:processorarchitectures}
The following processors (Table~\ref{tab:processors}) have the most mature developer workflows, combined with the widest availability of standalone systems.
More details are given in \cite{CamunasMesa19ma,Rajendran19msp}.

\textbf{SpiNNaker (Spiking Neural Network Architecture)}
uses general-purpose ARM cores to simulate biologically realistic models of the human brain~\cite{Furber14ieee}. 
\iflongversion Unlike the other processors that choose specific simplified neuron models to embed in custom transistor circuits, \fi
SpiNNaker implements neurons as software running on the cores, sacrificing hardware acceleration to maximize model flexibility.
The SpiNNaker has been coupled with event cameras for stereo depth estimation~\cite{Dikov17cbbs,Haessig19aicas},
optic flow computation~\cite{Richter14bc,Haessig19aicas}, and for object tracking \cite{Glover18irosw} and recognition~\cite{Serrano-Gotarredona15iscas}.

\textbf{TrueNorth}
uses digital neurons to perform real-time inference. 
Each chip simulates \SI{1}{\mega\nothing} (million) neurons and \SI{256}{\mega\nothing} synapses, distributed among 4096 neurosynaptic cores. 
There is no on-chip learning, so networks are trained offline using a GPU or other processor~\cite{Merolla14science}.

TrueNorth has been paired with event cameras to produce end-to-end, low power and low-latency event-based vision systems for gesture recognition~\cite{Amir17cvpr}, stereo reconstruction~\cite{Andreopoulos18cvpr} and optical flow estimation~\cite{Haessig18tbcas}.

\textbf{Loihi}
uses digital neurons to perform real-time inference and online learning. 
Each chip simulates up to 131 thousand spiking neurons and \SI{130}{\mega\nothing} synapses.
A learning engine in each neuromorphic core updates each synapse using rules that includes STDP and reinforcement learning~\cite{Davies18micro}.
Non-spiking networks can be trained in TensorFlow and approximated by spiking networks for Loihi using the Nengo Deep Learning toolkit from Applied Brain Research~\cite{Blouw18arxiv}.
\iflongversion
There are plans to build a larger system called Pohoiki Springs with 768 Loihi chips (\SI{100}{\mega\nothing} neurons and 100 billion synapses),
which is still small compared to the human brain's 800 trillion synapses.
\fi

\textbf{DYNAP}: The Dynamic Neuromorphic Asynchronous Processor 
\iflongversion (DYNAP) from aiCTX \fi 
has two variants, one optimized for scalable inference (Dynap-se), and another for online learning (Dynap-le).

\textbf{Braindrop} prototypes a single core of the \SI{1}{\mega\nothing}-neuron Brainstorm system~\cite{Neckar18thesis}.
It is programmed using Nengo~\cite{Bekolay14fni} and implements the Neural Engineering Framework~\cite{Eliasmith12science}.
\iflongversion Braindrop is Stanford University's follow-on to the Neurogrid processor. \fi

%% file: chapters/table_processors.tex
\begin{table}[t!]
\centering
\caption{\label{tab:processors}Comparison between selected neuromorphic processors,\newline ordered by neuron model type.}
\vspace{-1ex}
\begin{adjustbox}{max width=\linewidth}
\setlength{\tabcolsep}{2pt}
\begin{tabular}{lccccc}
\toprule
Processor & SpiNNaker & TrueNorth & Loihi & DYNAP & Braindrop\tabularnewline
Reference & \cite{Furber13c} & \cite{Akopyan15tcad} & \cite{Davies18micro} & \cite{Moradi18tbcas} & \cite{Neckar18thesis}\tabularnewline
\midrule
Manufacturer & U. Manchester & IBM & Intel & aiCTX & Stanford U.\tabularnewline
Year & 2011 & 2014 & 2018 & 2017 & 2018\tabularnewline
Neuron model & Software & Digital & Digital & Analog & Analog\tabularnewline
On-chip learning & Yes & No & Yes & No & No\tabularnewline
CMOS technol. & \SI{130}{\nano\meter} & \SI{28}{\nano\meter} & \SI{14}{\nano\meter} & \SI{180}{\nano\meter} & \SI{28}{\nano\meter}\tabularnewline
Neurons/chip & \SI{4}{\kilo\nothing}* & \SI{1024}{\kilo\nothing} & \SI{131}{\kilo\nothing} & \SI{1}{\kilo\nothing} & \SI{4}{\kilo\nothing}\tabularnewline
Neurons/core & 255* & 256 & 1024 & 256 & 4096\tabularnewline
Cores/chip & 16* & 4096 & 128 & 4 & 1\tabularnewline
Synapses/chip & 16 M & 268 M & 130 M & \SI{128}{\kilo\nothing}
& 16 M
\tabularnewline
Boards & 4- or 48-chip & 1- or 16-chip & 4- or 8-chip, & 1-chip & 1-chip\tabularnewline
Software stack & sPyNNaker & CPE/Eedn & Nengo & cAER & Nengo\tabularnewline
 & PACMAN & NSCP & Nx SDK & libcAER & \tabularnewline
\bottomrule
\end{tabular}
\end{adjustbox}
\vspace{-1ex}
\end{table}

%% file: chapters/fig_embedded_appls.tex
\begin{figure}
\centering
\global\long\def\heightplotrobots{3.3cm}
\subfloat[\label{fig:aplications:eDVS128_Pushbot.jpg}]{\includegraphics[height=\heightplotrobots]{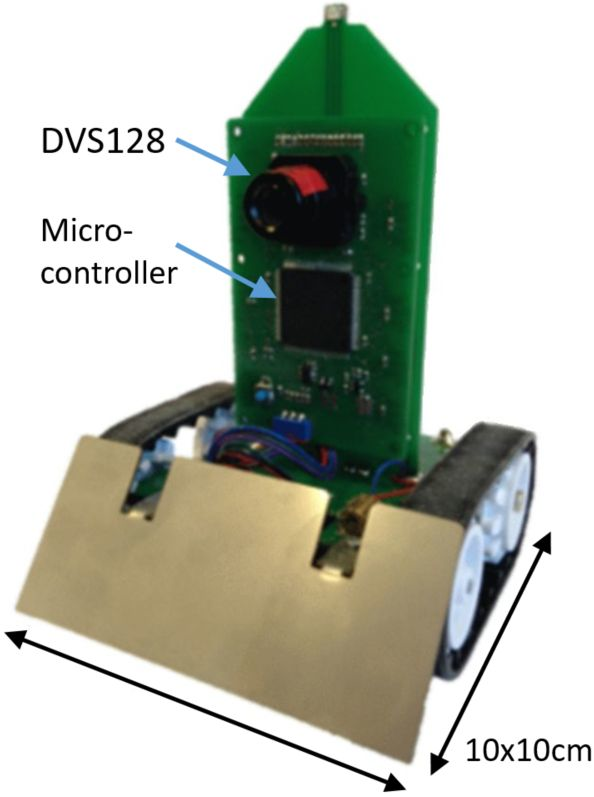}}\;\;
\subfloat[\label{fig:aplications:USLAM_drone}]{\includegraphics[height=\heightplotrobots]{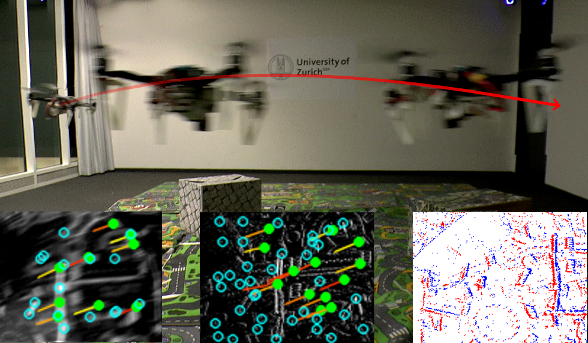}}
\caption{(a) Embedded DVS128 on Pushbot as standalone closed-loop perception-computation-action system, used in navigation and obstacle-avoidance tasks~\cite{Waniek15robio}.
(b) Drone with a down-looking DAVIS, used for autonomous flight~\cite{Rosinol18ral}.
The high speed and dynamic range of events are leveraged to operate in difficult illumination conditions.
The same visual-inertial odometry algorithm~\cite{Rosinol18ral} is also demonstrated on high-speed scenarios, 
such as an event camera spinning tied to a rope.
\label{fig:aplications:eDVS128_Pushbot}}
\end{figure}

%% file: chapters/06_resources.tex
\section{Resources}
\label{sec:resources}
The List of Event-based Vision Resources~\cite{Gallego17resources} is a collaborative effort to collect relevant resources in the field: links to information (papers, videos, organizations, companies and workshops) as well as software, drivers, code, datasets, simulators and other essential tools in event-based vision.

\vspace{-.5ex}
\input{chapters/061_software.tex}

\vspace{-.5ex}
\input{chapters/062_datasets_and_simulators.tex}
\vspace{-1ex}
\input{chapters/063_workshops.tex}

%% file: chapters/061_software.tex
\subsection{Software}
\label{sec:software}
To date, there is no open-source standard library integrated to OpenCV that provides algorithms for event-based vision.
This would be a very desirable resource to accelerate the adoption of event cameras.
There are, however, many quite highly developed open-source software resources:

$\bullet$ \emph{jAER}~\cite{jAER-software}\footnote{\url{https://jaerproject.org}}
is a Java-based environment for event sensors and processing like noise reduction, feature extraction, optical flow, de-rotation using IMU, CNN and RNN inference, etc.
\td{Several non-mobile robots~\cite{Delbruck07iscas,Conradt09iscas,Delbruck13fns, Delbruck15iscas} and even one mobile DVS~\cite{Moeys16ebccsp} robot have been built in jAER, although Java is not ideal for mobile robots.
It provides a desktop GUI based interface for easily recording and playing data that also exposes the complex internal configurations of these devices.}
It mainly supports the sensors developed at the Inst. of Neuroinformatics (UZH-ETH Zurich) that are distributed by iniVation.

$\bullet$ \emph{libcaer}\footnote{\url{https://github.com/inilabs/libcaer}} is a
minimal C library to access, configure and get data from iniVation and aiCTX neuromorphic sensors and processors.
It supports the DVS and DAVIS cameras, and the Dynap-SE neuromorphic processor.

$\bullet$ The ROS DVS package\footnote{\url{https://github.com/uzh-rpg/rpg_dvs_ros}} developed in the Robotics and Perception Group of UZH-ETH Zurich is based on libcaer.
It provides C++ drivers for the DVS and DAVIS.
It is popular in robotics since it integrates with the Robot Operating System (ROS)~\cite{Quigley09icraoss} and therefore provides high-level tools for easily recording and playing data, connecting to other sensors and actuators, etc.
Popular datasets~\cite{Mueggler17ijrr,Zhu18ral} are provided in this format.
The package also provides a calibration tool for both intrinsic and stereo calibration.

$\bullet$ The event-driven YARP Project\footnote{\url{https://github.com/robotology/event-driven}}~\cite{Glover18frobt} comprises libraries to handle neuromorphic sensors, such as the DVS, installed on the iCub humanoid robot,
along with algorithms to process event data.
It is based on the Yet Another Robot Platform (YARP) middleware.

$\bullet$ pyAER\footnote{\url{https://github.com/duguyue100/pyaer}} is a python wrapper around libcaer developed at the Inst. of Neuroinformatics (UZH-ETH) that will probably become popular for rapid experimentation.

$\bullet$ \emph{DV}\footnote{\url{https://inivation.gitlab.io/dv/dv-docs/}} is the C++ open-source software for the iniVation DVS/DAVIS and is also the software of their official SDK. 
It enables fast, easy deployment of advanced processing algorithms for next-generation embedded and IoT applications.

Other open-source software utilities and processing algorithms (in Python, CUDA, Matlab, etc.) are spread throughout the web, on the pages of the research groups working on event-based vision~\cite{Gallego17resources}.
Proprietary software includes the development kits (SDKs) developed by companies.

%% file: chapters/062_datasets_and_simulators.tex
\subsection{Datasets and Simulators}
\label{sec:datasets}
Datasets and simulators are fundamental tools to facilitate adoption of event-driven technology and advance its research.
They allow to reduce costs (currently, event cameras are considerably more expensive than standard cameras) and to monitor progress with quantitative benchmarks (as in traditional computer vision: the case of datasets such as Middlebury, MPI Sintel, KITTI, EuRoC, etc.).

The number of event-based vision datasets and simulators is growing fast.
Several of them are listed in~\cite{Gallego17resources}, sorted by task.
Broadly, they can be categorized as those that target motion estimation or image reconstruction (regression) tasks, those that target recognition (classification) tasks, and end-to-end human-labeled data like driving~\cite{Hu20arxiv}.
In the first group, there are datasets for optical flow, SLAM, object tracking, segmentation, etc.
The second group comprises datasets for object and action recognition.

\emph{Datasets for optical flow} include~\cite{Rueckauer16fns,Barranco16fns,Zhu18rss}.
Since ground-truth optical flow is difficult to acquire, \cite{Rueckauer16fns} considers only flow during purely rotational motion recorded with an IMU, and so, the dataset lacks flow due to translational (parallax) motion.
The datasets in~\cite{Barranco16fns,Zhu18rss} provide optical flow as the motion field induced on the image plane by the camera motion and the depth of the scene (measured with a range sensor, such as an \mbox{RGB-D} camera, a stereo pair or a LiDAR).
Naturally, ground truth optical flow is subject to noise and inaccuracies in alignment and calibration of the different sensors involved.

\emph{Datasets for pose estimation and SLAM} include~\cite{Weikersdorfer14icra}\footnote{\url{http://ebvds.neurocomputing.systems}},
\cite{Mueggler17ijrr,Barranco16fns,Zhu18ral,Delmerico19icra}.
The most popular one is described in~\cite{Mueggler17ijrr}, which has been used to benchmark visual odometry and visual-inertial odometry methods~\cite{Zhu17cvpr,Rebecq17bmvc,Rosinol18ral,Mueggler18tro,Gallego17ral,Reinbacher17iccp,Gallego18cvpr}.
This dataset is also popular to evaluate corner detectors~\cite{Mueggler17bmvc,Alzugaray18ral} and feature trackers~\cite{Kueng16iros,Gehrig19ijcv}.

\emph{Datasets for recognition} are currently of limited size compared to traditional computer vision ones.
They consist of cards of a deck (4 classes), faces (7 classes), handwritten digits (36 classes), gestures (rocks, papers, scissors) in dynamic scenes, cars, etc.
Neuromorphic versions of popular frame-based computer vision datasets, such as MNIST and  Caltech101, have been obtained by using saccade-like motions~\cite{Orchard15fns,Tan15fns}.
Newer datasets~\cite{Amir17cvpr,Sironi18cvpr,Miao19fnbot,Calabrese19cvprw} are acquired in real scenarios (not generated from frame-based data). 
These datasets have been used in~\cite{PerezCarrasco13pami,Orchard15pami,Lagorce17pami,Lungu17iscas,Moeys16ebccsp,Sironi18cvpr}, among others, to benchmark event-based recognition algorithms.

\label{sec:simulators}
The DVS \emph{emulators} in~\cite{Katz12iscas,Pineda16ssci} and the \emph{simulators} in~\cite{Mueggler17ijrr}
are based on the working principle of an ideal DVS pixel~\eqref{eq:EventTriggeringCondition}.
Given a virtual 3D scene and the trajectory of a moving DAVIS within it,
the simulator generates the corresponding stream of events, intensity frames and depth maps.
The simulator has been extended in~\cite{Rebecq18corl}: it uses an adaptive rendering scheme, is more photo-realistic, includes a simple event noise model and returns estimated  optical flow. 
The \textsl{v2e} tool~\cite{Delbruck2020-arxiv-v2e} generates events from video with a realistic non-ideal noisy DVS pixel model that extends modeling to low lighting conditions.
A comprehensive characterization of the noise and dynamic effects of existing event cameras has not been carried out yet, and so, the noise models used are currently somewhat oversimplified.

%% file: chapters/063_workshops.tex
\subsection{Workshops}
\label{sec:workshops}
There are two yearly Summer schools fostering research, among other topics, on event-based vision:
the Telluride Neuromorphic Cognition Engineering Workshop (27th edition in 2020, in USA)
and the Capo Caccia Workshop (12th edition in 2020, in Europe).
Recently, workshops have been organized alongside major robotics conferences (IROS'15\footnote{\url{http://innovative-sensing.mit.edu/}}, ICRA'17\footnote{\url{http://rpg.ifi.uzh.ch/ICRA17_event_vision_workshop.html}} or IROS'18\footnote{\url{http://www.jmartel.net/irosws-home}}).
Live demos of event-based systems have been shown at top-tier conferences, such as ISSCC'06, NIPS'09, CVPR'18, ECCV'18, %
ICRA'17, IROS'18, CVPR'19\footnote{\url{http://rpg.ifi.uzh.ch/CVPR19_event_vision_workshop.html}}, multiple ISCAS, etc.
As the event-based vision community grows, more workshops and live demos are expected to happen in traditional computer vision venues.

%% file: chapters/07_discussion.tex
\section{Discussion}
\label{sec:discussion}

Event-based vision is a topic that spans many fields, such as computer vision, robotics and neuromorphic engineering.
Each community focuses on exploiting different advantages of the event-based paradigm.
Some focus on the low power consumption for ``always on'' or embedded applications on resource-constrained platforms; 
others favor low latency to enable highly reactive systems, and others prefer the availability of information to better perceive the environment (high temporal resolution and HDR), with fewer constraints on computational resources.

Event-based vision is an emerging technology in the era of mature frame-based camera hardware and software. 
Comparisons are, in some terms, unfair since they are not carried out under the same maturity level.
Nevertheless event cameras show potential, able to overcome some of the limitations of frame-based cameras, reaching new scenarios previously inaccessible.
There is considerable room for improvement (research and development), as pointed out in numerous opportunities throughout the paper.

There is no agreement on what is the best method (and representation)
to process events, notably because it depends on the application.
There are different trade-offs involved, such as latency vs.~power consumption and accuracy, 
or sensitivity vs.~bandwidth and processing capacity.
For example, reducing the contrast threshold and/or increasing the resolution produces more events, which will be processed by an algorithm and platform with finite capacity.
A challenging research area is to quantify such trade-offs and to 
develop techniques to dynamically adjust the sensor and/or algorithm parameters for optimal performance. %

Another big challenge is to develop bio-inspired systems that are natively event-based end-to-end (from perception to control and actuation) that are also more efficient and long-term solutions than synchronous, frame-based systems.
Event cameras pose the challenge of rethinking perception, control and actuation, 
and, in particular, the current main stream of deep learning methods in computer vision: 
adapting them or transferring ideas to process events while being as top-performing.
Active vision (pairing perception and control) is specially relevant on event cameras because the events distinctly depends on motion, which may be due to the actuation of a robot.

Event cameras can be seen as an entry point for more efficient, near-sensor processing, such that only high-level, non-redundant information is transmitted, thus reducing bandwidth, latency and power consumption.
This could be done by pairing an event camera with hardware on the same sensor device (Speck in Section~\ref{sec:embedded_robotics}), 
or by alternative bio-inspired imaging sensors, such as cellular processor arrays \cite{Carey13vlsics} \iflongversion \cite{Martel19thesis} \fi 
which every pixel has a processor that allows to perform several types of computations with the brightness of the pixel and its neighbors.

%% file: chapters/08_conclusion.tex
\section{Conclusion}
\label{sec:conclusion}
Event cameras are revolutionary sensors that offer many advantages over traditional, frame-based cameras, such as low latency, low power, high speed and high dynamic range.
Hence, they have a large potential for computer vision and robotic applications in challenging scenarios currently inaccessible to traditional cameras. 
We have provided an overview of the field of event-based vision, covering perception, computing and control, with a focus on 
the working principle of event cameras and the algorithms developed to unlock their outstanding properties in selected applications, from low-level vision to high-level vision.
Neuromorphic perception and control are emerging topics; 
and so, there are plenty of opportunities, as we have pointed out throughout the text.
Many challenges remain ahead, and we hope that this paper provides an introductory exposition of the topic, 
as a step in humanity's longstanding quest to build intelligent machines endowed with a more efficient, bio-inspired way of perceiving and interacting with the world.

%% file: main.bbl

%% file: chapters/biographies.tex
\vskip -2\baselineskip plus -1fil
\begin{IEEEbiographynophoto}{Guillermo Gallego} is Associate Professor at Technische Universit\"at Berlin, Berlin, Germany, in the Dept. of Electrical Engineering and Computer Science.
He received the PhD degree in Electrical and Computer Engineering from the Georgia Institute of Technology, USA, in 2011. 
From 2011 to 2014 he was a Marie Curie researcher with Universidad Politecnica de Madrid, Spain, and from 2014 to 2019 he was a post-doctoral researcher at the %
University of Zurich, Switzerland.
\end{IEEEbiographynophoto}
\vskip -2\baselineskip plus -1fil
\begin{IEEEbiographynophoto}{Tobi Delbr\"uck} (M’99 SM’06 F’13) is a Professor of Physics and Electrical Engineering at ETH Zurich with the  Institute of Neuroinformatics, Zurich, Switzerland, where he has been since 1998.
He received the B.Sc. degree in physics from UC San Diego in 1986 and the Ph.D. degree from Caltech in 1993.
His group with S.-C. Liu focuses on neuromorphic sensory processing and efficient deep learning.
\end{IEEEbiographynophoto}
\vskip -2\baselineskip plus -1fil
\begin{IEEEbiographynophoto}{Garrick Orchard} is Researcher at the Neuromorphic Computing Laboratory of Intel Labs, USA.
He received the Ph.D. in Electrical and Computer Engineering in 2012 from Johns Hopkins University, USA.
From 2012 to 2019 he was Senior Research Scientist at Temasek Laboratories and Singapore Institute for Neurotechnology of the National University of Singapore. 
\end{IEEEbiographynophoto}
\vskip -2\baselineskip plus -1fil
\begin{IEEEbiographynophoto}{Chiara Bartolozzi} (IEEE Member) is Researcher at the Istituto Italiano di Tecnologia (IIT), Italy.
She earned a degree in Engineering at University of Genova (Italy) and a Ph.D. in Neuroinformatics at ETH Zurich (Switzerland), developing analog subthreshold circuits for emulating biophysical neuronal properties onto silicon and modelling selective attention on hierarchical multi-chip systems.
She leads the Neuromorphic Systems and Interfaces group at IIT, with the aim of applying neuromorphic engineering to design autonomous robotic machines.
\end{IEEEbiographynophoto}
\vskip -2\baselineskip plus -1fil
\begin{IEEEbiographynophoto}{Brian Taba} is Researcher at IBM, within the SyNAPSE project.
He received the B.S. in Electrical Engineering from California Institute of Technology in 1999 and the Ph.D. in Bioengineering from University of Pennsylvania.
\end{IEEEbiographynophoto}
\vskip -2\baselineskip plus -1fil
\begin{IEEEbiographynophoto}{Andrea Censi} is Deputy Director for the Chair of Dynamic Systems and Control (Prof. Frazzoli) at the Institute for Dynamic Systems and Control in the Department of Mechanical and Process Engineering at ETH Z\"urich.
He received the Ph.D. in Control \& Dynamical Systems from California Institute of Technology in 2012.
From 2013 to 2017 he was a postdoctoral researcher at the Laboratory for Information and Decision Systems, Massachusetts Institute of Technology, Cambridge, MA, USA. 
\end{IEEEbiographynophoto}
\vskip -2\baselineskip plus -1fil
\begin{IEEEbiographynophoto}{Stefan Leutenegger} is Senior Lecturer in Robotics at Imperial College London, UK, in the Department of Computing.
He received the PhD in Mechanical Engineering from ETH Zurich, Switzerland, in 2014, at the Autonomous Systems Lab.
Since 2014 he leads the Smart Robotics Lab at Imperial College London and co-leads research in the Dyson Robotics Lab together with Prof. A. Davison. He is co-founder of the startup SLAMcore.
\end{IEEEbiographynophoto}
\vskip -2\baselineskip plus -1fil
\begin{IEEEbiographynophoto}{Andrew Davison} is Professor of Robot Vision and Director of the Dyson Robotics Laboratory at Imperial College London. 
His research focus is on SLAM and its evolution towards general ``Spatial AI''.
He has also had strong involvement in taking this technology into real applications, in particular through his work with Dyson and as co-founder of SLAMcore. He was elected Fellow of the Royal Academy of Engineering in 2017.
\end{IEEEbiographynophoto}
\vskip -2\baselineskip plus -1fil
\begin{IEEEbiographynophoto}{J\"org Conradt}
is Associate Professor at the KTH in Stockholm, Sweden, in the School of Electrical Engineering and Computer Science. 
Before joining KTH, he was W1 Professor at the Technische Universit\"at M\"unchen, Germany. 
He holds a Ph.D. in Physics / Neuroscience from ETH Zurich, Switzerland.
He is an IEEE Senior Member, and was the founding director of the Elite Master Program NeuroEngineering at Technische Universit\"at M\"unchen. 
\end{IEEEbiographynophoto}
\vskip -2\baselineskip plus -1fil
\begin{IEEEbiographynophoto}{Kostas Daniilidis} is the Ruth Yalom Stone Professor of Computer and Information Science at the University of Pennsylvania where he has been faculty since 1998. 
He is an IEEE Fellow.
He was the director of the interdisciplinary GRASP laboratory from 2008 to 2013, Associate Dean for Graduate Education from 2012-2016, and Director of Online Learning since 2016. 
He obtained 
his PhD in Computer Science from the University of Karlsruhe, 1992.
His main interest today is in deep learning of 3D representations, data association, event-based cameras, semantic localization and mapping, and vision based manipulation. 
\end{IEEEbiographynophoto}
\vskip -2\baselineskip plus -1fil
\begin{IEEEbiographynophoto}{Davide Scaramuzza} is Associate Professor of Robotics and Perception at the University of Z\"urich, Switzerland, where he does research on autonomous, vision-based navigation of mini drones and event cameras.
He received the Ph.D. degree in robotics and computer vision from ETH Z\"urich, Switzerland, in 2008.
For his research contributions, he received a European Research Council (ERC) Grant, the IEEE Robotics and Automation Early Career Award, and several industry and paper awards.
\end{IEEEbiographynophoto}